%% file: main.tex
\documentclass[10pt,twocolumn,letterpaper]{article}

\usepackage{iccv}
\usepackage{times}
\usepackage[dvipsnames]{xcolor}
\usepackage{epsfig}
\usepackage{graphbox}
\usepackage{amsmath}
\usepackage{amssymb}
\usepackage{subfig}
\usepackage{algorithm}
\usepackage[noend]{algpseudocode}
\usepackage{varwidth}
\usepackage{times}
\usepackage{epsfig}
\usepackage{graphicx}
\usepackage{bbold}
\usepackage{amsthm}
\usepackage{capt-of}
\usepackage{multirow}
\usepackage{booktabs} 
\usepackage{CJK}
\usepackage{caption}
\usepackage{makecell}

\makeatletter
\def\BState{\State\hskip-\ALG@thistlm}
\makeatother

\definecolor{Highlight}{HTML}{39b54a}  
\newcommand{\sprv}[1]{\textcolor{Gray}{\textbf{#1}}}
\newcommand{\del}[1]{\textcolor{Highlight}{{\small\textbf{#1}}}}

\usepackage[pagebackref=true,breaklinks=true,letterpaper=true,colorlinks,bookmarks=false]{hyperref}

\iccvfinalcopy 

\def\iccvPaperID{3316} 

\begin{document}

\title{Generic Attention-model Explainability for Interpreting\\ Bi-Modal and Encoder-Decoder Transformers}

\author{Hila Chefer\textsuperscript{\rm 1} \quad Shir Gur\textsuperscript{\rm 1} \quad Lior Wolf\textsuperscript{\rm 1,2}\\
\textsuperscript{\rm 1}The School of Computer Science, Tel Aviv University\\
\textsuperscript{\rm 2}Facebook AI Research (FAIR)\\
}
\maketitle

\begin{abstract}

Transformers are increasingly dominating multi-modal reasoning tasks, such as visual question answering, achieving state-of-the-art results thanks to their ability to contextualize information using the self-attention and co-attention mechanisms. These attention modules also play a role in other computer vision tasks including object detection and image segmentation. Unlike Transformers that only use self-attention, Transformers with co-attention require to consider multiple attention maps in parallel in order to highlight the information that is relevant to the prediction in the model's input. In this work, we propose the first method to explain prediction by any Transformer-based architecture, including bi-modal Transformers and Transformers with co-attentions. We provide generic solutions and apply these to the three most commonly used of these architectures: (i) pure self-attention, (ii) self-attention combined with co-attention, and (iii) encoder-decoder attention. We show that our method is superior to all existing methods which are adapted from single modality explainability. 
Our code is available at: \url{https://github.com/hila-chefer/Transformer-MM-Explainability}.

\end{abstract}

\section{Introduction}

Multi-modal Transformers may change the way that computer vision is practiced. While the state of the art computer vision models are often trained as task-specific models that infer a fixed number of labels, Radford et al.~\cite{radford2021learning} have demonstrated that by training an image-text model that employs Transformers for encoding each modality, tens of downstream tasks can be performed, without further training (``zero-shot''), at comparable accuracy to the state of the art. Subsequently, Ramesh et al.~\cite{ramesh2021zero} used a bi-modal Transformer to generate images that match a given description in unseen domains with unprecedented performance.

These two contributions merge text and images differently. The first encodes the text with a Transformer~\cite{vaswani2017attention}, the image by either a ResNet~\cite{he2016deep} or a Transformer, and then applies a symmetric contrastive loss. The second concatenates the quantized image representation to the text tokens and then employs a Transformer model. There are also many other methods of combining text and images~\cite{tan2019lxmert, lu2019vilbert, li2020oscar, li2019visualbert}. What is common to all of these is that the mapping from the two inputs to the prediction contains interaction between the two modalities. 
These interactions often challenge the existing explainability methods that are aimed at attention-based models, since, as far as we can ascertain, all existing Transformer explainability methods (\eg, ~\cite{chefer2020transformer, abnar2020quantifying}) heavily rely on self-attention, and do not provide adaptations to any other form of attention, which is commonly used in multi-modal Transformers.

Another class of Transformer models that is not restricted to self-attention is that of Transformer encoder-decoders, \ie generative models, in which the model typically receives an input from a single domain, and produces output from a different one. 
These models are used in an emerging class of object detection~\cite{carion2020end,zhu2021deformable} and image segmentation~\cite{wang2020end,paul2021local,wang2020max} methods, and are also widely used for various NLP tasks, such as machine translation~\cite{vaswani2017attention, lewis2019bart}. In these object detection methods, for example, embeddings of the position-specific and class-specific queries are crossed with the encoded image information.

We propose the first explainability method that is applicable to all Transformer architectures, and demonstrate its effectiveness on the three most commonly used Transformer architectures: (i) pure self-attention, (ii) self-attention combined with co-attention, and (iii) encoder-decoder attention.  We use an exemplar model from each architecture, and prove our method's superiority over existing Transformer explainability methods, adapted from their single modality origin. Our explainability prescription is easier to implement than existing methods, such as ~\cite{chefer2020transformer}, and can be readily applied to any attention-based architecture.



\section{Related work}


\noindent{\bf Explainability in computer vision\quad} Interpreting computer vision algorithms usually entails the synthesis of a heatmap that depicts the computed relevancy at each image location. This can be class-dependent (for every possible label), or class-agnostic, in which case it depends only on the input and the model. Unlike most methods below, our method is of the first type. There are multiple families of explainability methods, including saliency-based  methods~\cite{dabkowski2017real,simonyan2013deep,mahendran2016visualizing,zhou2016learning,zeiler2014visualizing,zhou2018interpreting}, methods that consider activations~\cite{erhan2009visualizing} using the forward pass or the backprop~\cite{zhang2018top}, perturbation based methods~\cite{fong2019understanding,fong2017interpretable}, and methods based on Shapley-values~\cite{lundberg2017unified,chen2018lshapley}. The latter enjoy clear theoretical motivation. Theoretical justification is also given to attribution-based methods, through the theory of the Deep Taylor Decomposition~\cite{montavon2017explaining}. Such methods assign relevancy recursively from the top layer, backward, such that the sum of relevancies remains fixed. The LRP method~\cite{bach2015pixel}, is one such prominent method. Since LRP and most variants~\cite{nam2019relative,shrikumar2017learning,lundberg2017unified} are class agnostic~\cite{iwana2019explaining}, class-specific extensions were introduced~\cite{gu2018understanding,iwana2019explaining,gur2021visualization}.

Gradient-based methods directly consider the gradient of the loss with respect to the input of each layer, as computed through backpropagation. Examples include class agnostic methods~\cite{shrikumar2017learning,sundararajan2017axiomatic,smilkov2017smoothgrad,srinivas2019full}. A related class-specific approach is the  Grad-CAM method~\cite{selvaraju2017grad}, which considers the input features with the class-dependent gradient at the top layers.

\noindent{\bf Explainability for Transformers\quad}
Most attempts to explain Transformers directly  employ the attention maps. This, however, neglects the intermediate attention scores, as well as the other components of the Transformers. As noted by Chefer et al~\cite{chefer2020transformer}, the computation in each attention head mixes queries, keys, and values and cannot be fully captured by considering only the inner products of queries and keys, which is what is referred to as attention. 

LRP was applied to capture the relative importance of the attention heads within each Transformer block by Voita et al.~\cite{voita2019analyzing}. This method, however, does not propagate the relevancy scores back to the input to produce a heatmap.

Abnar et al.~\cite{abnar2020quantifying} propose a way to combine the attention scores across multiple layers. Two methods are suggested: attention rollout and attention flow. The first combines attention linearly along alternative paths in the pairwise attention graph. It is shown in~\cite{chefer2020transformer} that this method fails to distinguish between positive and negative contributions to the decision, leading to an accumulation of relevancy scores across the layers in cases for which these should be cancelled out. The attention flow method is formulated as a max-flow problem on the same pairwise attention graph. While it was shown in~\cite{abnar2020quantifying} to somewhat outperform rollout in specific scenarios, this method is too slow to support large-scale evaluations.

In contrast to these methods, Chefer et al.~\cite{chefer2020transformer} provide a comprehensive treatment of the information propagation within all components of the Transformer model, which back propagates the information through all layers from the decision back to the input.  The solution is based on Layer-wise Relevance Propagation~\cite{bach2015pixel}, with gradient integration for the self-attention layers, and is shown to be very effective for single modality Transformer encoders, such as~\cite{dosovitskiy2020image}. This method, however, does not provide a solution for attention modules other than self-attention, thus can not provide explanations for all Transformer architectures.

\noindent{\bf Transformers in computer vision\quad} Transformer technology has become increasingly prevalent for bi-modal tasks, such as image captioning and text-based image retrieval. We distinguish between networks that rely on self-attention, such as VisualBERT~\cite{li2019visualbert} and Oscar~\cite{li2020oscar} and those that also employ co-attention modules, such as LXMERT~\cite{tan2019lxmert} and ViLBERT~\cite{lu2019vilbert}. Our method provides suitable visualization for both types.

Our method also provides the first complete solution, as far as we can ascertain, for Transformer encoder-decoders~\cite{vaswani2017attention,raffel2019exploring,lewis2019bart}, which have been increasingly prevalent in computer vision. In the DETR Transformer-based detection method~\cite{carion2020end}, the image is encoded by a Transformer encoder, and the obtained information is co-attended together with queries that are both positional and class-based. Our method can be also applied to encoder-based visual Transformers, such as those used for image recognition~\cite{chen2020generative,dosovitskiy2020image,touvron2020training}, and image segmentation with a CNN decoder~\cite{zheng2020rethinking}. However, in this case, existing Transformer explainability methods can also be applied.

\section{Method}

Our method uses the model's attention layers to produce relevancy maps for each of the interactions between the input modalities in the network. In this work, we focus on image and text interactions, and attention modules for generative models, \ie, encoder-decoder attention. However, our method is easily applicable to any Transformer-based architecture, and can also be generalized to address more than two modalities. In the following, we discuss the method's propagation rules under the assumption of two modalities, \eg text and image for simplicity, followed by a detailed description of how to apply our method to each of the model types used in this work.  

Let $t, i$ be the number of text and image  input tokens respectively. To simplify notation, we use the same symbols ($t,i$) to identify variables that are associated with the two domains. 
Multi-modal attention networks contain four types of interactions between the input tokens: $\mathbf{A}^{tt}$ and $\mathbf{A}^{ii}$ are the self-attention interactions for the text and image tokens, respectively. $\mathbf{A}^{ti}$, $\mathbf{A}^{it}$ are the multi-modal attention interactions, where $\mathbf{A}^{ti}$ represents the influence of the image tokens on each text token, and $\mathbf{A}^{it}$ represents the influence of the text tokens on each image token.

In accordance with the attention interactions described, we construct a relevancy map per interaction, \ie $\mathbf{R}^{tt}$, $\mathbf{R}^{ii}$ for self-attention, and $\mathbf{R}^{ti}$, $\mathbf{R}^{it}$ for bi-modal attention.

The method calculates the relevancy maps by a forward pass on the attention layers, with each layer contributing to the aggregated relevance matrices using the update rules we will describe in the following subsections. 

\noindent{\bf Relevancy initialization\quad} Before the attention operations, each token is self-contained. Thus, self-attention interactions are initialized with the identity matrix. For bi-modal interactions, before the attention layers, each modality is separate and does not contain context from the other modality, therefore, the relevancy maps are initialized to zeros.
\begin{align}
    \label{eq:initialize-self}
    \mathbf{R}^{ii} = \mathbb{I}^{i\times i},~~~& \mathbf{R}^{tt} = \mathbb{I}^{t\times t}\\
    \mathbf{R}^{it} = \boldsymbol{0}^{i\times t},~~~& \mathbf{R}^{ti} = \boldsymbol{0}^{t\times i}
\end{align}

\noindent{\bf Relevancy update rules\quad}
  \begin{figure}
    \centering
    \begin{tabular}{cc}
    \includegraphics[width=2.5cm]{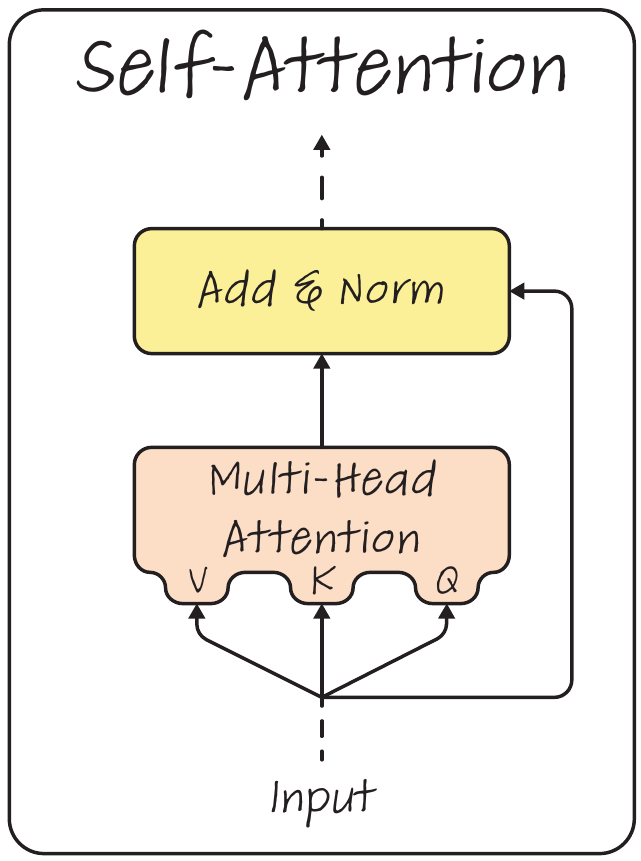}     &  \includegraphics[width=2.5cm]{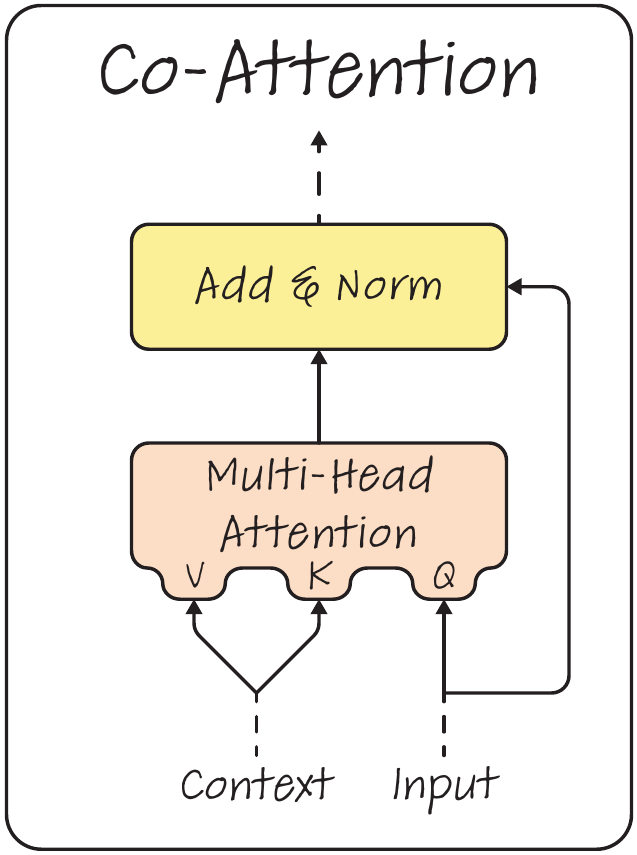}\\
         (a) &  (b)\\
    \end{tabular}
        \caption{(a) Self-attention and (b) co-attention modules.}
        \label{fig:attnmodules}
  \end{figure}
As the attention layers contextualize the tokens, our method modifies the relevancy maps that are impacted by the mixture of token embeddings. Recall the attention mechanism presented in~\cite{vaswani2017attention}:
\begin{align}
    \label{eq:attention}
    \mathbf{A}& = softmax(\frac{\mathbf{Q}\cdot{\mathbf{K}}^\top}{\sqrt{d_h}})\\
    \mathbf{O} &= \mathbf{A}\cdot\mathbf{V}
\end{align}
where $(\cdot)$ denotes matrix multiplication, $\mathbf{O} \in \mathbb{R}^{h \times s \times d_h}$ is the output of the attention module, $\mathbf{Q}\in \mathbb{R}^{h \times s \times d_h}$ is the queries matrix, and $ \mathbf{K},\mathbf{V}\in \mathbb{R}^{h \times q \times d_h}$ are the keys and values matrices. $h$ is the number of heads, $d_h$ is the embedding dimension, and $s,q\in \{i,t\}$ indicate the domains and the number of tokens in each domain, \ie, the attention takes place between $s$ query tokens and $q$ key tokens. Note that, as can be seen in Fig.~\ref{fig:attnmodules}, for self-attention layers, it holds that $s=q$ and $\mathbf{Q, K, V}$ are all projections of the input to the attention unit, while in co-attention $\mathbf{Q}$ is a projection of the input, and $\mathbf{K,V}$ are projections of the context input from the other modality. $\mathbf{A} \in \mathbb{R}^{h \times s \times q}$ is the attention map, which intuitively defines connections between each pair of tokens from $s,q$. 
Since the attention module is followed by a residual connection, as shown in Fig.~\ref{fig:attnmodules}, we accumulate the relevancies by adding each layer's contribution to the aggregated relevancies, similar to~\cite{abnar2020quantifying} in which the identity matrix is added to account for residual connections. 

Our method uses the attention map $\mathbf{A}$ of each attention layer to update the relevancy maps. Since each such map is comprised of $h$ heads, we follow~\cite{chefer2020transformer} and use gradients to average across heads. Note that Voita et al.~\cite{voita2019analyzing} show that attention heads differ in importance and relevance, thus a simple average across heads results in distorted relevancy maps. The final attention map $\bar{\mathbf{A}} \in \mathbb{R}^{s \times q}$ of our method is then defined as follows:
\begin{align}
    \label{eq:modified_att}
    \mathbf{\bar{A}} &= \mathbb{E}_h ((\nabla \mathbf{A} \odot \mathbf{A})^+)
\end{align}
where $\odot$ is the Hadamard product, 
$\nabla \mathbf{A} := \frac{\partial y_t}{\partial \mathbf{A}}$ for $y_t$ which is the model's output for the class we wish to visualize $t$, and $\mathbb{E}_h$ is the mean across the heads dimension. Following~\cite{chefer2020transformer} we remove the negative contributions before averaging. 

For self-attention layers that satisfy $\mathbf{\bar{A}}\in \mathbb{R}^{s\times s}$ the update rules for the affected aggregated relevancy scores are: 
\begin{align}
    \label{eq:self-attention-ss}
    \mathbf{R}^{ss} &=  \mathbf{R}^{ss} + \mathbf{\bar{A}} \cdot  \mathbf{R}^{ss} \\
    \label{eq:self-attention-sq}
    \mathbf{R}^{sq} &= \mathbf{R}^{sq} + \mathbf{\bar{A}} \cdot \mathbf{R}^{sq}
\end{align}
In Eq.~\ref{eq:self-attention-ss} we account for the fact that the tokens were already contextualized in previous attention layers by applying matrix multiplication with the aggregated self-attention matrix $\mathbf{R^{ss}}$, as done in~\cite{abnar2020quantifying, chefer2020transformer}. For Eq.~\ref{eq:self-attention-sq}, notice that the previous bi-modal attention layers inserted context from $q$ into $s$, therefore, when the self-attention mixes tokens from $s$, it also mixes the context $q$ in each token from $s$. The previous layers' mixture of context is embodied by $\mathbf{R}^{sq}$.  Thus, we calculate the added context from the self-attention process. 

In the case of $\mathbf{\bar{A}}\in \mathbb{R}^{s\times q}$, where a bi-modal attention is applied, the update rules of the relevancy accumulators include normalization for the self-attention matrices $\mathbf{R}^{xx}, x\in \{s,q\}$. Since we initialized $\mathbf{R}^{xx}=\mathbb{I}^{x\times x}$, and Eq.~\ref{eq:self-attention-ss} accumulates the relevancy matrices at each layer, we can consider an aggregated self-attention matrix $\mathbf{R}^{xx}$ as a matrix comprised of two parts, the first is the identity matrix from the initialization, and the second, $\mathbf{\hat{R}}^{xx} = \mathbf{R}^{xx} - \mathbb{I}^{x\times x}$ is the matrix created by the aggregation of self-attention across the layers. Since Eq.~\ref{eq:modified_att} uses gradients to average across heads, the values of $\mathbf{\hat{R}}^{xx}$ are typically reduced. We wish to account equally both for the fact that each token influences itself and for the contextualization by the self-attention mechanism. Therefore, we normalize each row in $\mathbf{\hat{R}}^{xx}$ so that it sums to 1. Intuitively, row $i$ in $\mathbf{\hat{R}}^{xx}$ disclosed the self-attention value of each token w.r.t. the $i$-th token, and the identity matrix $\mathbb{I}^{x\times x}$ sets that value for each token w.r.t. itself as 1. Thus:
\begin{align}
    \label{eq:norm-sum}
     \mathbf{\hat{S}}^{xx}_{m,n}&= \sum_{k=1}^{|x|} \mathbf{\hat{R}}^{xx}_{m,k} \\
    \label{eq:norm}
    \mathbf{\bar{R}}^{xx} &= \mathbf{\hat{R}}^{xx} / \mathbf{\hat{S}}^{xx} + \mathbb{I}^{x\times x}\,,
\end{align}
where $/$ stands for matrix division element by element. In the above, we normalize each row in $\mathbf{\hat{R}}^{xx}$ by dividing each element in the row by the sum of the row. 
Next, we define the following aggregation rules for bi-modal attention units:

\begin{align}
    \label{eq:bi-attention-sq}
    \mathbf{R}^{sq} &= \mathbf{R}^{sq} + (\mathbf{\bar{R}}^{ss})^\top \cdot \mathbf{\bar{A}} \cdot \mathbf{\bar{R}}^{qq}\\
    \label{eq:bi-attention-ss}
    \mathbf{R}^{ss} &= \mathbf{R}^{ss} + \mathbf{\bar{A}} \cdot \mathbf{R}^{qs}
\end{align}
Eq.~\ref{eq:bi-attention-sq} accounts for the fact that the tokens of each modality were already contextualized in previous attention layers by applying matrix multiplication with the normalized aggregated self-attention matrices $\mathbf{\bar{R}}^{ss}, \mathbf{\bar{R}}^{qq}$. 

For Eq.~\ref{eq:bi-attention-ss}, notice that the previous bi-modal attention layers integrate the embeddings of the two modalities, thus when contextualizing $s$ with $q$, $q$ also contains information from $s$, embodied in $\mathbf{R}^{qs}$.

Note that the above rules are described w.r.t. input from modality $s\in \{i,t\}$, and context from modality $q\in \{i,t\}$ \ie the rules are symmetrically applied to both modalities, image and text.

\subsection{Obtaining classification relevancies}
\label{sec:33}
In order to make the final classification, Transformer-based models usually regard the \texttt{[CLS]} token, which is a token that is added to the input tokens and constructs a general representation of all the input tokens. To retrieve per-token relevancies for classification tasks, one can consider the row corresponding to the \texttt{[CLS]} token in the corresponding relevancy map. For instance, assuming the \texttt{[CLS]} token is the first token in the text modality, to extract relevancies per text token, one should consider the first row of $\mathbf{R}^{tt}$, and to extract the image token relevancies, consider the first row in $\mathbf{R}^{ti}$ which describes the connections between the \texttt{[CLS]} token and each image token.

\subsection{Adaptation to attention types}
\label{sec:adaptation}

\begin{figure*}[t!]%
    \centering
    \begin{tabular}{ccc}
    {{\includegraphics[width=3cm]{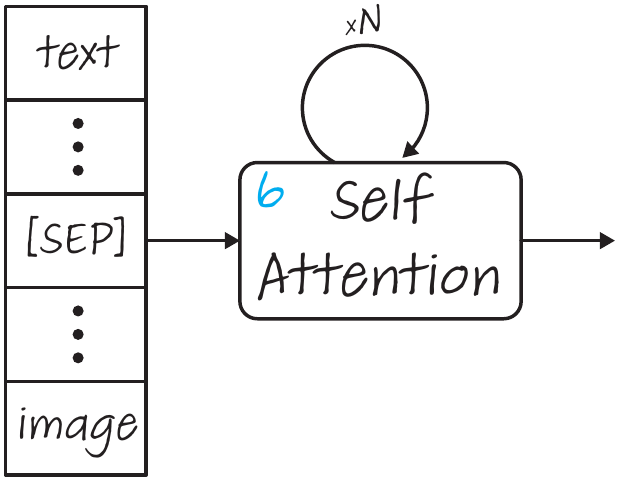}}}&%
    {{\includegraphics[width=6cm]{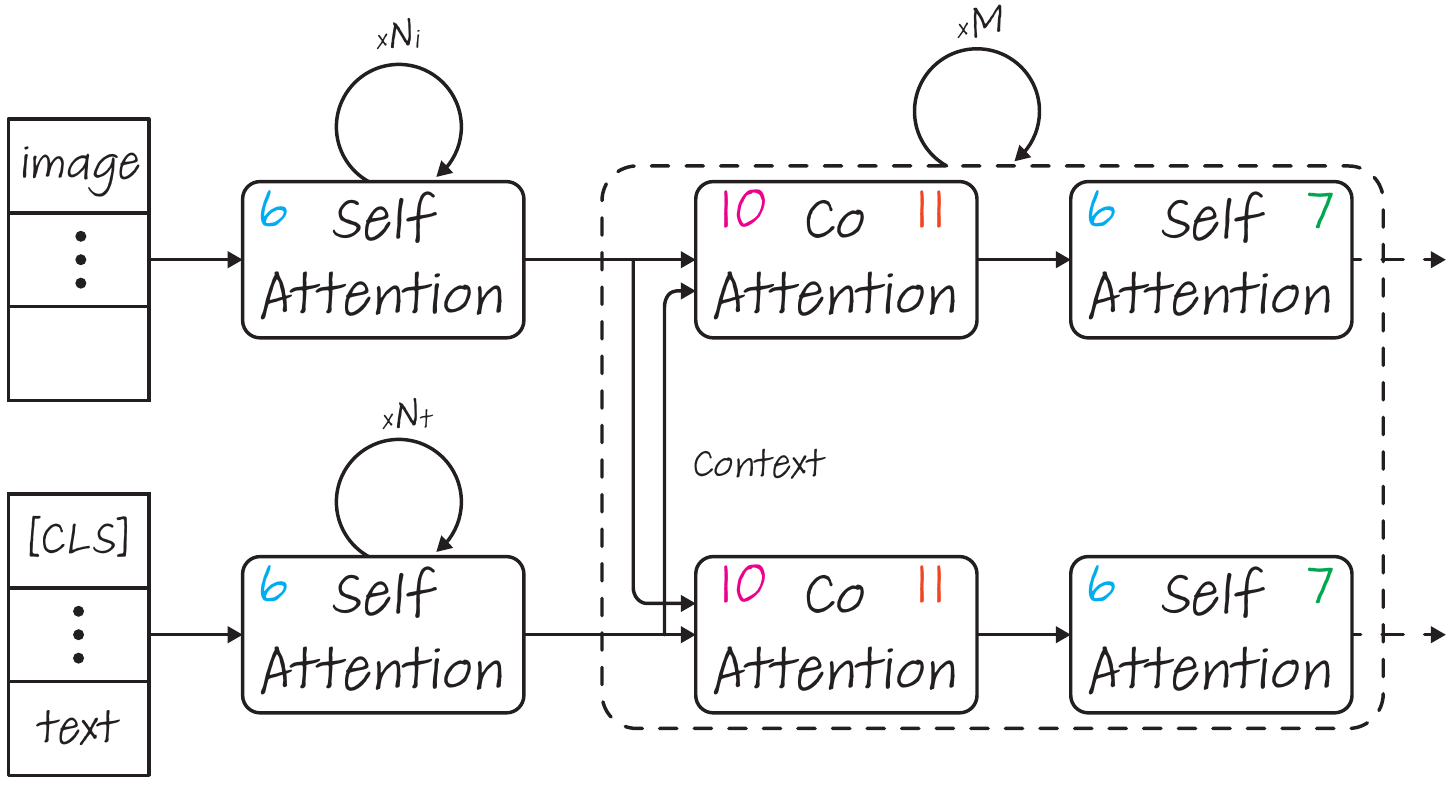} }}&%
    {{\includegraphics[width=6cm]{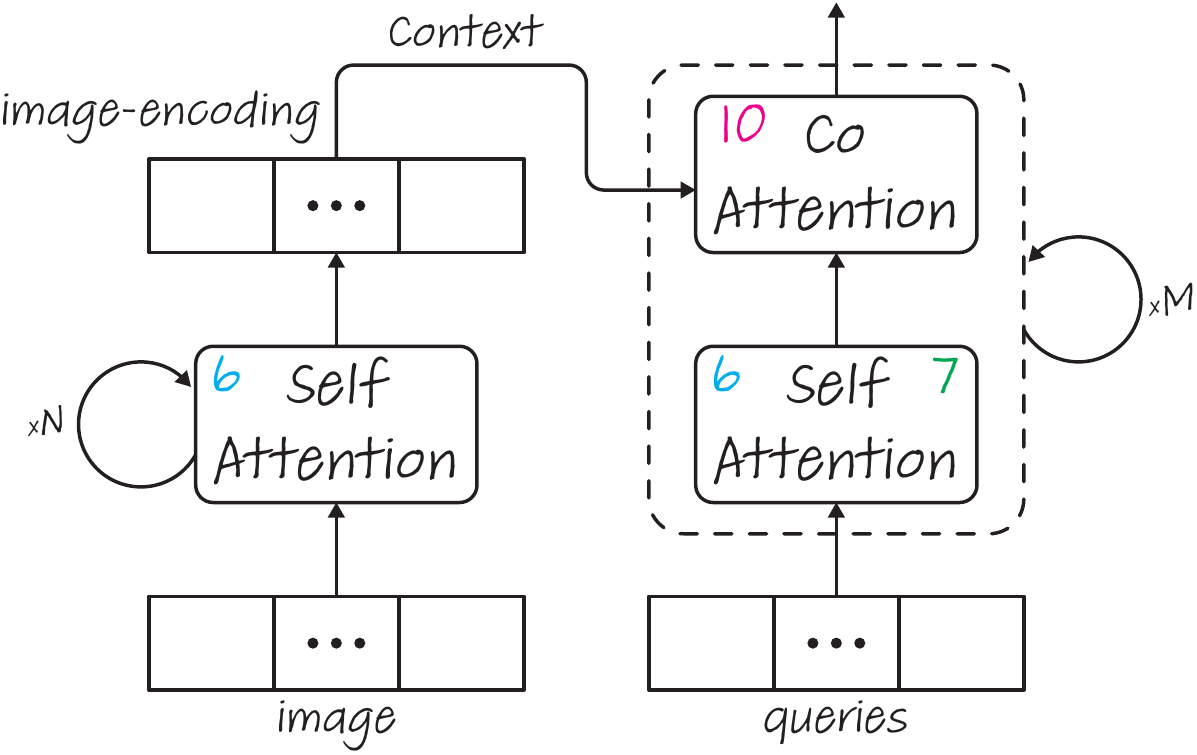} }}\\
    (a) & (b) & (c)\\
    \end{tabular}
    \caption{Illustration of the three architecture types presented in our work. The numbers in each attention module represent the Eq. number of the rule applied by our method on the module's forward pass. (a) VisualBERT: a pure self-attention architecture. (b) LXMERT: self-attention with co-attention encoder architecture. (c)  DETR: encoder-decoder architecture.}%
    \label{fig:threetypes}%
\end{figure*}

In this work, we examine our method on three different types of attention mechanisms used in Transformer-based networks. The architectures and matching propagation rules are visualized in Fig.~\ref{fig:threetypes}. The first architecture type is a multi-modal Transformer, where the two modalities are concatenated and separated by the \texttt{[SEP]} token~\cite{li2019visualbert,li2020oscar}, as demonstrated in Fig.~\ref{fig:threetypes}(a). Such networks only use self-attention to contextualize the modalities, \ie only Eq.~\ref{eq:self-attention-ss}. Since the model is based on pure self-attention, we produce one relevancy map $\mathbf{R}^{(t+i,t+i)}$ which defines connections between the modalities, as well as within each modality. In order to visualize the tokens related to the classification, one should consider the row of $\mathbf{R}^{(t+i,t+i)}$ which corresponds to the token used for classification. This row $\mathbf{R_{cls}}^{(t+i)}$ yields a relevancy score per image token and per text token. 

The second type is a multi-modal attention network that incorporates co-attention modules that contextualize each modality with the other modality~\cite{tan2019lxmert,lu2019vilbert}, as can be seen in Fig.~\ref{fig:threetypes}(b). Such networks require all propagation rules described above, for each modality. To produce relevancies for the classification, we simply follow the example in Sec~\ref{sec:33}, since as Fig.~\ref{fig:threetypes}(b) depicts, the $\texttt{[CLS]}$ token in this case is the first token of the text modality.

The third and last type is a generative model where there is one input modality, and the output is from a different domain~\cite{carion2020end,zhu2021deformable,wang2020end,paul2021local,wang2020max,vaswani2017attention, lewis2019bart}, which is visualized in Fig.~\ref{fig:threetypes}(c). Such networks contain an encoder that utilizes self-attention on the input and a decoder. The decoder has two types of inputs. The first is the encoded data, which remains unchanged,
and the second are inputs from the decoder's domain. 
The decoder proceeds to utilize self-attention on the decoder domain's tokens, followed by a co-attention layer contextualizing them with the encoder's output. To clarify, in this case, the relevance update rules are as follows: notate by $e$ the encoder's tokens, 
and by $d$ the decoder's tokens. The relevancy matrices are: $\mathbf{R}^{ee}, \mathbf{R}^{dd}$ for the self-attention interactions, and $\mathbf{R}^{de}$ for the bi-modal interactions between the decoder's tokens and the encoder's tokens. Notice that since the encoder is not contextualized, we do not have a relevancy matrix $\mathbf{R}^{ed}$. The encoder's self-attention calculation for $\mathbf{R}^{ee}$ simply follows Eq.~\ref{eq:self-attention-ss}. For the decoder's self-attention, we apply Eq.~\ref{eq:self-attention-ss},~\ref{eq:self-attention-sq}.
For the bi-modal attention in the decoder, we follow Eq.~\ref{eq:bi-attention-sq} to account for self-attention in the encoder and the decoder. Notice that Eq.~\ref{eq:bi-attention-ss} is irrelevant since we do not have a relevancy map for $\mathbf{R}^{qs}=\mathbf{R}^{ed}$. In order to extract relevancies in this case, we consider the relevancy map $\mathbf{R}^{de}$. In this work, we use an object detection model as our exemplar encoder-decoder architecture. For such models, each token from $d$ is a query representing an object in the input image. In order to produce relevancy for each of the image regions w.r.t. an object $j$ that was detected, one should consider the $j$-th row of $\mathbf{R}^{de}$, which corresponds to the $j$-th detection. $\mathbf{R}_j^{de}$ contains a relevancy score per each encoder token, which is in this case an image region.   


\section{Baselines}

We focus on methods that are both common in the explainability literature, and applicable to the extensive tests we report in this work. 
We present baselines of three classes, following~\cite{chefer2020transformer}: attention map baselines, gradient baselines, and relevancy map baselines. Our attention map baselines are raw attention and rollout. Raw attention regards only the last layer's attention map as the relevancy map, \eg $\mathbf{R}^{tt}= \mathbf{A}^{tt}$, where $\mathbf{A}^{tt}$ is the last text self-attention map. The second is rollout, which follows~\cite{abnar2020quantifying} for all the self-attention layers. Since the rollout baseline is based solely on self-attention, to distinguish from raw attention,
we employ the following for $\mathbf{R}^{sq}, s,q\in \{t,i\}$:
\begin{align}
    \label{eq:baselines-cross}
    \mathbf{R}^{sq} &= (\mathbf{R}^{ss})^\top \cdot \mathbf{\bar{A}} \cdot \mathbf{R}^{qq}
\end{align}
where $\mathbf{R}^{ss}, \mathbf{R}^{qq}$ are the self-attention relevancies computed by rollout, and $\mathbf{\bar{A}\in \mathbb{R}^{s\times q}}$ is the last bi-attention map.
For our gradient baselines, we use the Grad-CAM~\cite{selvaraju2017grad} adaptation described in~\cite{chefer2020transformer}, \ie, we examine the last attention layer, and perform Grad-CAM on the attention map's heads.
Lastly, our relevancy map baselines include partial LRP, following~\cite{voita2019analyzing},  which uses the LRP relevancy values of the last attention layer to average across the heads, 
and the Transformer attribution method described in~\cite{chefer2020transformer}. The method in ~\cite{chefer2020transformer} employs Eq.~\ref{eq:modified_att} for all attention layers in order to average across heads, in the following way:
\begin{align}
    \label{eq:modifie_att_lrp}
    \mathbf{\bar{A}} &= \mathbb{E}_h ((\nabla \mathbf{A} \odot \mathbf{R^A})^+)
\end{align}
where the only difference compared to Eq.~\ref{eq:modified_att}, is that \cite{chefer2020transformer} uses the LRP~\cite{bach2015pixel} relevancy values of $\mathbf{A}$, \ie $\mathbf{R^A}$, instead of using the raw attention maps as done in Eq.~\ref{eq:modified_att}. 
Additionally,~\cite{chefer2020transformer} uses Eq.~\ref{eq:self-attention-ss} for all self-attention layers. For non self-attention layers, our version of~\cite{chefer2020transformer} takes the last attention map, and averages across heads using Eq.~\ref{eq:modifie_att_lrp}.
Note that while applying our method only requires a few simple hooks for the attention modules, LRP requires a custom implementation of all network layers.

\begin{figure*}[t!]%
    \centering
    \begin{tabular}{cccc}
    {{\includegraphics[width=4cm]{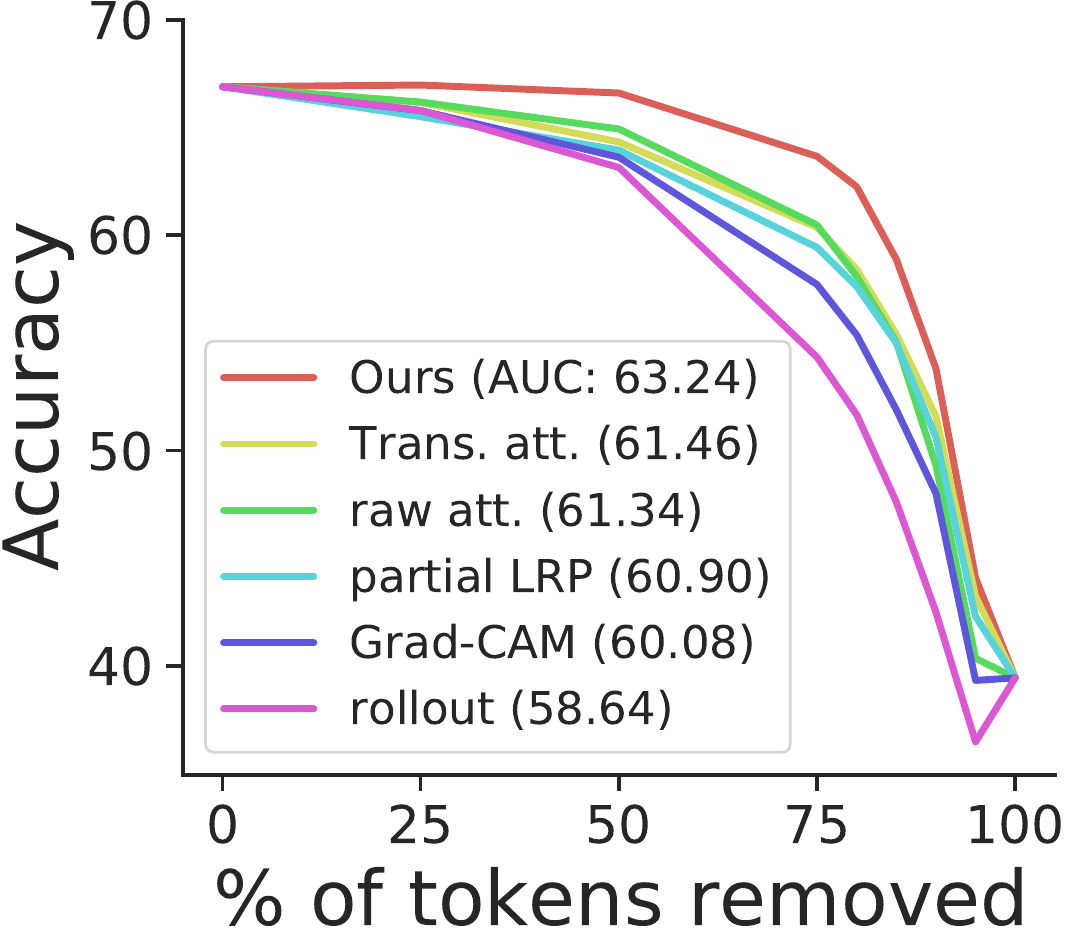}}}&%
    {{\includegraphics[width=4cm]{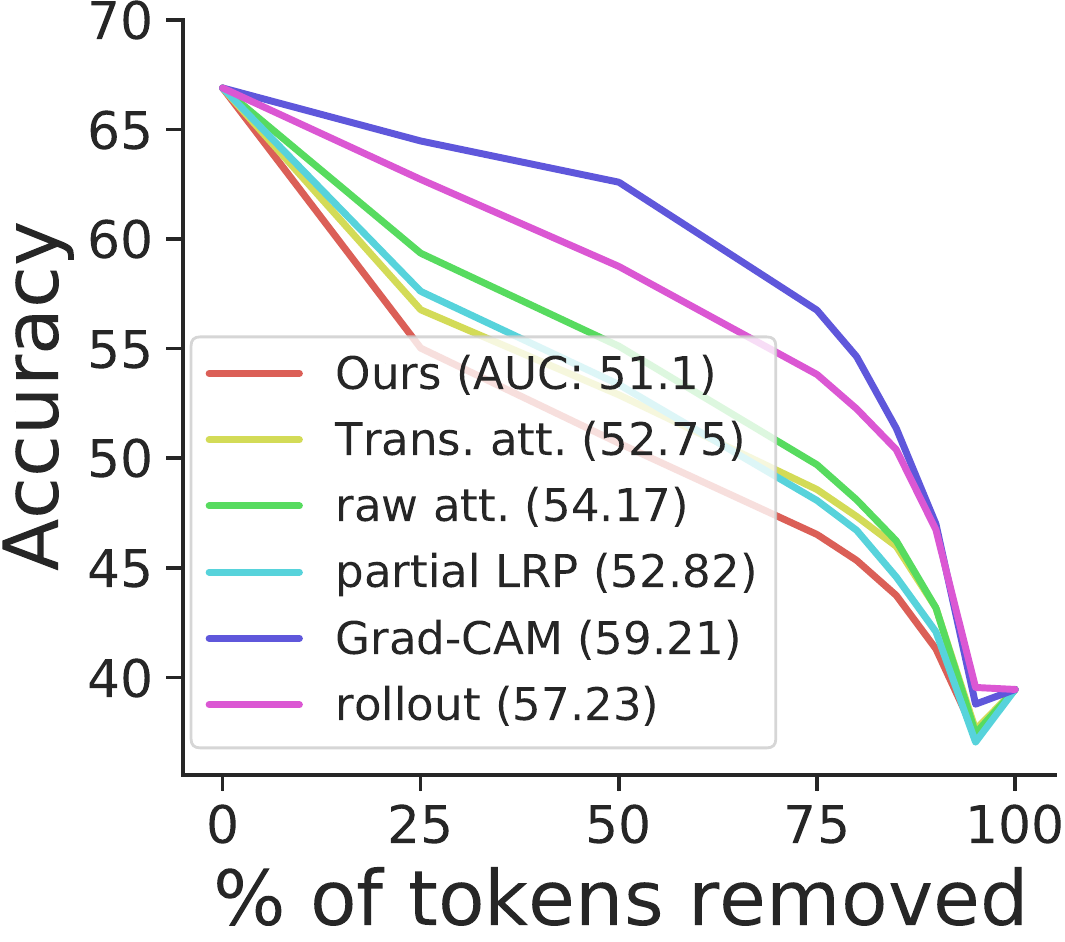} }}&%
    {{\includegraphics[width=4cm]{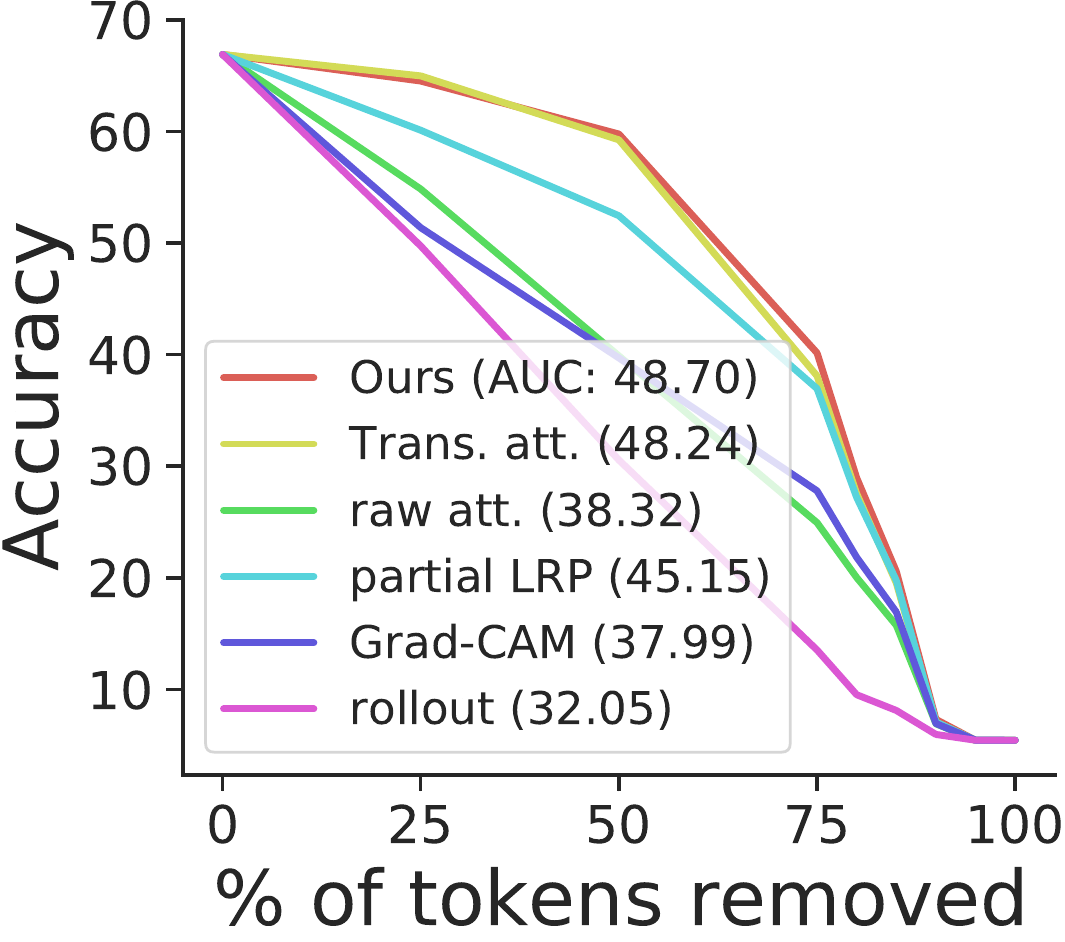}}}&%
    {{\includegraphics[width=4cm]{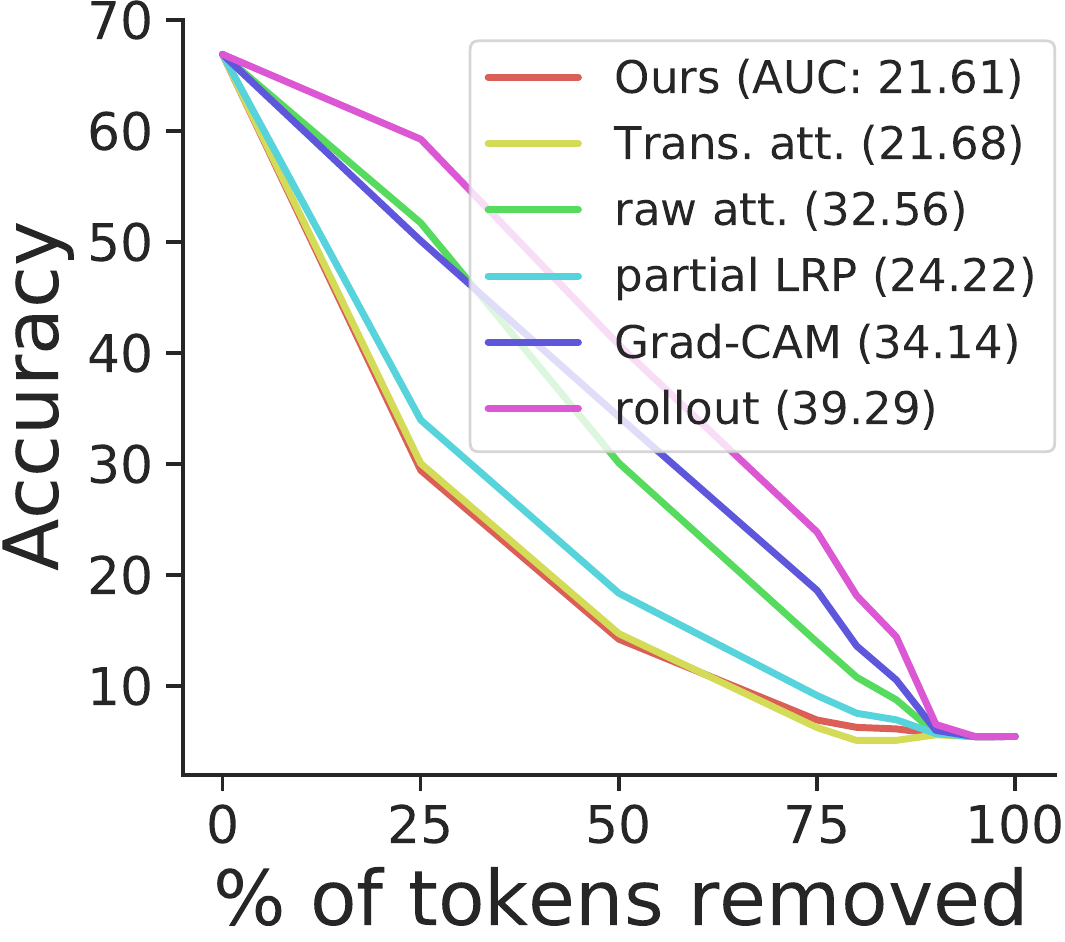} }}\\
    (a) & (b) & (c) & (d)\\
    \end{tabular}
    \caption{LXMERT perturbation test results. For negative perturbation, larger AUC is better; for positive perturbation, smaller AUC is better. (a) negative perturbation on image tokens, (b) positive perturbation on image tokens, (c) negative perturbation on text tokens, and (d) positive perturbation on text tokens.}%
    \label{fig:LXMERT-pert}%
\end{figure*}



\begin{figure*}[t]
    \setlength{\tabcolsep}{1pt} 
    \renewcommand{\arraystretch}{1} 
    \begin{tabular*}{\linewidth}{@{\extracolsep{\fill}}lcccc}
     &
    \begin{CJK*}{UTF8}{gbsn}
{\setlength{\fboxsep}{0pt}\colorbox{white!0}{\parbox{0.2\textwidth}{    \colorbox{red!58.1370735168457}{\strut is} \colorbox{red!0.0}{\strut the} \colorbox{red!0.0}{\strut animal} \colorbox{red!65.44914245605469}{\strut eating} \colorbox{red!100.0}{\strut ?} \colorbox{red!1.2035624980926514}
}}}
\end{CJK*}&
    \begin{CJK*}{UTF8}{gbsn}
{\setlength{\fboxsep}{0pt}\colorbox{white!0}{\parbox{0.2\textwidth}{   \colorbox{red!59.360687255859375}{\strut did} \colorbox{red!47.44766616821289}{\strut he} \colorbox{red!41.36971664428711}{\strut catch} \colorbox{red!0.0}{\strut the} \colorbox{red!100.0}{\strut ball} \colorbox{red!87.9761734008789}{\strut ?}
}}}
\end{CJK*} &
    \begin{CJK*}{UTF8}{gbsn}
{\setlength{\fboxsep}{0pt}\colorbox{white!0}{\parbox{0.2\textwidth}{ \colorbox{red!4.920843124389648}{\strut is} \colorbox{red!0.0}{\strut the} \colorbox{red!100.00000762939453}{\strut tub} \colorbox{red!40.00966262817383}{\strut white} \colorbox{red!42.100624084472656}{\strut ?}
}}}
\end{CJK*} &
    \begin{CJK*}{UTF8}{gbsn}
{\setlength{\fboxsep}{0pt}\colorbox{white!0}{\parbox{0.2\textwidth}{   \colorbox{red!30.391498565673828}{\strut did} \colorbox{red!5.028392791748047}{\strut the} \colorbox{red!26.076675415039062}{\strut man} \colorbox{red!35.311214447021484}{\strut just} \colorbox{red!100.0}{\strut catch} \colorbox{red!16.184215545654297}{\strut the} \colorbox{red!5.848325729370117}{\strut fr}\colorbox{red!0.0}{\strut is}\colorbox{red!20.73851203918457}{\strut bee} \colorbox{red!94.43649291992188}{\strut ?} 
}}}
\end{CJK*}\\
    \rotatebox[origin=c]{90}{~~~~~Ours} &
    \includegraphics[align=c,height=0.165\linewidth]{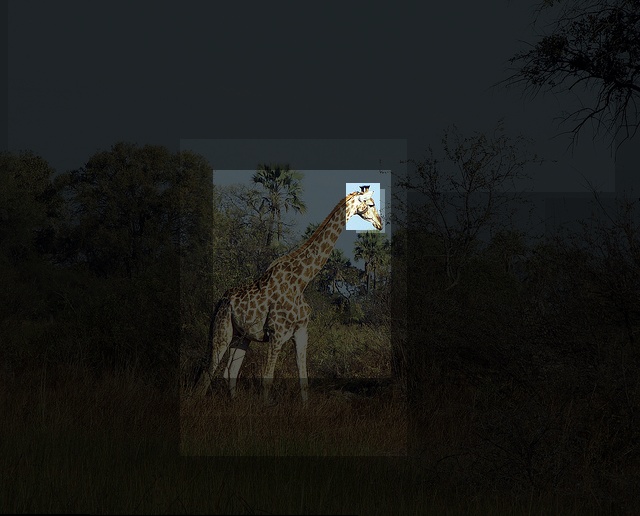} &
    \includegraphics[align=c,height=0.165\linewidth]{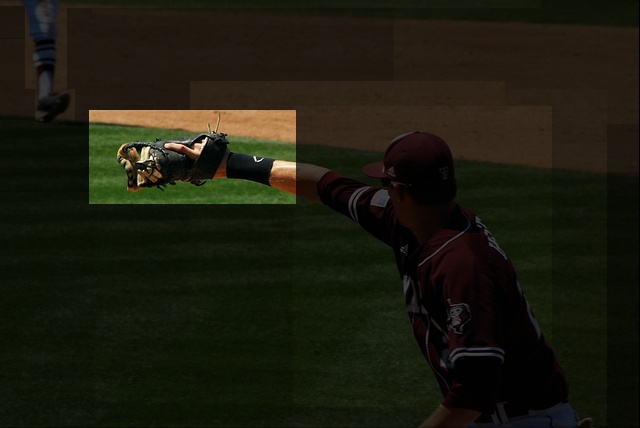} &
    \includegraphics[align=c,height=0.165\linewidth]{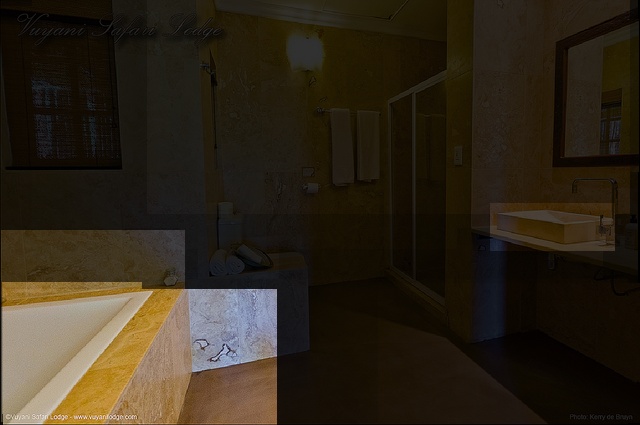} &
    \includegraphics[align=c,height=0.165\linewidth]{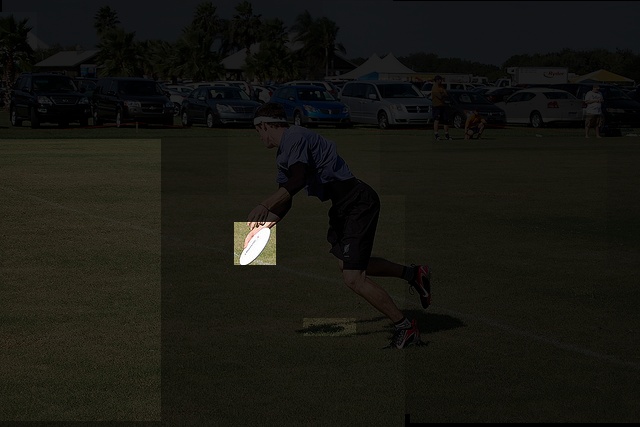} 
    \\\\
    &

\begin{CJK*}{UTF8}{gbsn}
{\setlength{\fboxsep}{0pt}\colorbox{white!0}{\parbox{0.2\textwidth}{   \colorbox{red!54.17789077758789}{\strut is} \colorbox{red!0.0}{\strut the} \colorbox{red!3.314396858215332}{\strut animal} \colorbox{red!59.200435638427734}{\strut eating} \colorbox{red!100.0}{\strut ?}
}}}
\end{CJK*} &
    \begin{CJK*}{UTF8}{gbsn}
{\setlength{\fboxsep}{0pt}\colorbox{white!0}{\parbox{0.2\textwidth}{  \colorbox{red!38.96585464477539}{\strut did} \colorbox{red!22.602096557617188}{\strut he} \colorbox{red!44.653099060058594}{\strut catch} \colorbox{red!6.535938262939453}{\strut the} \colorbox{red!100.0}{\strut ball} \colorbox{red!69.22191619873047}{\strut ?}
}}}
\end{CJK*} &
    \begin{CJK*}{UTF8}{gbsn}
{\setlength{\fboxsep}{0pt}\colorbox{white!0}{\parbox{0.2\textwidth}{ \colorbox{red!18.032072067260742}{\strut is} \colorbox{red!2.069951057434082}{\strut the} \colorbox{red!64.09111022949219}{\strut tub} \colorbox{red!99.99999237060547}{\strut white} \colorbox{red!53.09128189086914}{\strut ?}
}}}
\end{CJK*} &
    \begin{CJK*}{UTF8}{gbsn}
{\setlength{\fboxsep}{0pt}\colorbox{white!0}{\parbox{0.2\textwidth}{  \colorbox{red!21.42181968688965}{\strut did} \colorbox{red!0.0}{\strut the} \colorbox{red!1.4645568132400513}{\strut man} \colorbox{red!21.948606491088867}{\strut just} \colorbox{red!15.582681655883789}{\strut catch} \colorbox{red!0.0}{\strut the} \colorbox{red!2.117180347442627}{\strut fr}\colorbox{red!2.7323646545410156}{\strut is}\colorbox{red!2.486506700515747}{\strut bee} \colorbox{red!100.00000762939453}{\strut ?}
}}}
\end{CJK*}\\
\rotatebox[origin=c]{90}{~~~~~Trans. att.~\cite{chefer2020transformer}} &
    \includegraphics[align=c,height=0.165\linewidth]{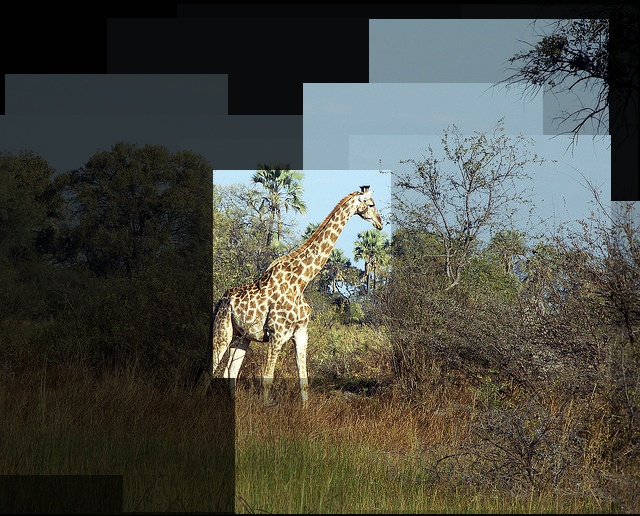} &
    \includegraphics[align=c,height=0.165\linewidth]{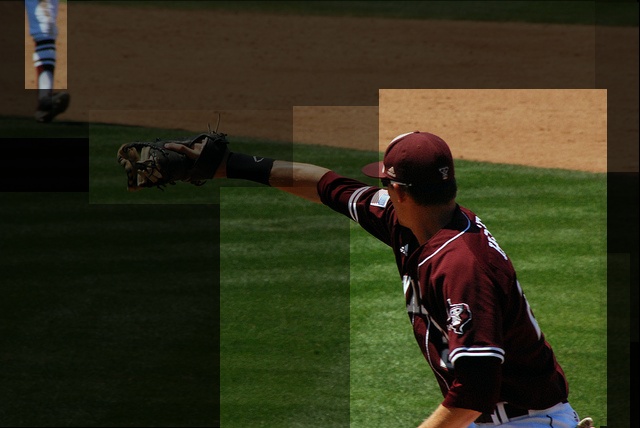} &
    \includegraphics[align=c,height=0.165\linewidth]{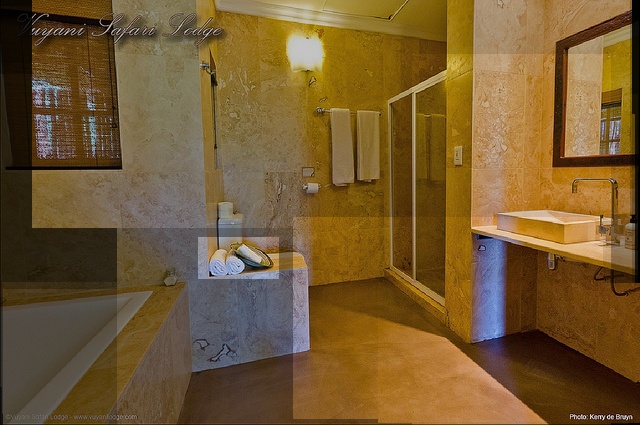} &
    \includegraphics[align=c,height=0.165\linewidth]{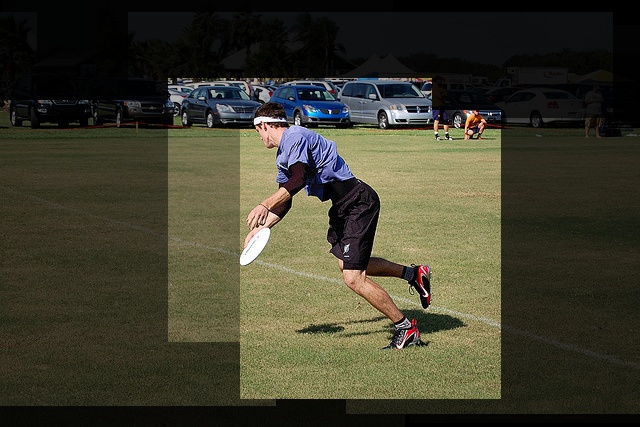} 
    \\
    
    \end{tabular*}
    \caption{A comparison between our method (top) and the method of~\cite{chefer2020transformer} (bottom) for VQA with the LXMERT model. Relevancy for text is given as shades of red. Relevancy for images is given by multiplying each region by the relative relevancy. The results for the text part are similar. For the images, our method provides much more focused results. Both observations are aligned with the quantitative results. Answers (left to right): no, yes, yes, no.} 
    \label{fig:LXMERT-vis}
\end{figure*}

\begin{figure*}[t!]%
    \centering
    \begin{tabular}{cccc}
    {{\includegraphics[width=4cm]{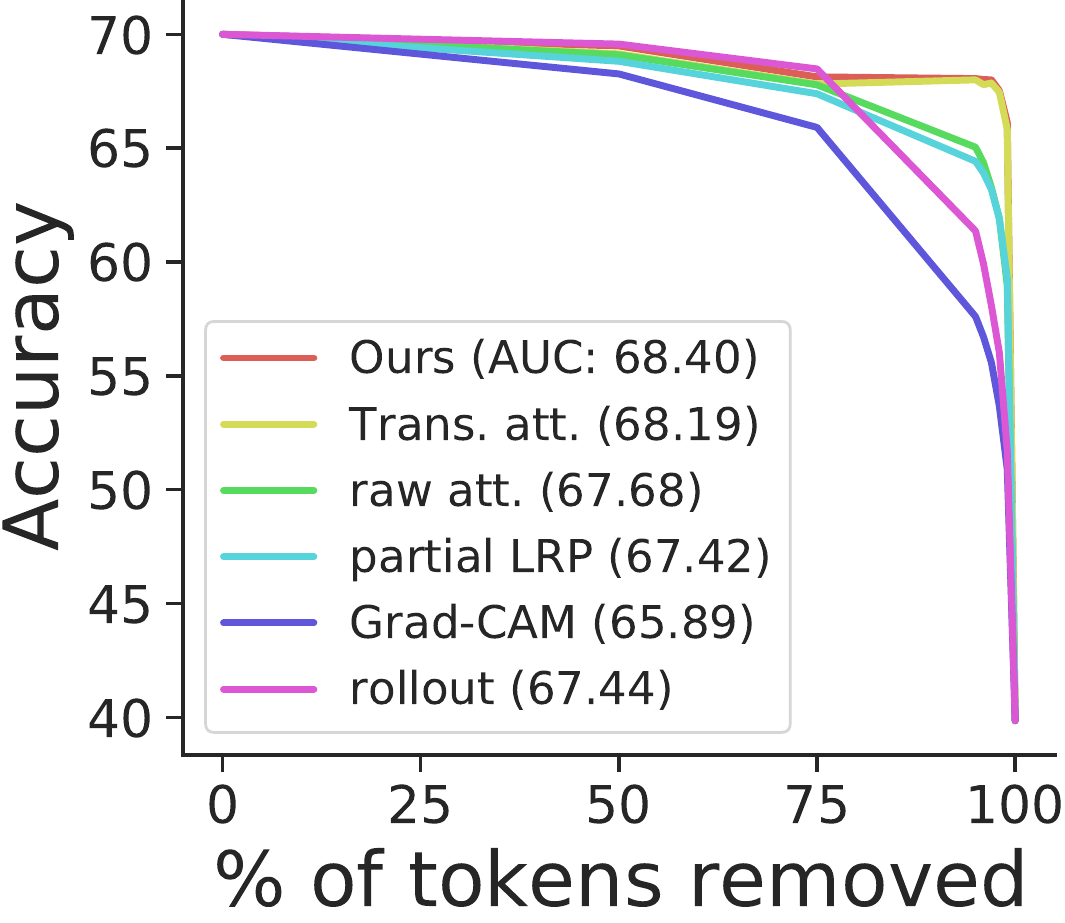}}}&%
    {{\includegraphics[width=4cm]{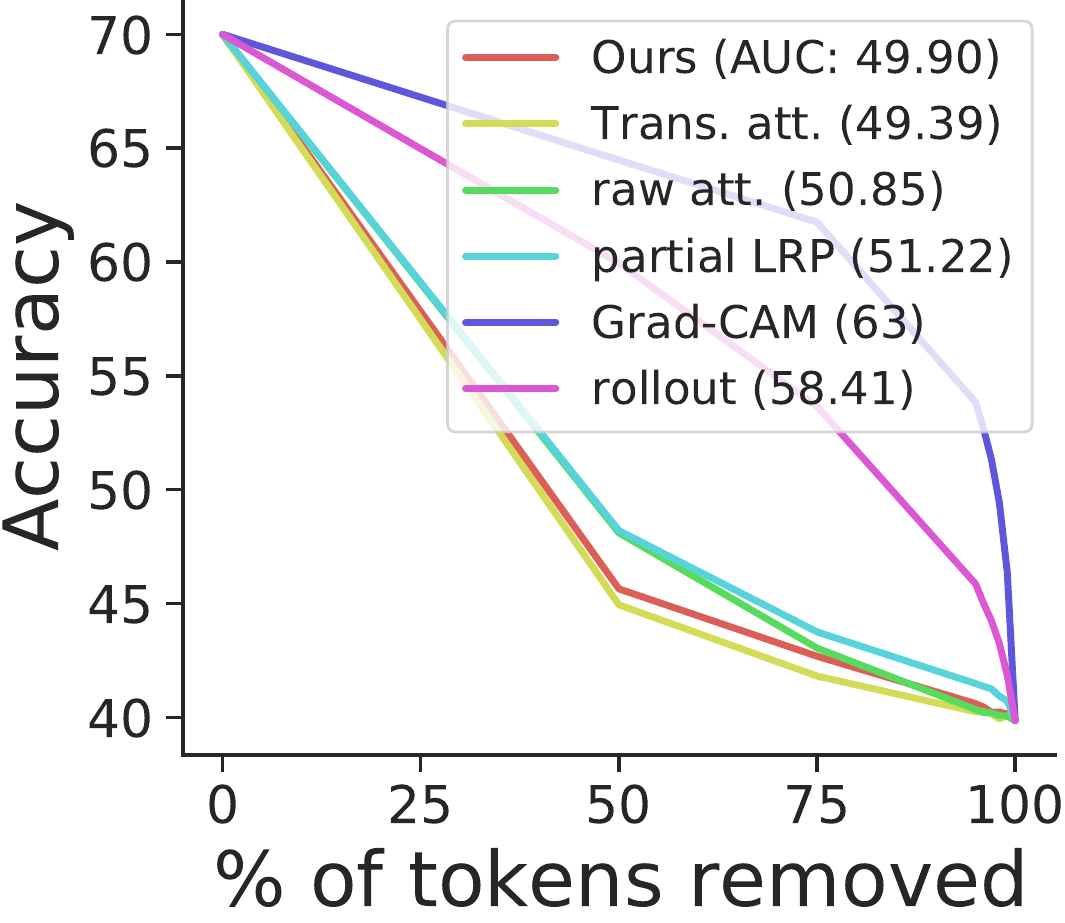} }}&
     {{\includegraphics[width=4cm]{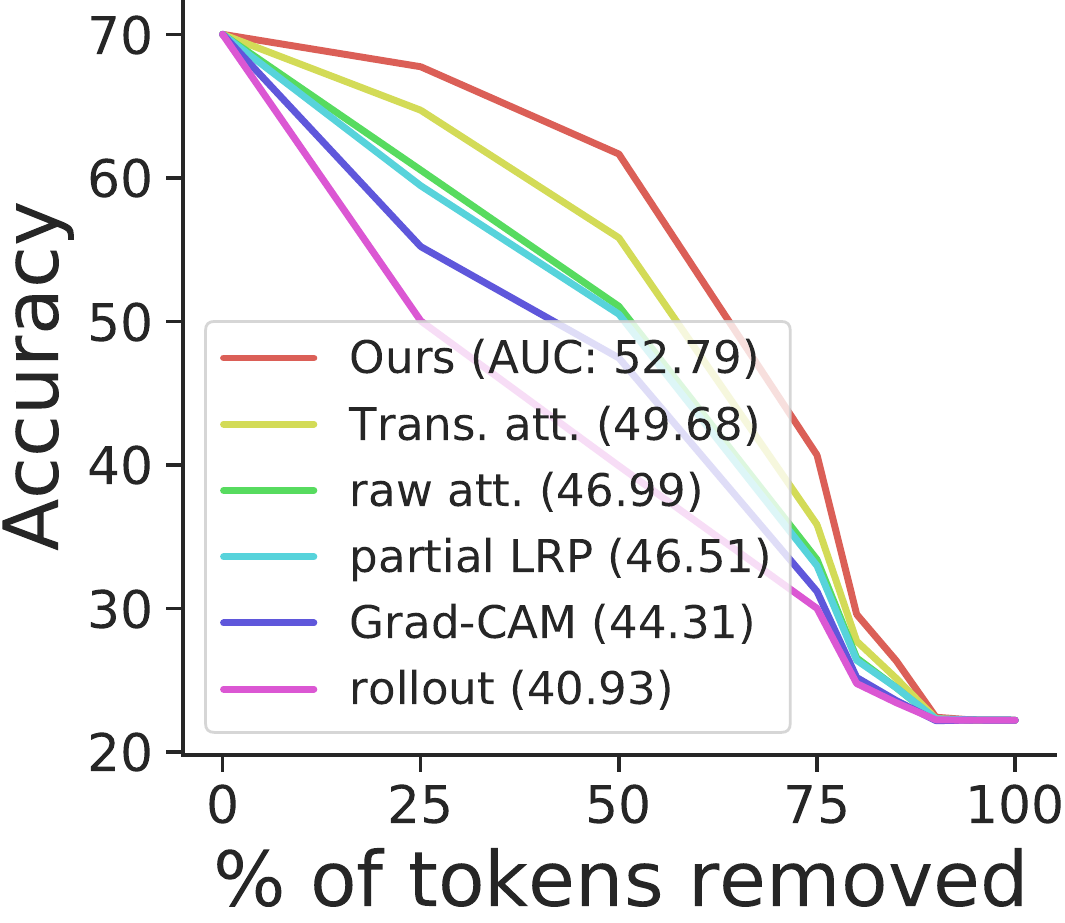}}}& 
     {{\includegraphics[width=4cm]{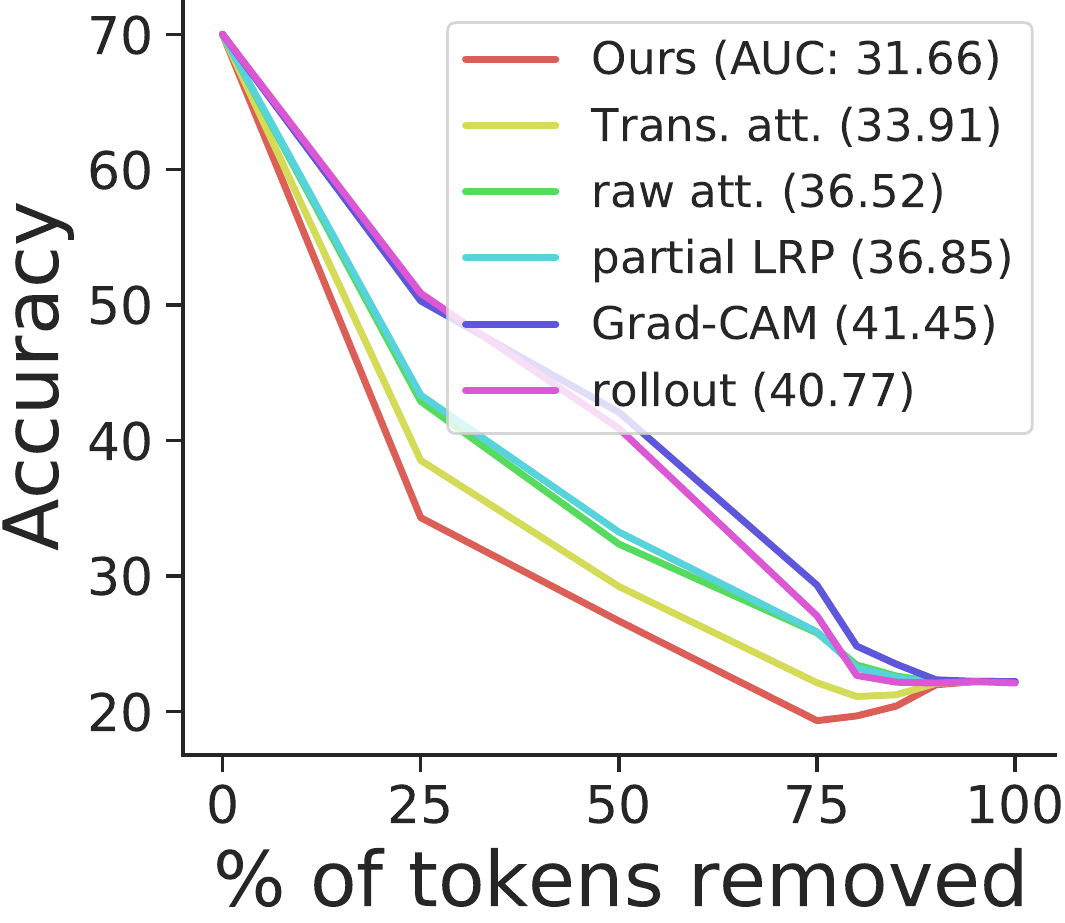}}}
    \\
    (a) & (b) & (c) & (d)\\
    \end{tabular}
    \caption{VisualBERT perturbation test results. For negative perturbation, larger AUC is better; positive perturbation, smaller AUC is better. (a) negative perturbation on image tokens, (b) positive perturbation on image tokens, (c) negative perturbation on text tokens, and (d) positive perturbation on text tokens.}%
    \label{fig:VisualBERT-pert}%
\end{figure*}

\begin{table*}[t]
    \centering
    \begin{tabular*}{\linewidth}{@{\extracolsep{\fill}}llcccccc}
        \toprule
        &Supervised & \multicolumn{6}{c}{Weakly supervised segmentation}\\
        \cmidrule{3-8}
        &\multirow{2}{*}{detection} &\multirow{2}{*}{rollout~\cite{abnar2020quantifying}} & raw & \multirow{2}{*}{Grad-CAM~\cite{selvaraju2017grad}}  & partial & Trans. & \multirow{2}{*}{Ours}\\
        &&&attention&&LRP~\cite{voita2019analyzing}& attribution~\cite{chefer2020transformer}&\\
        \midrule
        AP & \sprv{51.8} & 0.1 & 5.6 & 2.3 & 4.7 & 7.2 & \textbf{13.1} \del{(+5.9)}\\
        $\text{AP}_{medium}$ & \sprv{56.3} & 0.1 & 9.6 & 2.3 & 8.0 & 10.4 & \textbf{14.4} \del{(+4.0)}\\
        $\text{AP}_{large}$ & \sprv{67.6} & 0.2 & 6.9 & 4.7 & 5.1 & 12.4 & \textbf{24.6} \del{(+12.2)}\\
        
        AR & \sprv{67.4} & 0.4 & 11.7 & 5.5 & 10.4 & 13.4 & \textbf{19.3} \del{(+5.9)}\\
        $\text{AR}_{medium}$ & \sprv{72.8} & 0.1 & 21.8 & 5.9 & 19.9 & 21.0 & \textbf{23.9} \del{(+2.1)}\\
        $\text{AR}_{large}$ & \sprv{85.1} & 0.9 & 10.8 & 10.7 & 8.0 & 19.4 & \textbf{33.2} \del{(+13.8)}\\
        \bottomrule
    \end{tabular*}
   
    \caption{DETR~\cite{zhu2021deformable}-based weakly supervised segmentation results on the MSCOCO~\cite{ty2014coco} validation set, higher is better. AP=average precision, AR=average recall. The subscripts indicate benchmark subsets. The first column is the DETR~\cite{zhu2021deformable} bounding boxes detection scores obtained for each category, while the rest of the columns are for segmentation maps.}
    \label{tab:DETR}
\end{table*}
        
   

\begin{table}[t]
\centering
\begin{tabular}{@{}l@{~~}l@{~}c@{~}c@{~}c@{~}c@{~}c@{~}c@{}}
        \toprule
        &&rollout & raw att.& GCAM & LRP& T. Attr & Ours\\
        \midrule
        \multirow{2}{*}{N} &Predicted & 53.10 & 45.55 & 41.52  & 50.49 & 54.16 &\textbf{54.61}\\
        &Target & - & - & 42.02 & 50.49 & 55.04 & \textbf{55.67} \\
        \midrule
        \multirow{2}{*}{P} &Predicted & 20.05 & 23.99 & 34.06 & 19.64 & \textbf{17.03} & 17.32\\
        &Target & - & - & 33.56 & 19.64 & \textbf{16.04} & 16.72\\
        \bottomrule
    \end{tabular}
    \caption{ViT~\cite{dosovitskiy2020image} positive (P) and negative (N) perturbation AUC results for the predicted and target classes, on the ImageNet~\cite{russakovsky2015ImageNet} validation set. For negative perturbation, larger AUC is better; positive perturbation, smaller AUC is better. GCAM=Grad-CAM; T. Attr = Transformer attribution~\cite{chefer2020transformer}.}
    \label{tab:ViT}
\end{table}

\begin{figure*}[t]
    \setlength{\tabcolsep}{1pt} 
    \renewcommand{\arraystretch}{1} 
    \begin{center}
    \begin{tabular*}{\linewidth}{@{\extracolsep{\fill}}lcccccccccc}
    \parbox{2cm}{Detection\\bounding\\box} &
    \includegraphics[align=c,width=0.09\linewidth]{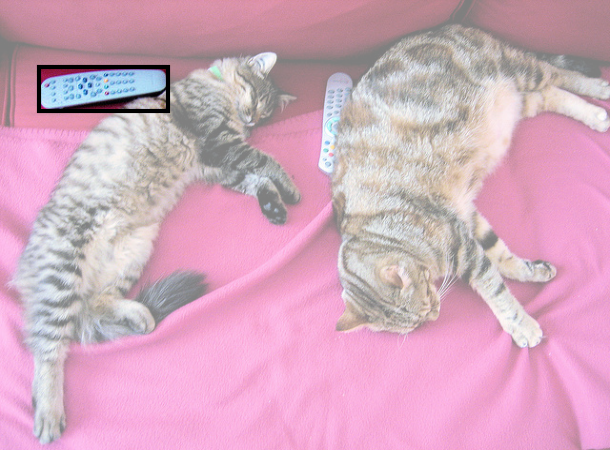} &
    \includegraphics[align=c,width=0.09\linewidth]{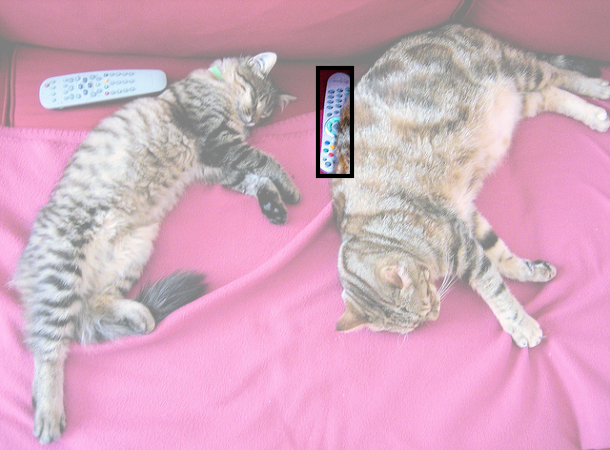} &
    \includegraphics[align=c,width=0.09\linewidth]{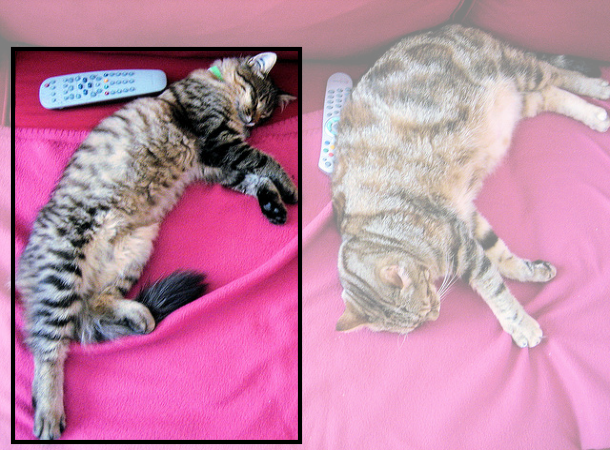} &
    \includegraphics[align=c,width=0.09\linewidth]{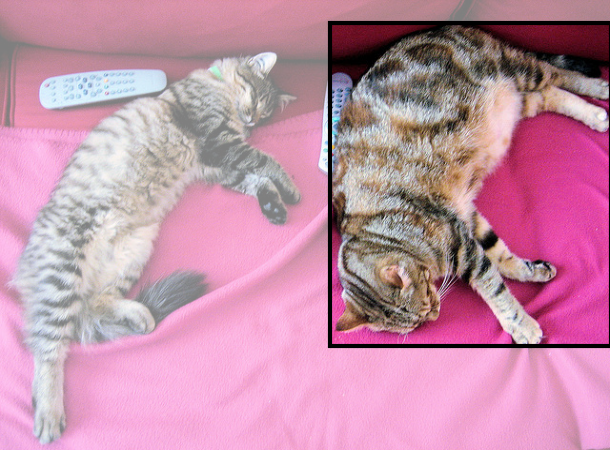} &
    \includegraphics[align=c,width=0.09\linewidth]{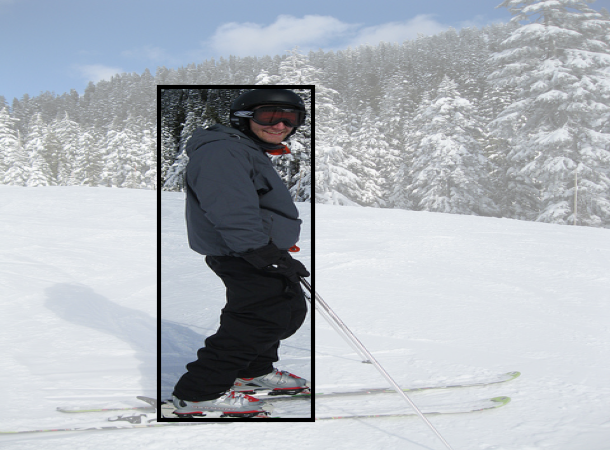} &
    \includegraphics[align=c,width=0.09\linewidth]{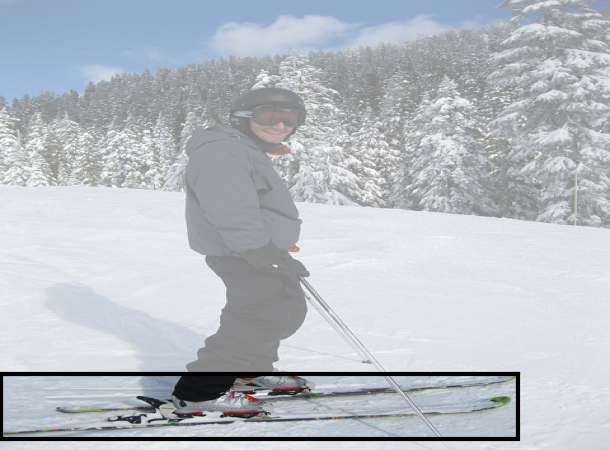} &
    \includegraphics[align=c,width=0.09\linewidth]{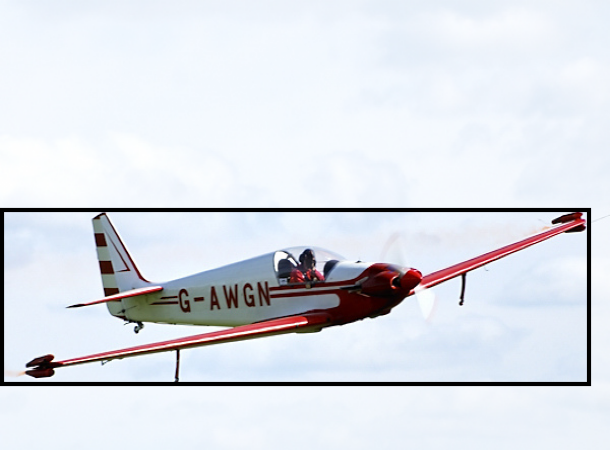} &
    \includegraphics[align=c,width=0.09\linewidth]{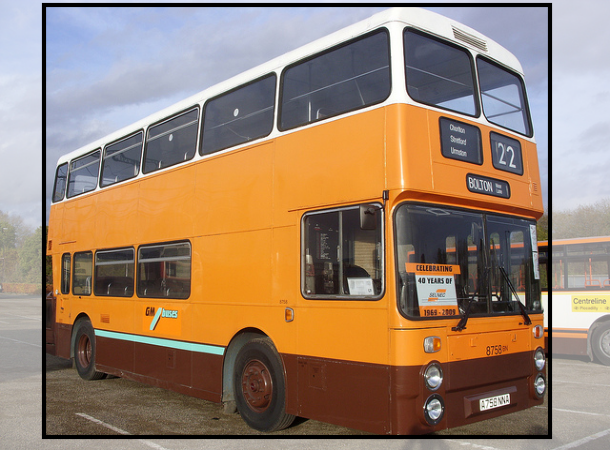} &
    \includegraphics[align=c,width=0.09\linewidth]{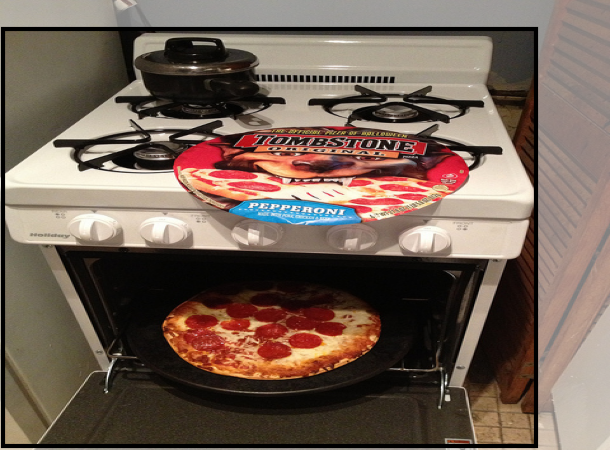} &\\
    Ours &
    \includegraphics[align=c,width=0.09\linewidth]{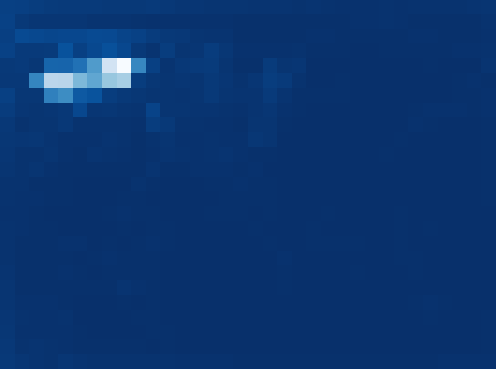} &
    \includegraphics[align=c,width=0.09\linewidth]{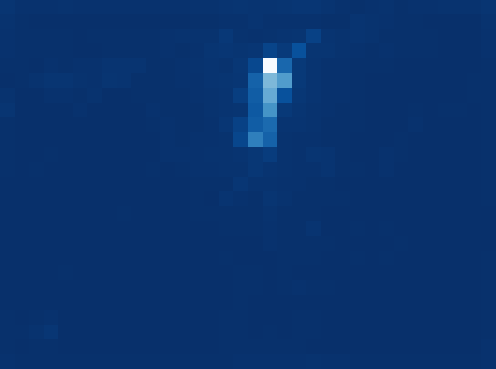} &
    \includegraphics[align=c,width=0.09\linewidth]{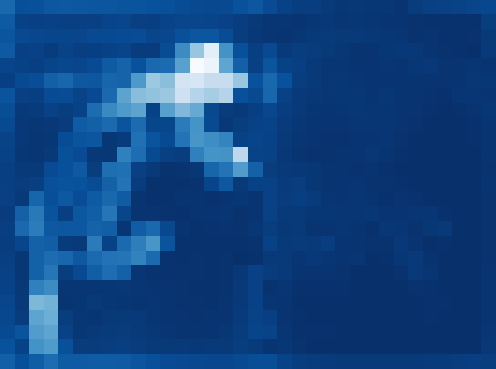} &
    \includegraphics[align=c,width=0.09\linewidth]{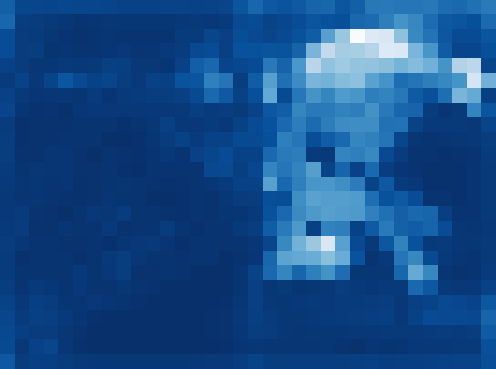} &
    \includegraphics[align=c,width=0.09\linewidth]{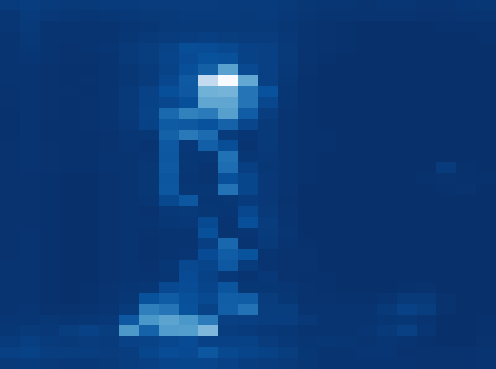} &
    \includegraphics[align=c,width=0.09\linewidth]{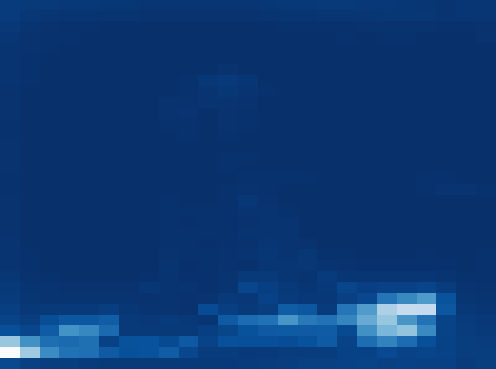} &
    \includegraphics[align=c,width=0.09\linewidth]{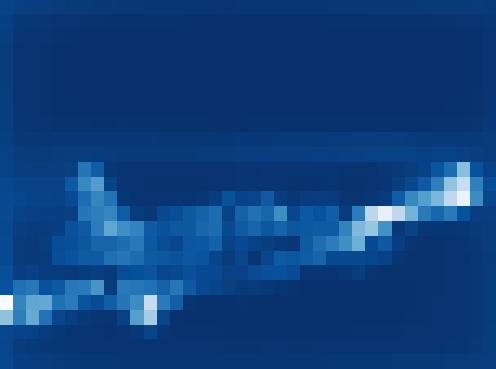}&
    \includegraphics[align=c,width=0.09\linewidth]{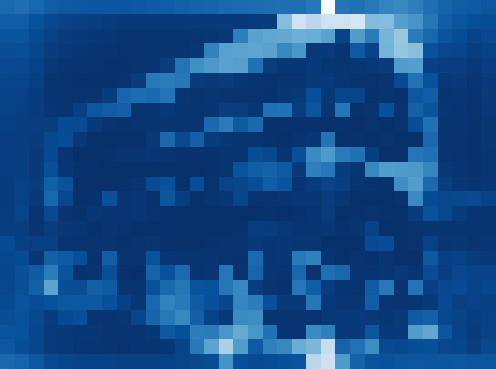} &
    \includegraphics[align=c,width=0.09\linewidth]{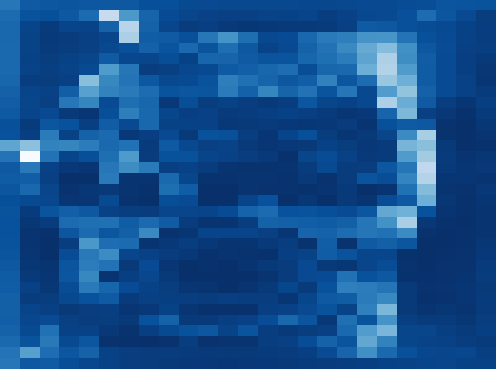} &\\
    Trans. att.~\cite{chefer2020transformer} &
    \includegraphics[align=c,width=0.09\linewidth]{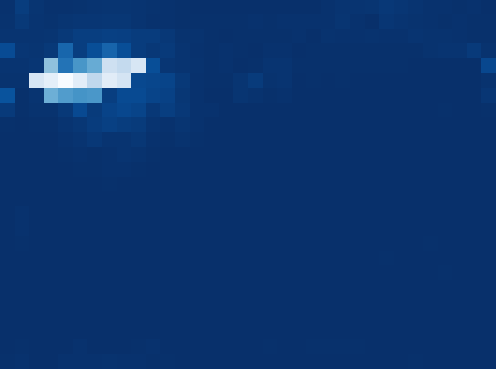} &
    \includegraphics[align=c,width=0.09\linewidth]{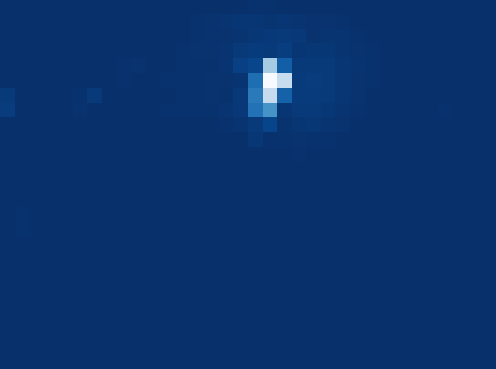} &
    \includegraphics[align=c,width=0.09\linewidth]{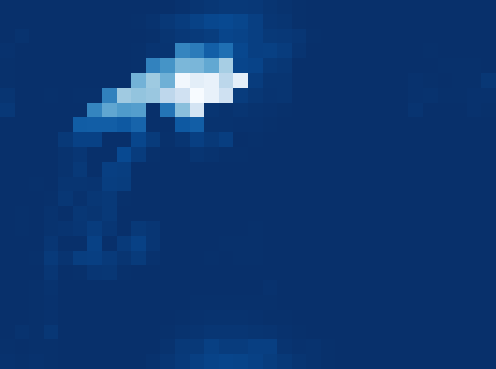} &
    \includegraphics[align=c,width=0.09\linewidth]{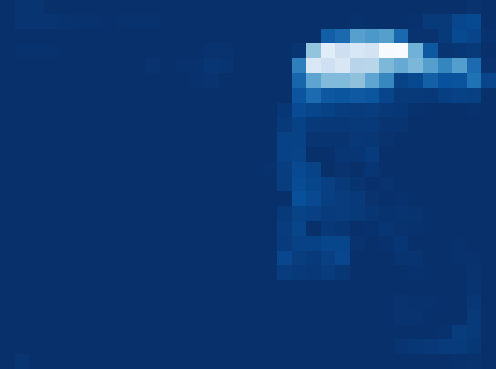} &
    \includegraphics[align=c,width=0.09\linewidth]{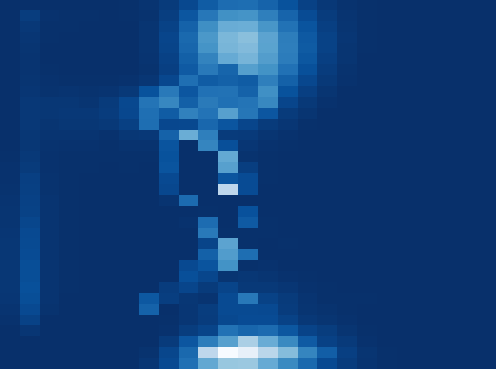} &
    \includegraphics[align=c,width=0.09\linewidth]{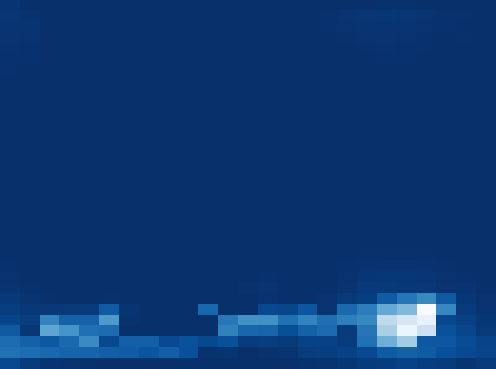} &
    \includegraphics[align=c,width=0.09\linewidth]{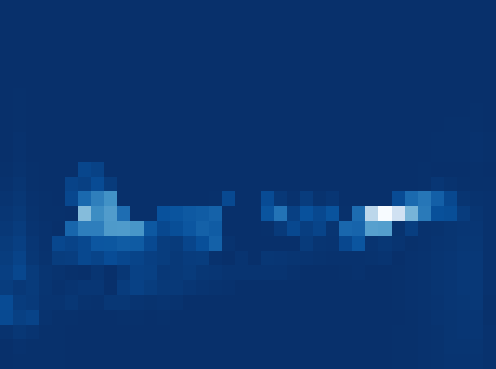}&
    \includegraphics[align=c,width=0.09\linewidth]{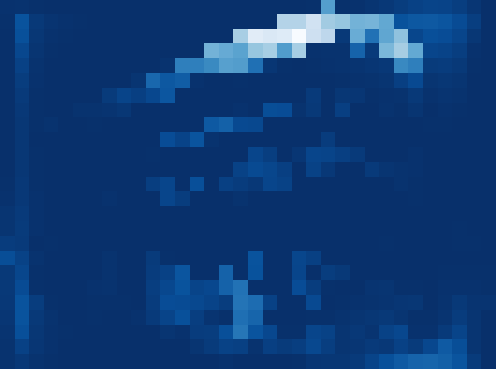} &
    \includegraphics[align=c,width=0.09\linewidth]{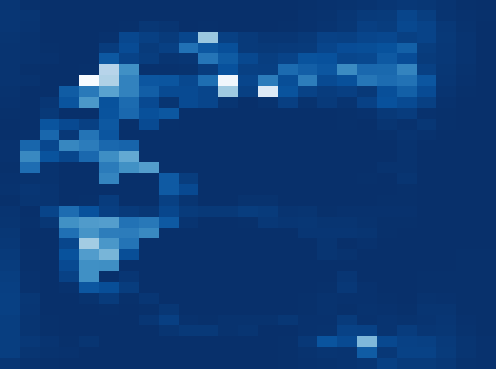} &\\
    partial LRP~\cite{voita2019analyzing} &
    \includegraphics[align=c,width=0.09\linewidth]{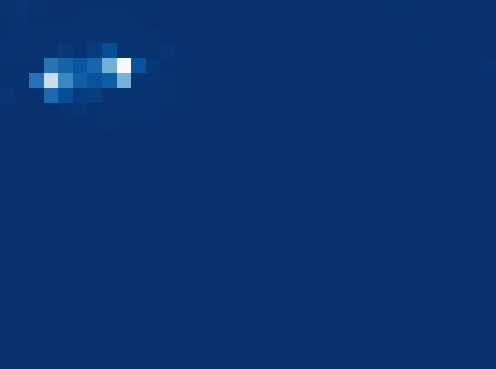} &
    \includegraphics[align=c,width=0.09\linewidth]{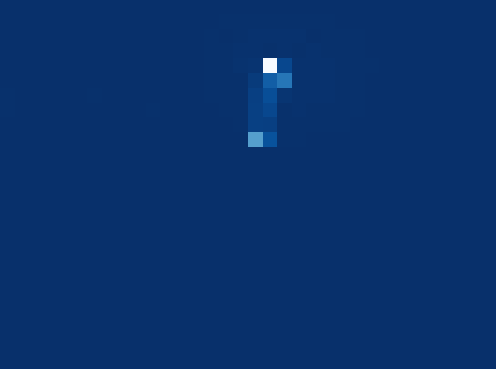} &
    \includegraphics[align=c,width=0.09\linewidth]{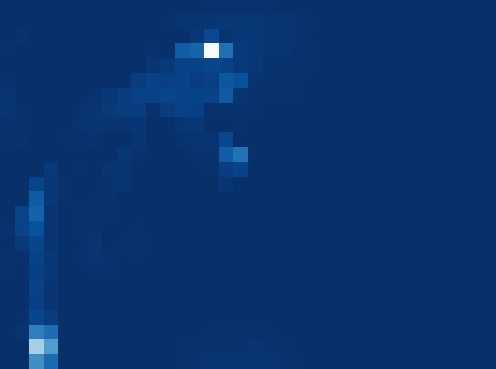} &
    \includegraphics[align=c,width=0.09\linewidth]{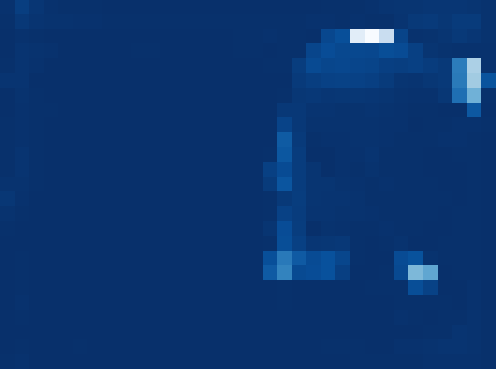} &
    \includegraphics[align=c,width=0.09\linewidth]{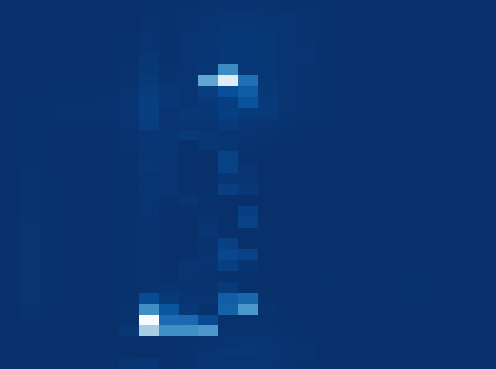} &
    \includegraphics[align=c,width=0.09\linewidth]{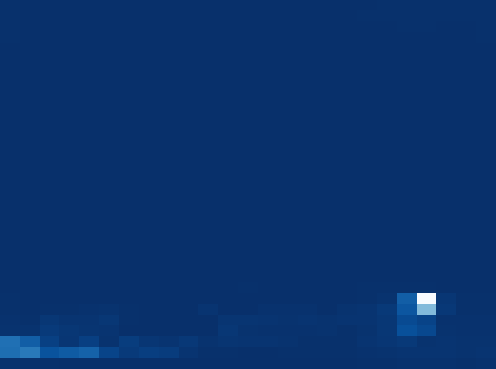} &
    \includegraphics[align=c,width=0.09\linewidth]{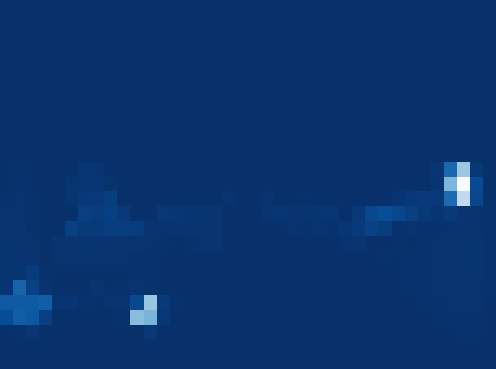}&
    \includegraphics[align=c,width=0.09\linewidth]{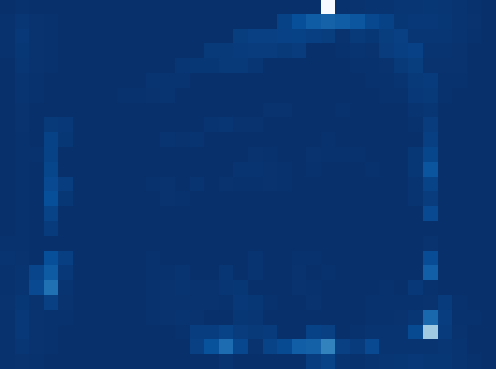} &
    \includegraphics[align=c,width=0.09\linewidth]{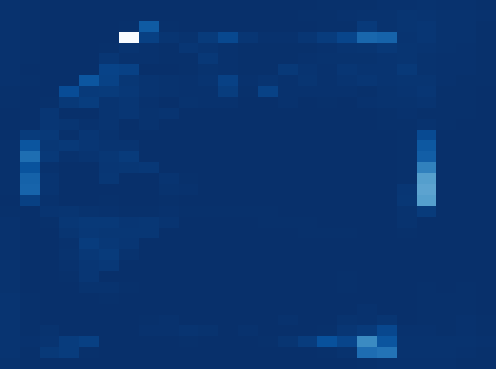} &\\
    Grad-CAM~\cite{selvaraju2017grad} &
    \includegraphics[align=c,width=0.09\linewidth]{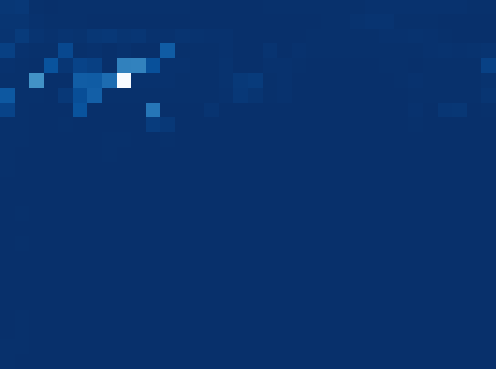} &
    \includegraphics[align=c,width=0.09\linewidth]{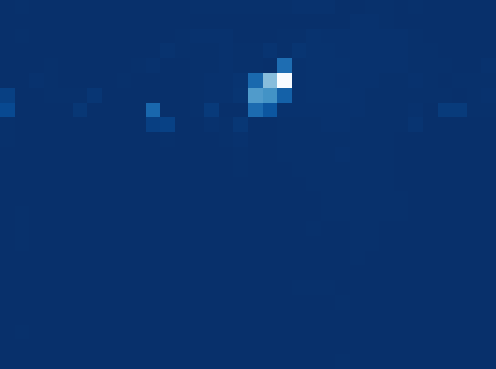} &
    \includegraphics[align=c,width=0.09\linewidth]{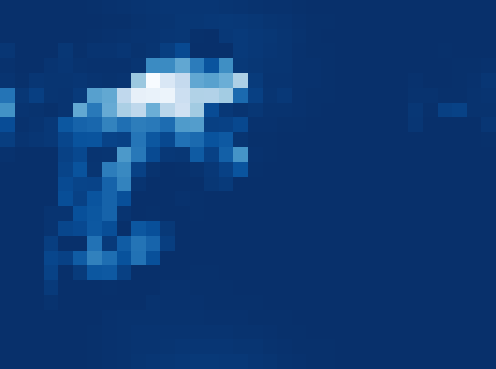} &
    \includegraphics[align=c,width=0.09\linewidth]{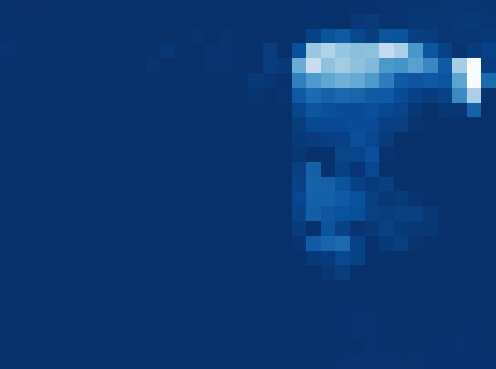} &
    \includegraphics[align=c,width=0.09\linewidth]{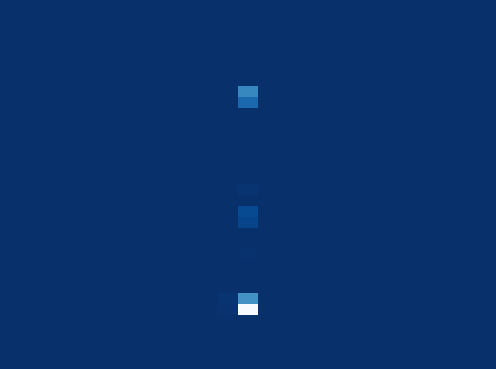} &
    \includegraphics[align=c,width=0.09\linewidth]{figures/DETR_vis/airplain_gradcam_71.png} &
    \includegraphics[align=c,width=0.09\linewidth]{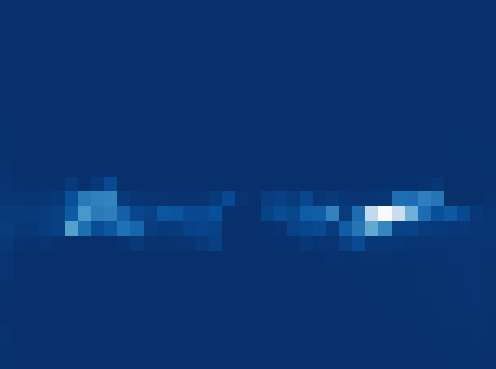}&
    \includegraphics[align=c,width=0.09\linewidth]{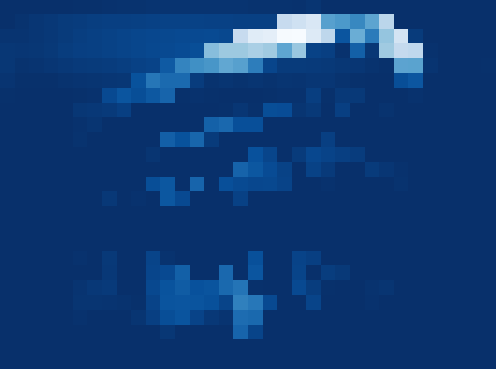} &
    \includegraphics[align=c,width=0.09\linewidth]{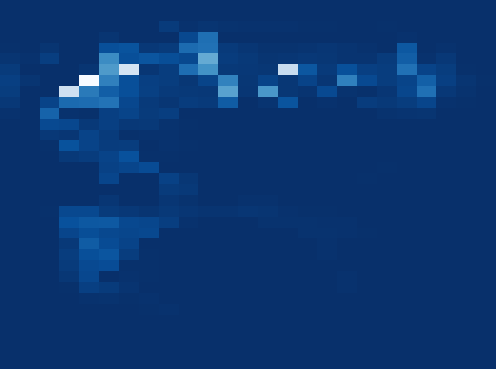} &\\
    raw att. &
    \includegraphics[align=c,width=0.09\linewidth]{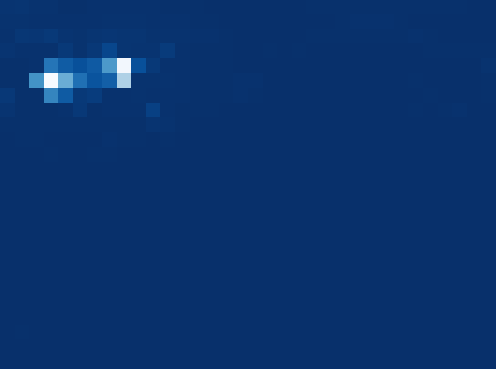} &
    \includegraphics[align=c,width=0.09\linewidth]{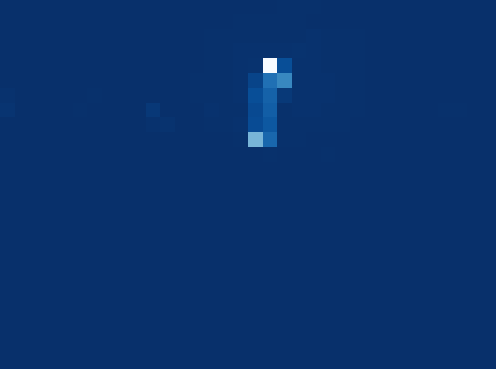} &
    \includegraphics[align=c,width=0.09\linewidth]{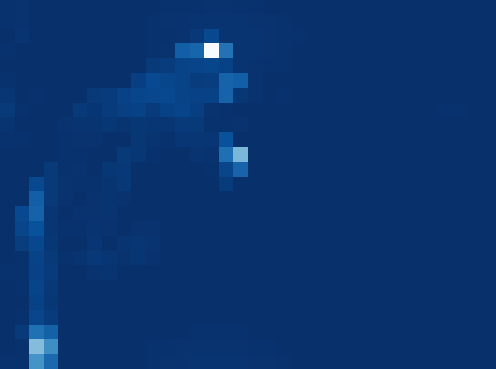} &
    \includegraphics[align=c,width=0.09\linewidth]{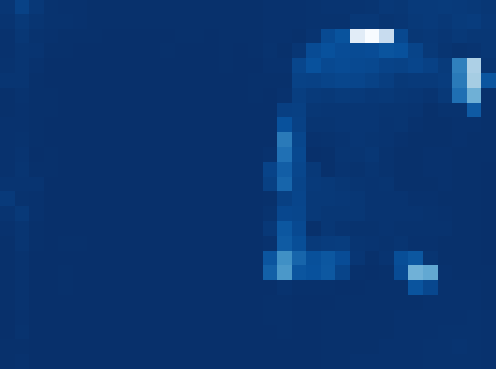} &
    \includegraphics[align=c,width=0.09\linewidth]{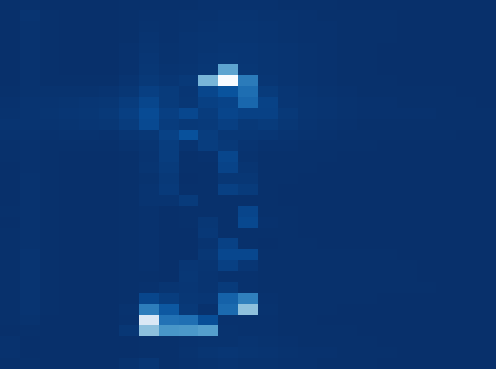} &
    \includegraphics[align=c,width=0.09\linewidth]{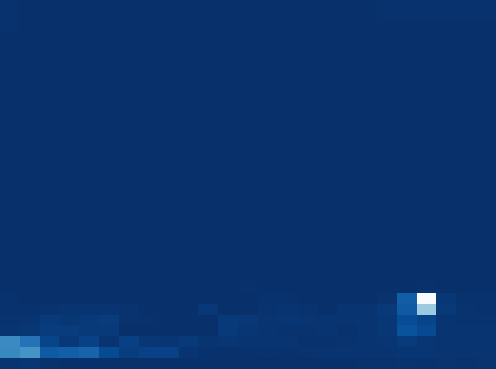} &
    \includegraphics[align=c,width=0.09\linewidth]{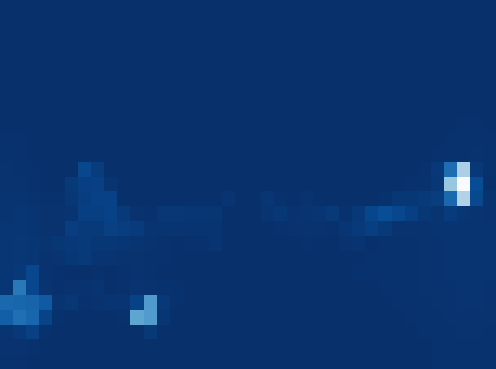}&
    \includegraphics[align=c,width=0.09\linewidth]{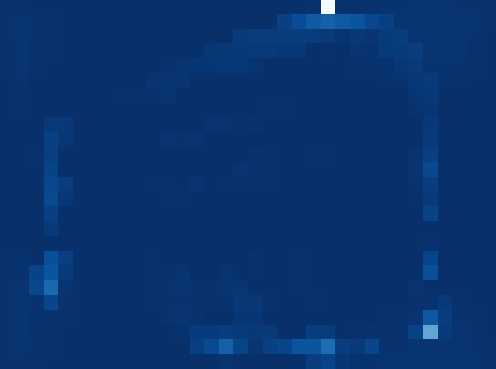} &
    \includegraphics[align=c,width=0.09\linewidth]{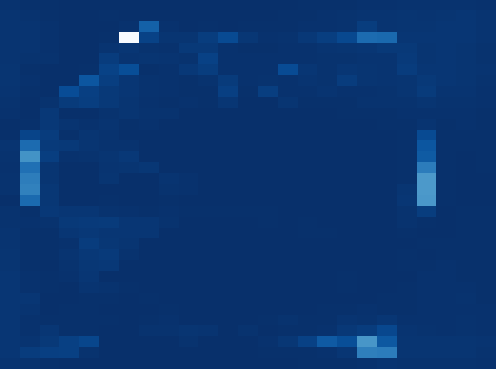} &\\
    rollout~\cite{abnar2020quantifying} &
    \includegraphics[align=c,width=0.09\linewidth]{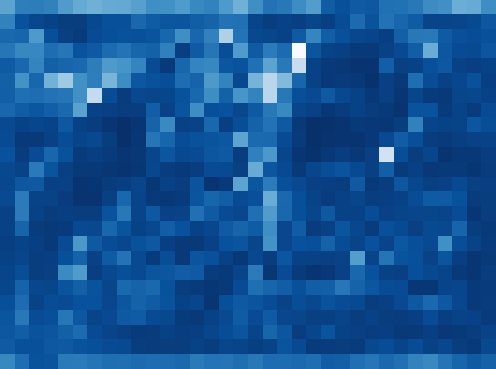} &
    \includegraphics[align=c,width=0.09\linewidth]{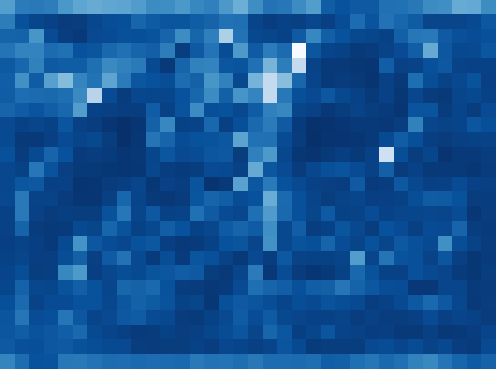} &
    \includegraphics[align=c,width=0.09\linewidth]{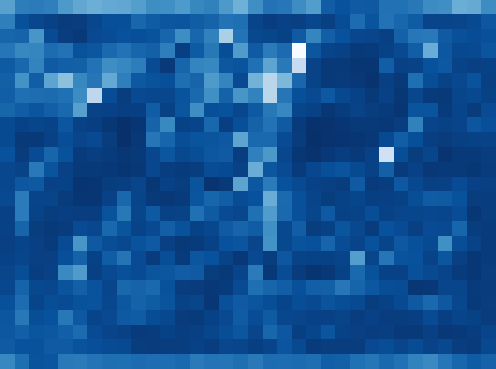} &
    \includegraphics[align=c,width=0.09\linewidth]{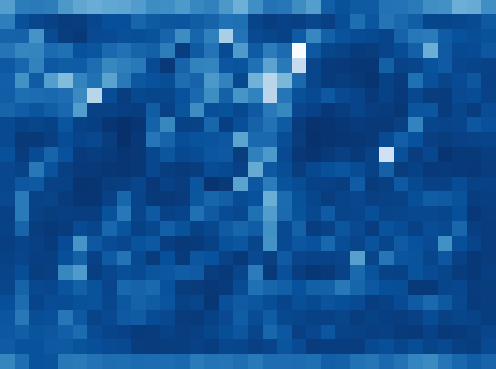} &
    \includegraphics[align=c,width=0.09\linewidth]{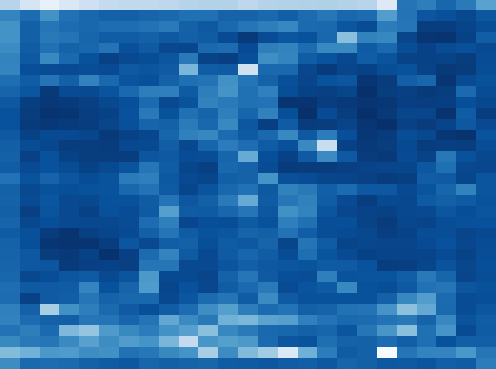} &
    \includegraphics[align=c,width=0.09\linewidth]{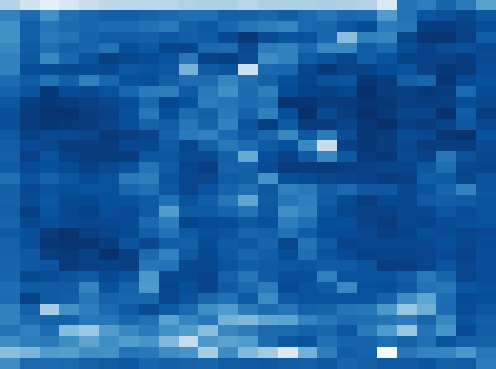} &
    \includegraphics[align=c,width=0.09\linewidth]{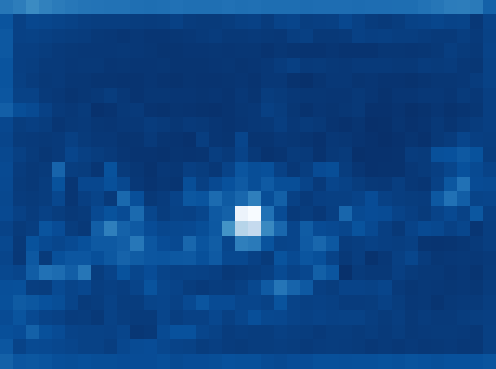}&
    \includegraphics[align=c,width=0.09\linewidth]{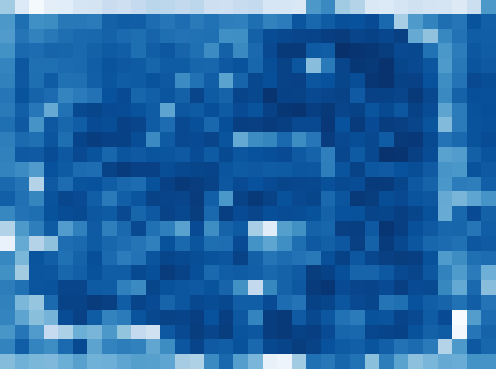} &
    \includegraphics[align=c,width=0.09\linewidth]{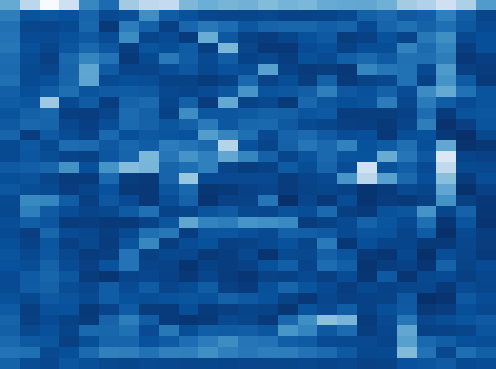} &\\
    \end{tabular*}
    \caption{Sample segmentation masks for DETR~\cite{zhu2021deformable}. Each row represents a method. Detections (from left to right): remote, remote, cat, cat, person, skis, airplane, bus, oven (in the last two samples, the bounding box is almost the entire frame). Our method produces the most accurate results, and the segmentations are consistent with the detections produced by DETR.} 
    \label{fig:DETR-vis}
    \end{center}
    \vspace{-10px}
\end{figure*}

\section{Experiments}
Our experiments include three Transformer-based models, each representing one of the three types of architectures we refer to in this work. See Fig.~\ref{fig:threetypes} for illustrations of each of the architectures. In addition, to compare with previous work~\cite{chefer2020transformer, abnar2020quantifying} in the same setting for which these methods were conceived, we also consider ViT~\cite{dosovitskiy2020image}. The relevancy propagation for each model follows Sec.~\ref{sec:adaptation}.  

The first model we examine is VisualBERT~\cite{li2019visualbert}, which represents a self-attention based architecture, and the second model is LXMERT~\cite{tan2019lxmert}, which represents an architecture combining self-attention and co-attention in a Transformer encoder for two modalities.

For both models, we perform positive and negative perturbation tests on each modality separately to evaluate the quality of the relevancy matrices produced by the methods. We use the visual question answering~\cite{Stanislaw2015vqa} task in testing the explanations since this task requires the models to demonstrate an understanding of both input modalities and the connections between them.

The perturbation tests are performed as follows: first, a pre-trained network is used for extracting relevancy maps for $10,000$ randomly picked samples from the validation set of the VQA dataset. Second, we gradually remove the tokens of a given modality and measure the mean top-1 accuracy of the network. In positive perturbation, tokens are removed from the highest relevance to the lowest, while in the negative version, from lowest to highest. In positive perturbation, one expects to see a steep decrease in performance, which indicates that the removed tokens are important to the classification score. In negative perturbation, a good explanation would maintain the accuracy of the model while removing tokens that are not related to the classification. In both cases, we measure the area-under-the-curve (AUC), to evaluate the decrease in the model's accuracy. 

We note that in all perturbation tests, the accuracy does not reach $0\%$, even when removing $100\%$ of the tokens of each modality. This is since the input from the other modality remains intact therefore the models can rely on a single modality to provide a reasonable answer.

Notice that the LXMERT~\cite{tan2019lxmert} image perturbation test results, which are depicted in Fig.~\ref{fig:LXMERT-pert}(a,b), demonstrate a clear advantage to our method compared to other methods. For negative perturbation, the AUC using our method is the largest by a sizeable margin, and the accuracy is well-preserved even after removing more than $80\%$ of the image tokens, and for positive perturbation, notice the very steep decrease in accuracy, and the low AUC. 

As can be seen in Fig.~\ref{fig:threetypes}(b), the \texttt{[CLS]} token for LXMERT~\cite{tan2019lxmert} is the first token of the text modality, thus following Sec.~\ref{sec:adaptation}, $\mathbf{R^{ti}}$ is the map used for extracting relevancies in the image perturbation case. Since $\mathbf{R^{ti}}$ is a multi-modal relevancy map, the image perturbation tests best demonstrate the advantage of using our method over all existing methods, which fall short in evaluating relevancies from the co-attention modules. 

For the LXMERT~\cite{tan2019lxmert} text perturbation tests which are depicted in Fig.~\ref{fig:LXMERT-pert}(c,d), notice that by Sec.~\ref{sec:adaptation}, we visualize $\mathbf{R^{tt}}$ which is a self-attention map, where the dominating update rule is Eq.~\ref{eq:self-attention-ss}. This rule is identical to the rule employed by the Transformer attribution \cite{chefer2020transformer} baseline, except for the head averaging in Eq.~\ref{eq:modifie_att_lrp}. Therefore, the main difference between our proposed method and the method described in~\cite{chefer2020transformer} is the choice to use LRP~\cite{bach2015pixel} in the head averaging process. This results in very similar results for both methods. 
For completeness, we provide in the supplementary results for our method when adding LRP, as is done in Eq.~\ref{eq:modifie_att_lrp}. The rest of the methods fall far behind.

Fig.~\ref{fig:LXMERT-vis} presents typical results for our method and for Transformer attribution~\cite{chefer2020transformer}. The rest of the methods are not competitive and their matching samples are presented in the supplementary. As can be seen, the text results are similar, as predicted by the quantitative results. Our image attention results are much more focused on the relevant image parts than those of the baseline method. 


Note that since VisualBERT~\cite{li2019visualbert} is based on pure self-attention, the difference between our method and the Transformer attribution~\cite{chefer2020transformer} method stems from the choice of whether or not to use LRP~\cite{bach2015pixel} for head averaging in Eq.~\ref{eq:modified_att}, similarly to the LXMERT~\cite{tan2019lxmert} text (but not image) perturbation tests. As can be seen in Fig.~\ref{fig:VisualBERT-pert}, our method outperforms all methods and achieves very similar results to those of~\cite{chefer2020transformer}, and in some cases, such as the text perturbation test, even outperforms~\cite{chefer2020transformer} by a sizeable margin. This demonstrates that the use of LRP~\cite{bach2015pixel} is unnecessary, 
even for pure self-attention architectures.

The third model we experiment on is DETR~\cite{carion2020end}, which is an encoder-decoder model, as seen in Fig.~\ref{fig:threetypes}(c). We use a pre-trained DETR model with the ImageNet pre-trained backbone ResNet-50, which is trained for object detection on the MSCOCO~\cite{ty2014coco} dataset. Importantly, this model has only been trained for object detection, \ie, producing bounding boxes and classifications for each object in the input image. To evaluate the different explainability methods, 
our test uses each of the methods on the $5,000$ samples of the MSCOCO~\cite{ty2014coco} validation set to produce segmentation masks, \ie we consider the output of each method to be a segmentation mask. We first filter the queries to include only ones where the classification probability is higher than $50\%$ and then employ Otsu's thresholding method~\cite{otsu1979threshold} to separate the foreground and the background of the segmentation. See supplementary for the full details.

Our generated segmentation masks visualize the bounding boxes predicted by DETR, therefore it should be noted that the produced masks are inherently dependent on the quality of the corresponding bounding boxes, \ie, when the predicted bounding box is not sufficient, naturally, the mask produced for it will be at least equally inaccurate. In addition, since the explainability methods are not aimed at producing segmentation maps, they often do not output contiguous masks, and the Otsu threshold may also create "holes" in the produced masks. For all the reasons above, we decrease the minimal IoU used for MSCOCO evaluation from $0.5$ to $0.2$, which significantly benefits all the methods, and we present the results of the MSCOCO segmentation evaluation for the categories where the produced bounding boxes are good enough for the generation of segmentation masks, \eg, we do not present results for small objects\footnote{We choose this working point since using a stricter threshold leads to baseline results that are slightly better than chance and our method outperforms but provides a score that is only 2-3 times better than chance.}. 
As can be seen in Tab.~\ref{tab:DETR}, our method outperforms all other methods by a very large margin, which indicates that our novel formulations are necessary for non self-attention architectures. Notice the correlation in Tab.~\ref{tab:DETR} between the bounding box evaluation for DETR and our segmentation. See Fig.~\ref{fig:DETR-vis}
for visualizations of the masks.

Lastly, in order to compare our method with existing single-modality baselines, we present the positive and negative perturbation tests on ViT-Base~\cite{dosovitskiy2020image}, as performed by~\cite{chefer2020transformer}. As mentioned, since ViT-Base~\cite{dosovitskiy2020image} is a single-modality Transformer encoder, the only difference between our method and the Transformer attribution method of~\cite{chefer2020transformer} is the use of LRP~\cite{bach2015pixel} in Eq.~\ref{eq:modified_att}, as shown in Eq.~\ref{eq:modifie_att_lrp}. Therefore, as can be seen in Tab.~\ref{tab:ViT}, the differences between our method and the method proposed in~\cite{chefer2020transformer} are very mild, which is another indication that LRP~\cite{bach2015pixel} can be removed. 
Tab.~\ref{tab:ViT} also shows improvement in performance when using the target class instead of the predicted class for gradient propagation in Eq.~\ref{eq:modified_att}, which, as stated in~\cite{chefer2020transformer}, indicates that our method is able to produce class-specific visualizations. 

\noindent{\bf Ablation study\quad} We present in the supplementary three variations of our method that demonstrate the effectiveness of our normalization (Eq.~\ref{eq:norm-sum},\ref{eq:norm}), the necessity of the aggregation in all our rules~\ref{eq:self-attention-ss},~\ref{eq:self-attention-sq},~\ref{eq:bi-attention-sq},~\ref{eq:bi-attention-ss}, and the need for the self-attention updates to the bi-modal rule~\ref{eq:bi-attention-sq}.

\section{Conclusions}

Transformers play an increasingly dominant role in computer vision, with image-text Transformers and Transformers that perform tasks that have output domains that are more complex than the labels provided by a classifier, presenting groundbreaking results. In order to debug such models, as well as to support downstream tasks, and the increasing demand for model-interpretability, it is required to have complete and accurate explainability methods. However, the current explainability literature for Transformers is limited, overly focuses on pure attention maps, and lacks the methodology for treating co-attention maps.

Our method carefully tracks the evolution and mixing of the attention maps. It provides a generic prescription that is applicable to all attention models we are aware of. Empirically, it  outperforms the existing methods across Transformer architectures and evaluation metrics. In some cases, when self-attention is prominent, the recent method by Chefer et al.~\cite{chefer2020transformer} is the only method that can provide comparable results. However, in the majority of the experiments, our method leads over all methods by a very sizable margin.

\section*{Acknowledgment}
This project has received funding from the European Research Council (ERC) under the European Unions Horizon 2020 research and innovation programme (grant ERC CoG 725974). The contribution of the first author is part of a Master thesis research conducted at Tel Aviv University.

{\small
\bibliographystyle{ieee_fullname}
\bibliography{explainability}
}

\clearpage
\onecolumn
\appendix
\setcounter{equation}{0}
\setcounter{figure}{0}

\include{supplementary}

\end{document}

%% file: supplementary.tex












\section{Code}
The code contains Jupyter notebooks with the examples presented for LXMERT and DETR. Both notebooks also allow using images from the internet. For LXMERT, we also support the option of asking a free form question.

\begin{figure}[h]
    \begin{center}
    \centering
    \includegraphics[width=0.9\linewidth]{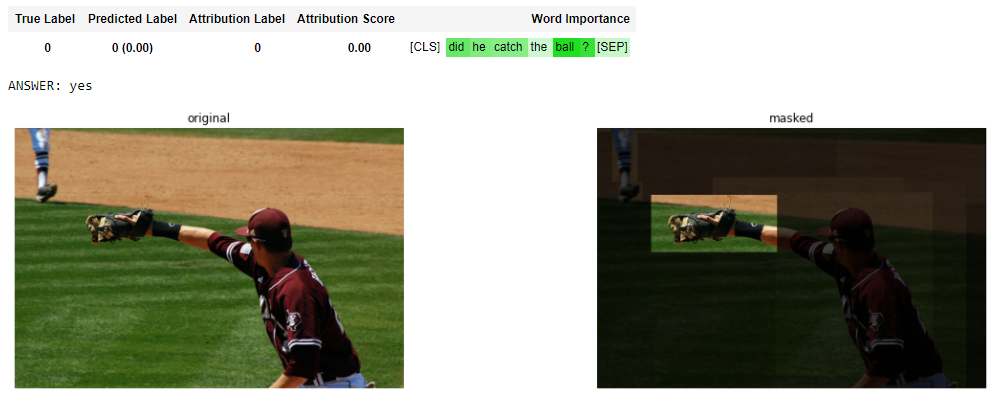}
    \includegraphics[width=0.9\linewidth]{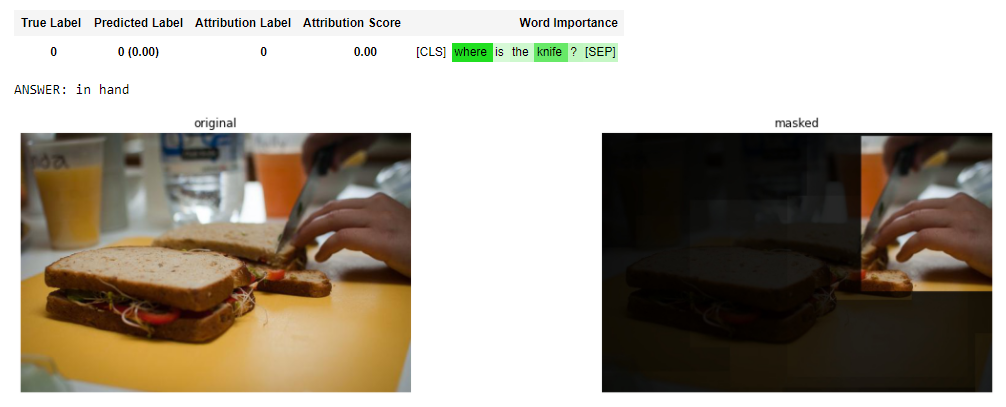}
    \end{center}
    \caption{LXMERT examples from the Jupyter notebook. The notebook contains both the examples from the paper (top), and examples of uploaded images and free form questions (bottom).}
    \label{fig:LXMERT}
\end{figure}

\begin{figure}[H]
    \begin{center}
    \centering
    \includegraphics[width=0.9\linewidth]{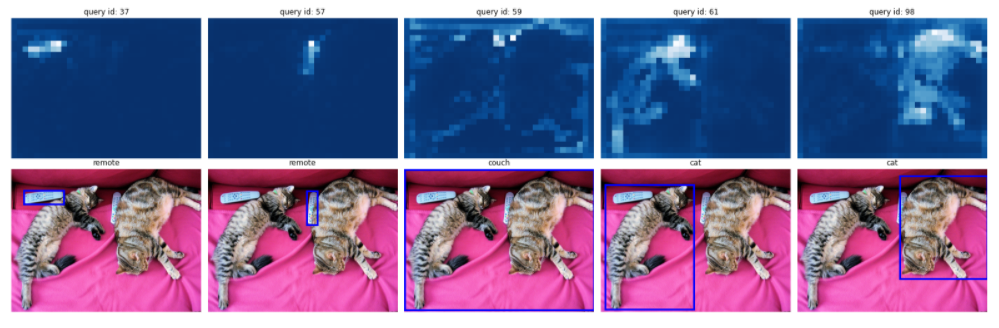}
    \end{center}
    \caption{DETR example from the Jupyter notebook. The notebook contains the examples from the paper.}
    \label{fig:DETR}
\end{figure}

\section{Extended LXMERT VQA visual results}
\begin{figure*}
    \setlength{\tabcolsep}{1pt} 
    \renewcommand{\arraystretch}{1} 
    \begin{tabular*}{\linewidth}{@{\extracolsep{\fill}}lcccc}
     &
    \begin{CJK*}{UTF8}{gbsn}
{\setlength{\fboxsep}{0pt}\colorbox{white!0}{\parbox{0.2\textwidth}{    \colorbox{red!58.1370735168457}{\strut is} \colorbox{red!0.0}{\strut the} \colorbox{red!0.0}{\strut animal} \colorbox{red!65.44914245605469}{\strut eating} \colorbox{red!100.0}{\strut ?} \colorbox{red!1.2035624980926514}
}}}
\end{CJK*}&
    \begin{CJK*}{UTF8}{gbsn}
{\setlength{\fboxsep}{0pt}\colorbox{white!0}{\parbox{0.2\textwidth}{   \colorbox{red!59.360687255859375}{\strut did} \colorbox{red!47.44766616821289}{\strut he} \colorbox{red!41.36971664428711}{\strut catch} \colorbox{red!0.0}{\strut the} \colorbox{red!100.0}{\strut ball} \colorbox{red!87.9761734008789}{\strut ?}
}}}
\end{CJK*} &
    \begin{CJK*}{UTF8}{gbsn}
{\setlength{\fboxsep}{0pt}\colorbox{white!0}{\parbox{0.2\textwidth}{ \colorbox{red!4.920843124389648}{\strut is} \colorbox{red!0.0}{\strut the} \colorbox{red!100.00000762939453}{\strut tub} \colorbox{red!40.00966262817383}{\strut white} \colorbox{red!42.100624084472656}{\strut ?}
}}}
\end{CJK*} &
    \begin{CJK*}{UTF8}{gbsn}
{\setlength{\fboxsep}{0pt}\colorbox{white!0}{\parbox{0.2\textwidth}{   \colorbox{red!30.391498565673828}{\strut did} \colorbox{red!5.028392791748047}{\strut the} \colorbox{red!26.076675415039062}{\strut man} \colorbox{red!35.311214447021484}{\strut just} \colorbox{red!100.0}{\strut catch} \colorbox{red!16.184215545654297}{\strut the} \colorbox{red!5.848325729370117}{\strut fr}\colorbox{red!0.0}{\strut is}\colorbox{red!20.73851203918457}{\strut bee} \colorbox{red!94.43649291992188}{\strut ?} 
}}}
\end{CJK*}\\
    \rotatebox[origin=c]{90}{~~~~~Ours} &
    \includegraphics[align=c,height=0.165\linewidth]{figures/LXMERT_vis/185590/COCO_val2014_000000185590_ours.jpg} &
    \includegraphics[align=c,height=0.165\linewidth]{figures/LXMERT_vis/127510/COCO_val2014_000000127510_ours.jpg} &
    \includegraphics[align=c,height=0.165\linewidth]{figures/LXMERT_vis/324266/COCO_val2014_000000324266_ours.jpg} &
    \includegraphics[align=c,height=0.165\linewidth]{figures/LXMERT_vis/200717/COCO_val2014_000000200717_ours.jpg} 
    \\\\
    &
    

\begin{CJK*}{UTF8}{gbsn}
{\setlength{\fboxsep}{0pt}\colorbox{white!0}{\parbox{0.2\textwidth}{   \colorbox{red!54.17789077758789}{\strut is} \colorbox{red!0.0}{\strut the} \colorbox{red!3.314396858215332}{\strut animal} \colorbox{red!59.200435638427734}{\strut eating} \colorbox{red!100.0}{\strut ?}
}}}
\end{CJK*} &
    \begin{CJK*}{UTF8}{gbsn}
{\setlength{\fboxsep}{0pt}\colorbox{white!0}{\parbox{0.2\textwidth}{  \colorbox{red!38.96585464477539}{\strut did} \colorbox{red!22.602096557617188}{\strut he} \colorbox{red!44.653099060058594}{\strut catch} \colorbox{red!6.535938262939453}{\strut the} \colorbox{red!100.0}{\strut ball} \colorbox{red!69.22191619873047}{\strut ?}
}}}
\end{CJK*} &
    \begin{CJK*}{UTF8}{gbsn}
{\setlength{\fboxsep}{0pt}\colorbox{white!0}{\parbox{0.2\textwidth}{ \colorbox{red!18.032072067260742}{\strut is} \colorbox{red!2.069951057434082}{\strut the} \colorbox{red!64.09111022949219}{\strut tub} \colorbox{red!99.99999237060547}{\strut white} \colorbox{red!53.09128189086914}{\strut ?}
}}}
\end{CJK*} &
    \begin{CJK*}{UTF8}{gbsn}
{\setlength{\fboxsep}{0pt}\colorbox{white!0}{\parbox{0.2\textwidth}{  \colorbox{red!21.42181968688965}{\strut did} \colorbox{red!0.0}{\strut the} \colorbox{red!1.4645568132400513}{\strut man} \colorbox{red!21.948606491088867}{\strut just} \colorbox{red!15.582681655883789}{\strut catch} \colorbox{red!0.0}{\strut the} \colorbox{red!2.117180347442627}{\strut fr}\colorbox{red!2.7323646545410156}{\strut is}\colorbox{red!2.486506700515747}{\strut bee} \colorbox{red!100.00000762939453}{\strut ?}
}}}
\end{CJK*}\\
\rotatebox[origin=c]{90}{~~~~~Trans. att.} &
    \includegraphics[align=c,height=0.165\linewidth]{figures/LXMERT_vis/185590/COCO_val2014_000000185590_transformer_attr.jpg} &
    \includegraphics[align=c,height=0.165\linewidth]{figures/LXMERT_vis/127510/COCO_val2014_000000127510_transformer_attr.jpg} &
    \includegraphics[align=c,height=0.165\linewidth]{figures/LXMERT_vis/324266/COCO_val2014_000000324266_transformer_attr.jpg} &
    \includegraphics[align=c,height=0.165\linewidth]{figures/LXMERT_vis/200717/COCO_val2014_000000200717_transformer_attr.jpg} 
    \\
  
    
    &
    \begin{CJK*}{UTF8}{gbsn}
{\setlength{\fboxsep}{0pt}\colorbox{white!0}{\parbox{0.2\textwidth}{   \colorbox{red!71.55523681640625}{\strut is} \colorbox{red!0.0}{\strut the} \colorbox{red!12.155278205871582}{\strut animal} \colorbox{red!13.623899459838867}{\strut eating} \colorbox{red!100.0}{\strut ?}
}}}
\end{CJK*} &
    \begin{CJK*}{UTF8}{gbsn}
{\setlength{\fboxsep}{0pt}\colorbox{white!0}{\parbox{0.2\textwidth}{  \colorbox{red!7.101400375366211}{\strut did} \colorbox{red!0.0}{\strut he} \colorbox{red!4.807400703430176}{\strut catch} \colorbox{red!0.0}{\strut the} \colorbox{red!0.0}{\strut ball} \colorbox{red!100.0}{\strut ?}
}}}
\end{CJK*} &
    \begin{CJK*}{UTF8}{gbsn}
{\setlength{\fboxsep}{0pt}\colorbox{white!0}{\parbox{0.2\textwidth}{ \colorbox{red!10.491823196411133}{\strut is} \colorbox{red!0.0}{\strut the} \colorbox{red!31.27155876159668}{\strut tub} \colorbox{red!16.228574752807617}{\strut white} \colorbox{red!100.0}{\strut ?}
}}}
\end{CJK*} &
    \begin{CJK*}{UTF8}{gbsn}
{\setlength{\fboxsep}{0pt}\colorbox{white!0}{\parbox{0.2\textwidth}{  \colorbox{red!1.761559247970581}{\strut did} \colorbox{red!1.8291882276535034}{\strut the} \colorbox{red!0.0}{\strut man} \colorbox{red!18.06475257873535}{\strut just} \colorbox{red!9.239696502685547}{\strut catch} \colorbox{red!0.0}{\strut the} \colorbox{red!1.9935507774353027}{\strut fr}\colorbox{red!1.8459583520889282}{\strut is}\colorbox{red!1.858442783355713}{\strut bee} \colorbox{red!100.0}{\strut ?}
}}}
\end{CJK*}\\
\rotatebox[origin=c]{90}{~~~~~partial LRP} &
    \includegraphics[align=c,height=0.165\linewidth]{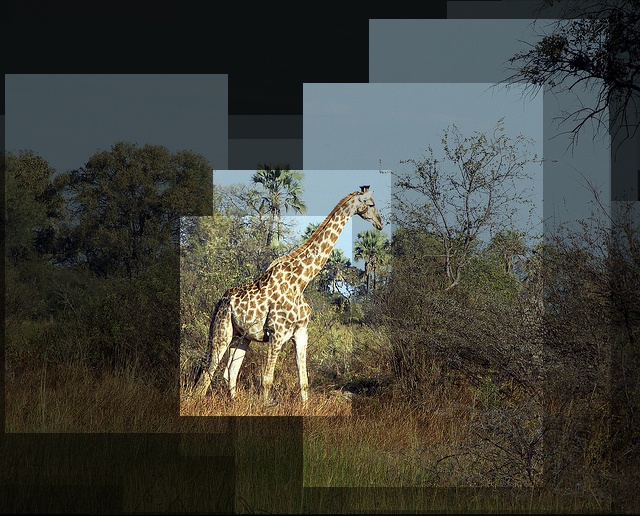} &
    \includegraphics[align=c,height=0.165\linewidth]{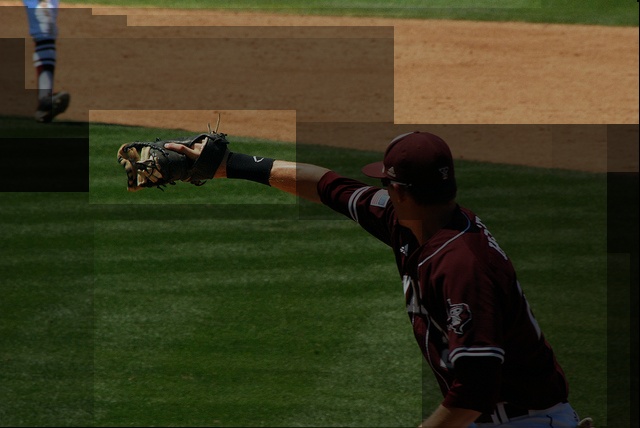} &
    \includegraphics[align=c,height=0.165\linewidth]{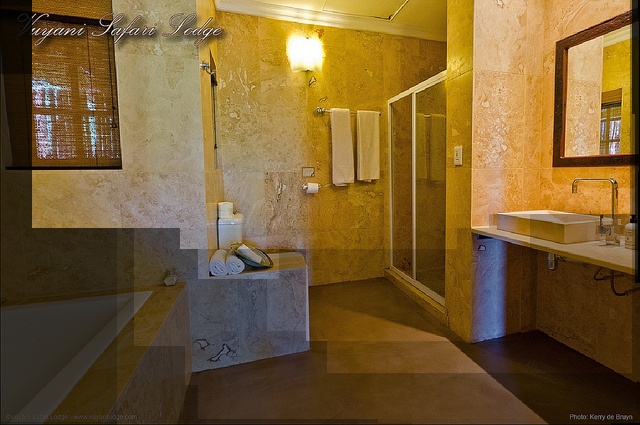} &
    \includegraphics[align=c,height=0.165\linewidth]{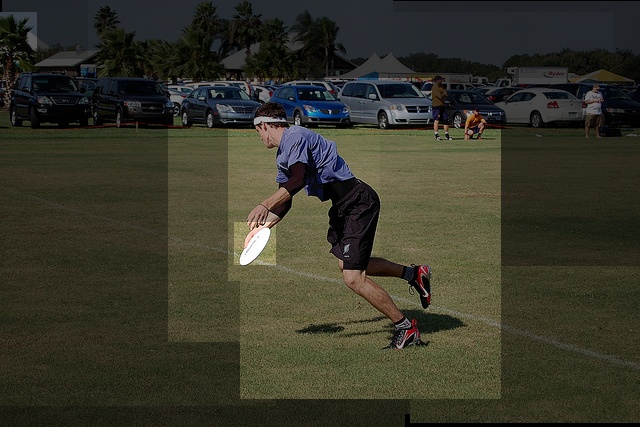} 
    \\

    
        &
    \begin{CJK*}{UTF8}{gbsn}
{\setlength{\fboxsep}{0pt}\colorbox{white!0}{\parbox{0.2\textwidth}{   \colorbox{red!86.70310974121094}{\strut is} \colorbox{red!62.29895782470703}{\strut the} \colorbox{red!6.177962779998779}{\strut animal} \colorbox{red!0.0}{\strut eating} \colorbox{red!100.0}{\strut ?}

}}}
\end{CJK*} &
    \begin{CJK*}{UTF8}{gbsn}
{\setlength{\fboxsep}{0pt}\colorbox{white!0}{\parbox{0.2\textwidth}{  \colorbox{red!11.449234962463379}{\strut did} \colorbox{red!0.0}{\strut he} \colorbox{red!3.343825578689575}{\strut catch} \colorbox{red!23.538013458251953}{\strut the} \colorbox{red!3.3261044025421143}{\strut ball} \colorbox{red!100.0}{\strut ?}

}}}
\end{CJK*} &
    \begin{CJK*}{UTF8}{gbsn}
{\setlength{\fboxsep}{0pt}\colorbox{white!0}{\parbox{0.2\textwidth}{ \colorbox{red!52.8731689453125}{\strut is} \colorbox{red!64.24579620361328}{\strut the} \colorbox{red!0.0}{\strut tub} \colorbox{red!41.77085494995117}{\strut white} \colorbox{red!100.0}{\strut ?}
}}}
\end{CJK*} &
    \begin{CJK*}{UTF8}{gbsn}
{\setlength{\fboxsep}{0pt}\colorbox{white!0}{\parbox{0.2\textwidth}{   \colorbox{red!2.2868552207946777}{\strut did} \colorbox{red!17.637758255004883}{\strut the} \colorbox{red!0.0}{\strut man} \colorbox{red!7.790342330932617}{\strut just} \colorbox{red!5.826866626739502}{\strut catch} \colorbox{red!12.080976486206055}{\strut the} \colorbox{red!0.0}{\strut fr}\colorbox{red!0.0}{\strut is}\colorbox{red!0.0}{\strut bee} \colorbox{red!100.0}{\strut ?} 
}}}
\end{CJK*}\\
\rotatebox[origin=c]{90}{~~~~~Grad-CAM} &
    \includegraphics[align=c,height=0.165\linewidth]{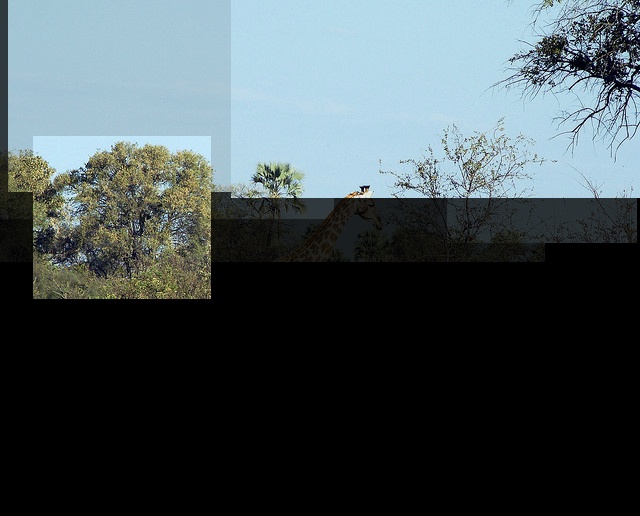} &
    \includegraphics[align=c,height=0.165\linewidth]{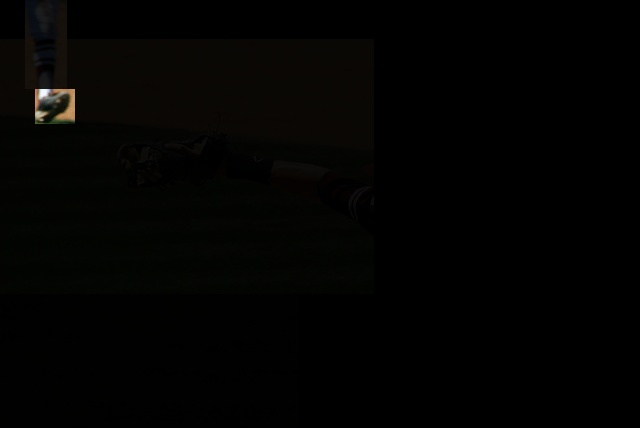} &
    \includegraphics[align=c,height=0.165\linewidth]{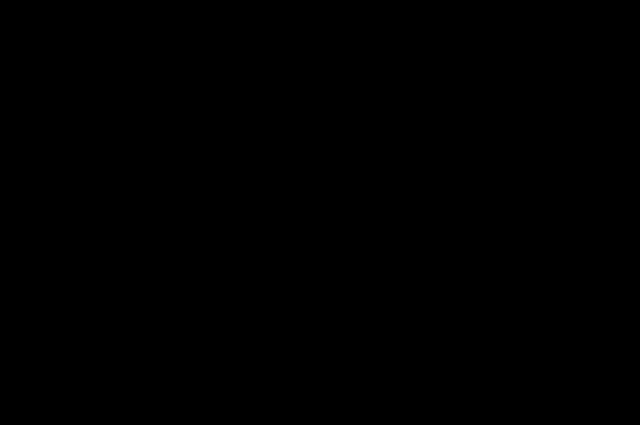} &
    \includegraphics[align=c,height=0.165\linewidth]{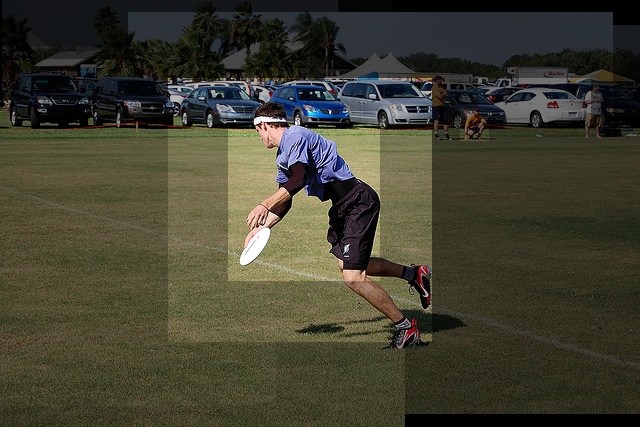} 
    \\


        &
    \begin{CJK*}{UTF8}{gbsn}
{\setlength{\fboxsep}{0pt}\colorbox{white!0}{\parbox{0.2\textwidth}{   \colorbox{red!78.64637756347656}{\strut is} \colorbox{red!65.93904113769531}{\strut the} \colorbox{red!0.0}{\strut animal} \colorbox{red!7.380090713500977}{\strut eating} \colorbox{red!100.00000762939453}{\strut ?}
}}}
\end{CJK*} &
    \begin{CJK*}{UTF8}{gbsn}
{\setlength{\fboxsep}{0pt}\colorbox{white!0}{\parbox{0.2\textwidth}{  \colorbox{red!6.997497081756592}{\strut did} \colorbox{red!0.0}{\strut he} \colorbox{red!2.3131580352783203}{\strut catch} \colorbox{red!32.17988967895508}{\strut the} \colorbox{red!0.0}{\strut ball} \colorbox{red!100.00000762939453}{\strut ?}
}}}
\end{CJK*} &
    \begin{CJK*}{UTF8}{gbsn}
{\setlength{\fboxsep}{0pt}\colorbox{white!0}{\parbox{0.2\textwidth}{  \colorbox{red!1.4908345937728882}{\strut is} \colorbox{red!87.8625717163086}{\strut the} \colorbox{red!14.453324317932129}{\strut tub} \colorbox{red!0.0}{\strut white} \colorbox{red!73.8957748413086}{\strut ?}
}}}
\end{CJK*} &
    \begin{CJK*}{UTF8}{gbsn}
{\setlength{\fboxsep}{0pt}\colorbox{white!0}{\parbox{0.2\textwidth}{   \colorbox{red!3.7937772274017334}{\strut did} \colorbox{red!26.993452072143555}{\strut the} \colorbox{red!1.7138662338256836}{\strut man} \colorbox{red!10.991415023803711}{\strut just} \colorbox{red!6.126044273376465}{\strut catch} \colorbox{red!16.603017807006836}{\strut the} \colorbox{red!0.0}{\strut fr}\colorbox{red!0.0}{\strut is}\colorbox{red!1.6056874990463257}{\strut bee} \colorbox{red!100.0}{\strut ?}}}}
\end{CJK*}\\
\rotatebox[origin=c]{90}{~~~~~raw attn.} &
    \includegraphics[align=c,height=0.165\linewidth]{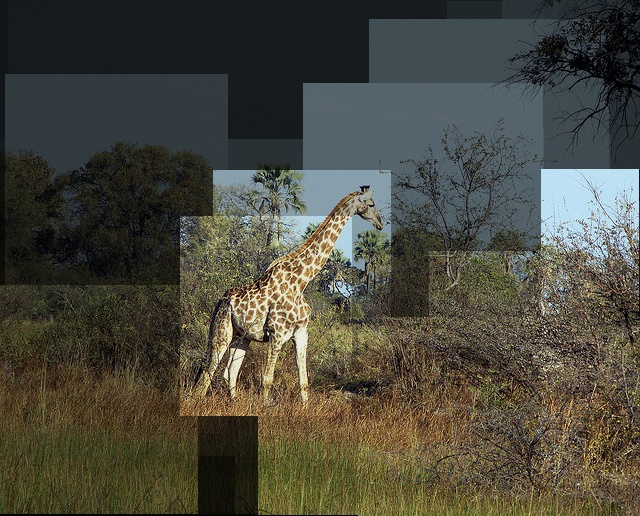} &
    \includegraphics[align=c,height=0.165\linewidth]{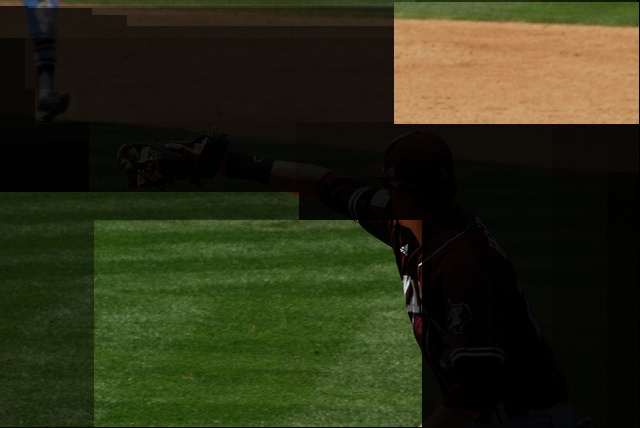} &
    \includegraphics[align=c,height=0.165\linewidth]{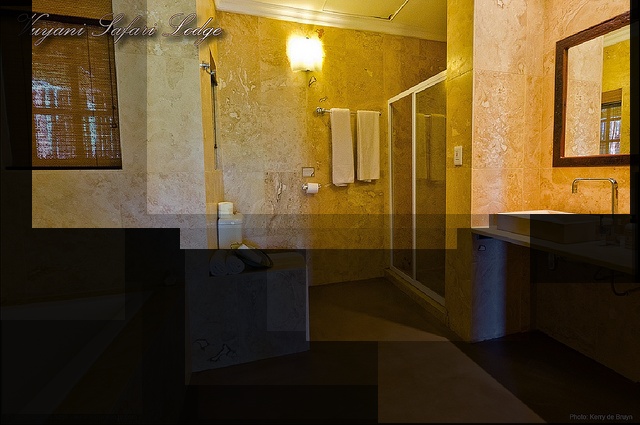} &
    \includegraphics[align=c,height=0.165\linewidth]{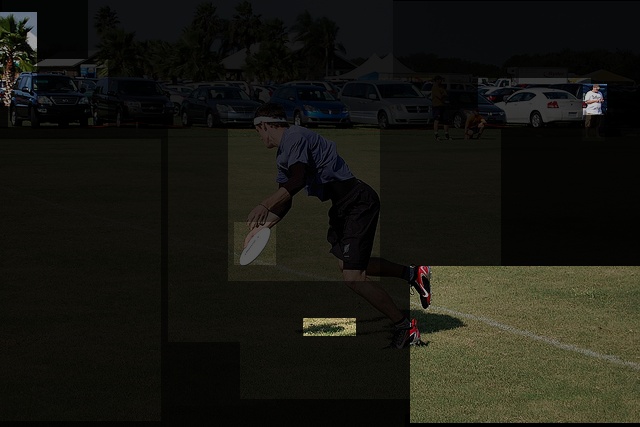} 
    \\


        &
    \begin{CJK*}{UTF8}{gbsn}
{\setlength{\fboxsep}{0pt}\colorbox{white!0}{\parbox{0.2\textwidth}{   \colorbox{red!0.0}{\strut is} \colorbox{red!47.170936584472656}{\strut the} \colorbox{red!8.081250190734863}{\strut animal} \colorbox{red!20.938230514526367}{\strut eating} \colorbox{red!14.840865135192871}{\strut ?}
}}}
\end{CJK*} &
    \begin{CJK*}{UTF8}{gbsn}
{\setlength{\fboxsep}{0pt}\colorbox{white!0}{\parbox{0.2\textwidth}{  \colorbox{red!6.312027931213379}{\strut did} \colorbox{red!0.0}{\strut he} \colorbox{red!10.827007293701172}{\strut catch} \colorbox{red!55.25422286987305}{\strut the} \colorbox{red!7.1024603843688965}{\strut ball} \colorbox{red!16.058643341064453}{\strut ?}
}}}
\end{CJK*} &
    \begin{CJK*}{UTF8}{gbsn}
{\setlength{\fboxsep}{0pt}\colorbox{white!0}{\parbox{0.2\textwidth}{  \colorbox{red!0.0}{\strut is} \colorbox{red!43.76888656616211}{\strut the} \colorbox{red!11.508163452148438}{\strut tub} \colorbox{red!1.344169020652771}{\strut white} \colorbox{red!12.809098243713379}{\strut ?}
}}}
\end{CJK*} &
    \begin{CJK*}{UTF8}{gbsn}
{\setlength{\fboxsep}{0pt}\colorbox{white!0}{\parbox{0.2\textwidth}{   \colorbox{red!7.530399322509766}{\strut did} \colorbox{red!23.75246238708496}{\strut the} \colorbox{red!0.0}{\strut man} \colorbox{red!5.081450462341309}{\strut just} \colorbox{red!15.589095115661621}{\strut catch} \colorbox{red!20.32027816772461}{\strut the} \colorbox{red!0.0}{\strut fr}\colorbox{red!2.322873830795288}{\strut is}\colorbox{red!2.8838162422180176}{\strut bee} \colorbox{red!14.764466285705566}{\strut ?}
}}}
\end{CJK*}\\
\rotatebox[origin=c]{90}{~~~~~rollout} &
    \includegraphics[align=c,height=0.165\linewidth]{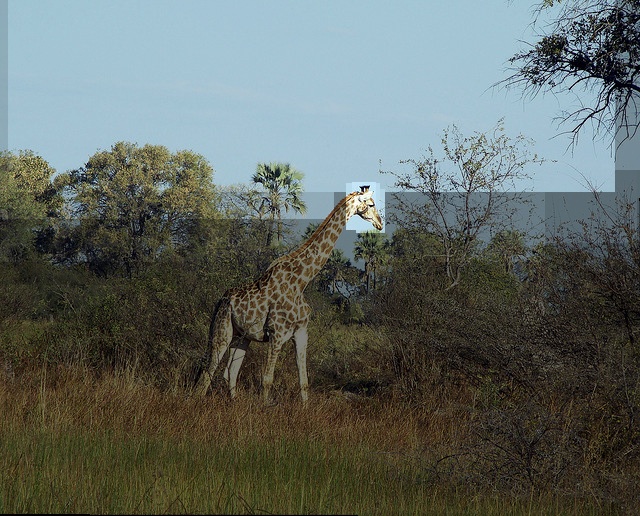} &
    \includegraphics[align=c,height=0.165\linewidth]{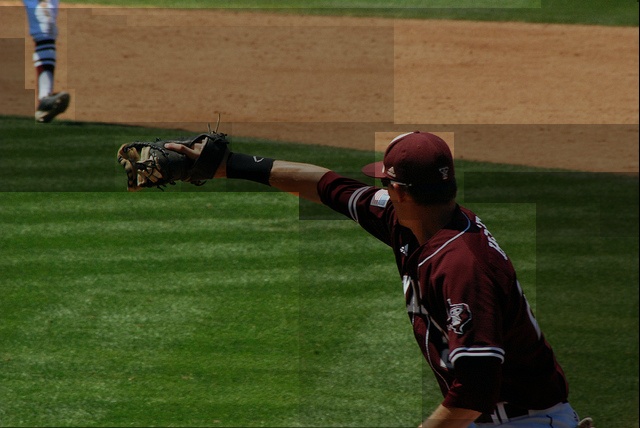} &
    \includegraphics[align=c,height=0.165\linewidth]{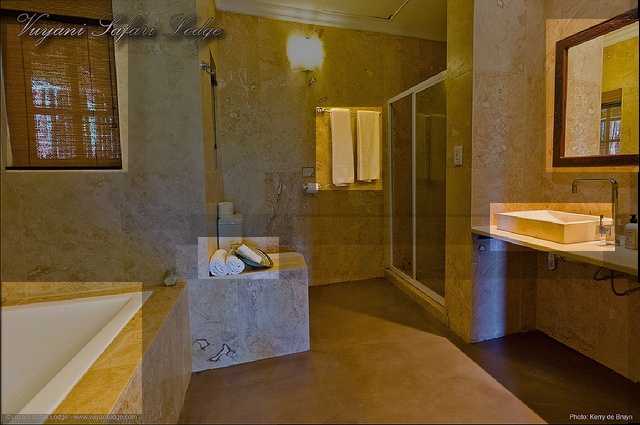} &
    \includegraphics[align=c,height=0.165\linewidth]{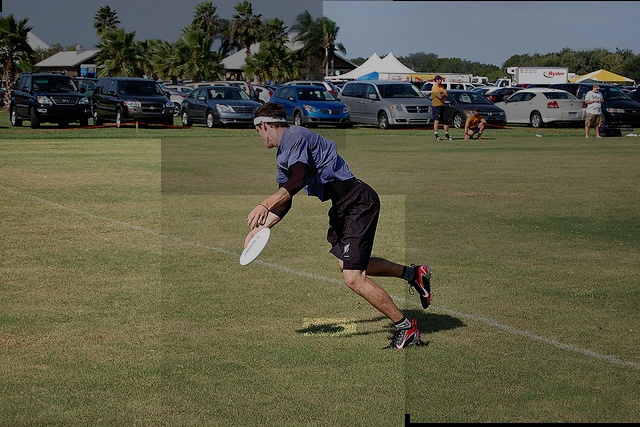} 
    \\

    
    \end{tabular*}
    \caption{A comparison between our method (top) and the baselines for VQA with the LXMERT model. Relevancy for text is given as shades of red. Relevancy for images is given by multiplying each region by the relative relevancy. Notice that both for the images and the text our method achieves favorable results. Answers (left to right): no, yes, yes, no. } 
    \label{fig:LXMERT-vis-full}
\end{figure*}

In Fig.~\ref{fig:LXMERT-vis-full} we present extended results for Fig.~4 in the paper, \ie we present the explanations extracted by each method for typical samples from the VQA dataset using the LXMERT model for question answering.

\section{Preparing the DETR relevancy maps for the COCO segmentation evaluation code}
In this section, we elaborate on the process of extracting segmentation masks from DETR's object detection results. The extracted segmentation masks are then used for our DETR tests, as presented in Sec.~5 of the paper.

DETR has been trained for object detection, \ie, producing a bounding box and a classification for each object in the input image. In order to evaluate the different explainability methods, we refer to the $\mathbf{R^{qi}}$ relevancy map, where the $j$-th row defines the relevance of each image feature to the $j$-th query, \ie the $j$-th bounding box, as described in Sec.~3.2 of the paper. Our test uses each of the explainability methods on the $5,000$ samples of the MSCOCO validation set to produce segmentation masks, as described in Alg.~\ref{alg:segm}. We first filter the queries to include only ones where the classification probability is higher than $50\%$ (Alg.~\ref{alg:segm}, L.~3). Then, for each query $j$ that is left, we use the relevancy matrix $\mathbf{R^{qi}}$ in row $j$ as a heatmap of the image features (Alg.~\ref{alg:segm}, L.~6), noting the important pixels for the $j$-th predicted bounding box. Since most of our baselines, as well as our method, produce non-negative relevancies, we use Otsu's thresholding method to separate the foreground and the background of the segmentation mask (Alg.~\ref{alg:segm}, L.~7). Then, the DETR segmentation evaluation code upsamples the masks to the target mask size, followed by a sigmoid operation, which only leaves the strictly positive values of the segmentation map (Alg.~\ref{alg:segm}, L.~8-9). Finally, the DETR segmentation evaluation code upsamples the generated map back to the size of the original image (Alg.~\ref{alg:segm}, L.~10). 

\begin{algorithm}[H]
\caption{Obtain Segmentation Masks from Heatmaps}\label{euclid}
$\bold{Input:}$ (i) input image (ii) $logits\in \mathbb{R}^{q\times c}$ obtained by the detection alg., where $q$ is the number of queries (bounding boxes), and $c$ is the number of object classes, 
(iii) $\mathbf{R^{qi}}$- relevancy matrix per query, from the explainability alg.\\
$\bold{Output:}$ $\textit{masks}\in \mathbb{R}^{q\times h\times w}$ where $q$ is the number of queries, and $h,w$ are the spatial dimensions of the input image. $\textit{masks}[j]$ is the segmentation map corresponding to the $j$-th bounding box.
\begin{algorithmic}[1]
\State $q \gets queries$
\State $\textit{probabilities} \gets softmax(logits)$
\State $\textit{keep} \gets j\in q\textit{, where max(probabilities}[j]) > 0.5$
\State $\textit{masks} \gets [[0, ..., 0], ..., [0, ..., 0]]$
\BState \emph{for $j\in$ keep}:
\State \hspace*{2em}$\textit{masks}[j] \gets \mathbf{R^{qi}}[j]$
\State \hspace*{2em}$\textit{masks}[j] \gets Otsu(\textit{masks}[j])$

\State \begin{varwidth}[t]{\linewidth}
      \hspace*{2em}$\textit{masks}[j] \gets Upsample(\textit{masks}[j],\textit{size=targetMaskSize, method="bilinear"})$
      \end{varwidth}

\State \hspace*{2em}$\textit{masks}[j] \gets \textit{sigmoid(masks[j])} > 0.5$


\State \begin{varwidth}[t]{\linewidth}
      \hspace*{2em}$\textit{masks}[j] \gets Upsample(\textit{masks}[j],\textit{size=origImageSize, method="nearest"})$
      \end{varwidth}

\end{algorithmic}
\label{alg:segm}
\end{algorithm}
\section{Ablation Study}

\begin{table}[!htb]
    \centering
    \begin{tabular}{lccccc}
        \toprule
        & Ours & w/o norm. &w/o aggregation & Eq.10 w/o self-att.\\
        \midrule
        AP & \textbf{13.1} & 11.7 &  0.1 & 11.5 \\
        $\text{AP}_{medium}$ & \textbf{14.4} & 13.9 & 0.0 & 13.8 \\
        $\text{AP}_{large}$ & \textbf{24.6} & 20.9 & 0.2& 20.5\\
         AR & \textbf{19.3} & 18.0 &  0.5 & 17.8 \\
        $\text{AR}_{medium}$ & \textbf{23.9} & 23.9 & 0.0 & 23.8 \\
        $\text{AR}_{large}$ & \textbf{33.2} & 29.2 & 1.0& 28.6\\
        \bottomrule
    \end{tabular}
    \caption{Performance for different ablation variants of our method on the DETR experiments. Higher is better.}
    \label{tab:ablation-DETR}
\end{table}

\begin{table}[!htb]
    \centering
    \begin{tabular}{lccccc}
        \toprule
        & Ours & w/o norm. & w/o aggregation & Eq.10 w/o self-att.\\
        \midrule
        Neg. img & \textbf{63.24} & 62.49 & 60.41 & 62.18 \\
        Pos. img & 51.10 & 50.60 & 60.40& \textbf{50.57} \\
        Neg. text & \textbf{48.70} & 48.64 & 41.72& 48.64\\
        Pos. text & 21.61 & \textbf{21.59} &41.72 & \textbf{21.59}\\
        \bottomrule
    \end{tabular}
    \caption{Area-under-the-curve for different ablation variants of our method on the LXMERT experiments. For negative perturbation, larger AUC is better; for positive perturbation, smaller AUC is better.}
    \label{tab:ablation-LXMERT}
\end{table}

We present three variations of our method. Firstly, we verify the effectiveness of our normalization for the self-attention relevancies presented in Eq.~8,9. Since the normalization is applied to rule 10, we expect it to affect mostly bi-modal relevancies, \ie the image perturbation experiments for LXMERT, and the DETR tests. The second ablation we present studies the necessity of the aggregation in all our rules~6,7,10,11, \ie instead of adding the former relevancy matrix to the newly constructed one, we only keep the new one, \eg for rule~6 the update becomes: $\mathbf{R}^{ss} = \mathbf{\bar{A}} \cdot  \mathbf{R}^{ss}$. Lastly, we explore the need for the self-attention updates to the bi-modal rule 10 by changing the update rule to: $\mathbf{R}^{sq} = \mathbf{R}^{sq} +\mathbf{\bar{A}}$. All our ablations are done on the LXMER, DETR experiments since, as mentioned several times, VisualBERT is based on pure self-attention, which yields similar results to the Transformer attribution baseline.

As can be seen from Tab.~\ref{tab:ablation-DETR}, all the components included in our method are crucial to its success on DETR, and the ablations cause a sizeable decrease in performance. It should be noted that for the reasonable ablations of not using normalization and not using self-attention in Eq.10, our ablations still outperform all other methods significantly for the DETR experiment. 

For the image perturbation test on LXMERT, presented in Tab.~\ref{tab:ablation-LXMERT}, we observe relatively mild differences between our method and the ablations of no normalization and no self-attention, this can be attributed to the fact that in contrast to DETR, LXMERT only uses 36 image regions that had gone through Non-maximum Suppression (NMS), therefore the added context from the self-attention to the multi-modal attention is not as crucial, since usually the top-1 image region is identical to that of the ablations, and is sufficient to make the classification.

\section{Using LRP with our method}

We present the results for the LXMERT perturbation tests evaluated by the area-under-the-curve measure for our method with LRP for completeness, \ie with head averaging as presented in the Transformer attribution method and in Eq.~13 instead of the head averaging in Eq.~5. The results in Tab.~\ref{tab:LRP} support and substantiate the conclusions presented in the paper: for the image perturbation tests, whether or not LRP is used, our method's contributions lead to a large gap in performance over all baseline methods. LRP itself leads to a small degradation in performance. For the text perturbation tests which are, as mentioned in the paper, self-attention based, our method is similar in performance to the Transformer Attribution method. Here, too, the choice of whether or not to use LRP is insignificant. Given the complexity of implementing LRP (see Sec.~4 of the main text), we advocate to eliminate it.

\begin{table}[h]
    \centering
    \begin{tabular}{lcccccccc}
        \toprule
        & Ours & Ours w/ LRP & Transformer att. & raw attn. & partial LRP & Grad-CAM & rollout\\
        \midrule
        Neg. img & \textbf{63.24} & 62.41 & 61.46 & 61.34 & 60.90 & 60.08 & 58.64  \\
        Pos. img & \textbf{51.10} & 51.10 & 52.75 & 54.17& 52.82 & 59.21 & 57.23  \\
        Neg. text & \textbf{48.70} & 48.25 & 48.24 & 38.32 & 45.15 & 37.99 & 32.05 \\
        Pos. text & \textbf{21.61} & 21.68 & 21.68 & 32.56 & 24.22 & 34.14 & 39.29 \\
        \bottomrule
    \end{tabular}
    \caption{Area-under-the-curve for all the baselines and our method with and without LRP on the LXMERT experiments. For negative perturbation, larger AUC is better; for positive perturbation, smaller AUC is better.}
    \label{tab:LRP}
\end{table}

\section{Perturbation experiments graphs}
\begin{figure*}%
    \centering
    \begin{tabular}{cccc}
    {{\includegraphics[width=6cm]{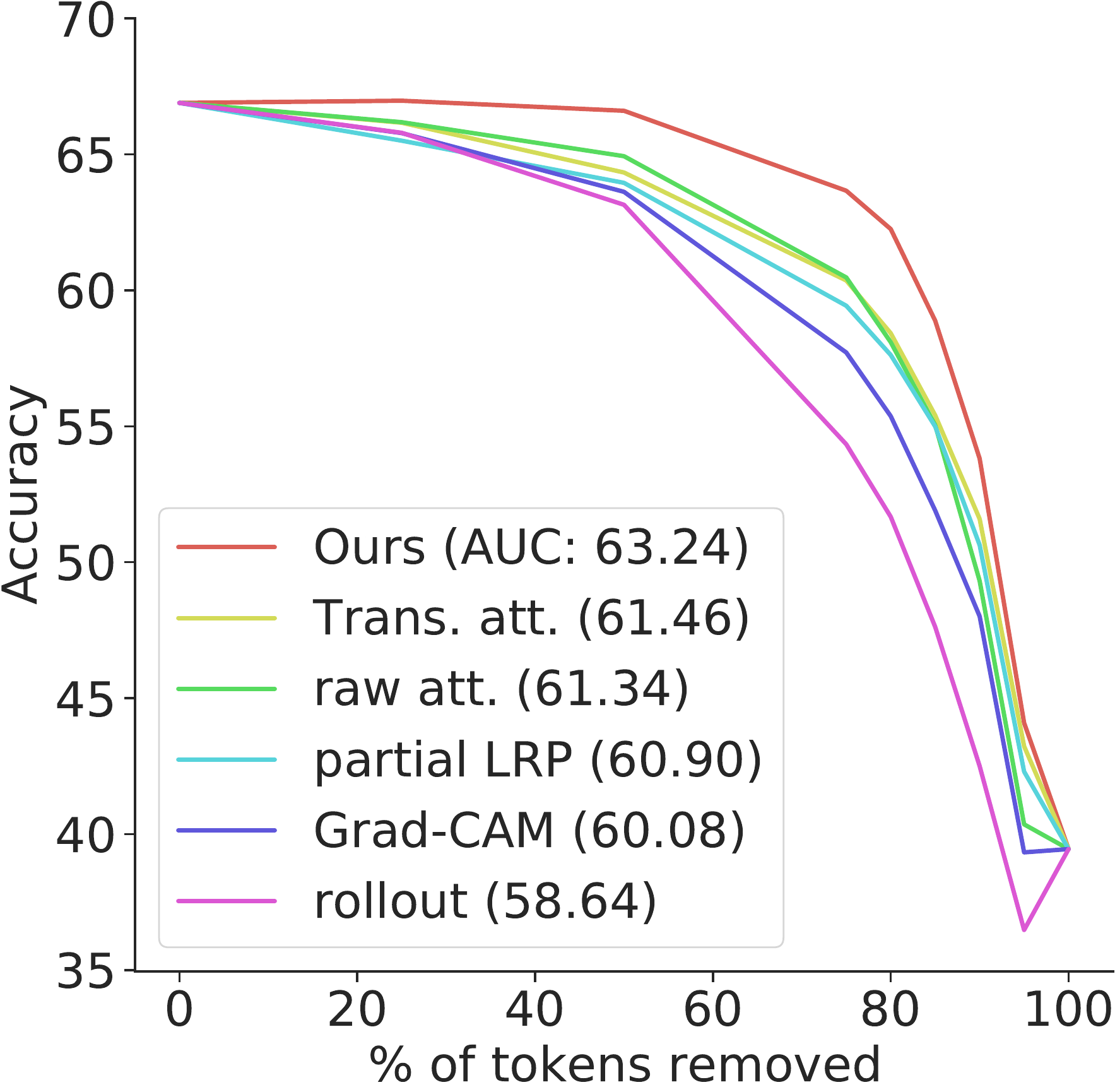} }}&%
    {{\includegraphics[width=6cm]{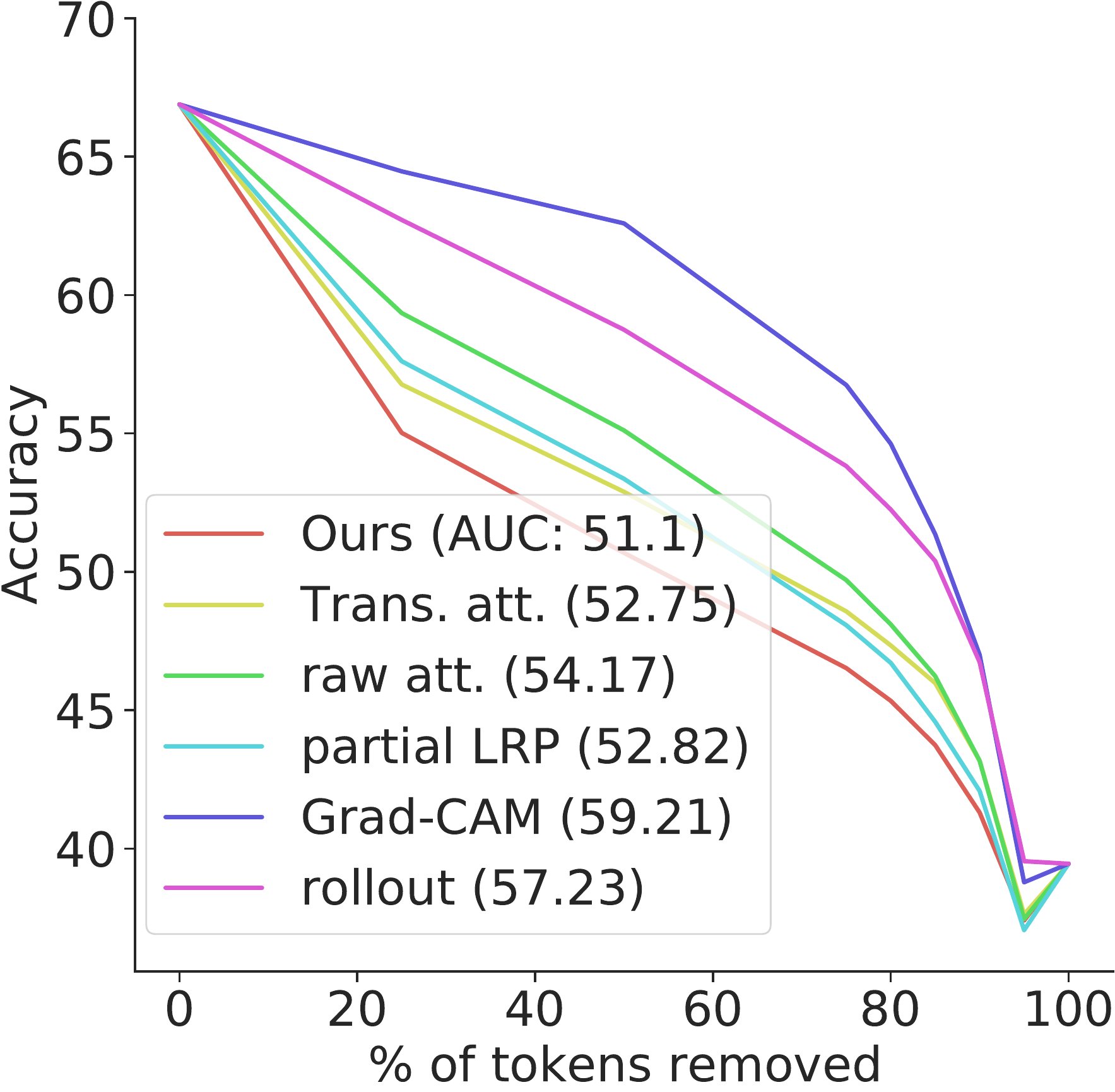} }}\\
    (a) & (b)\\
    {{\includegraphics[width=6cm]{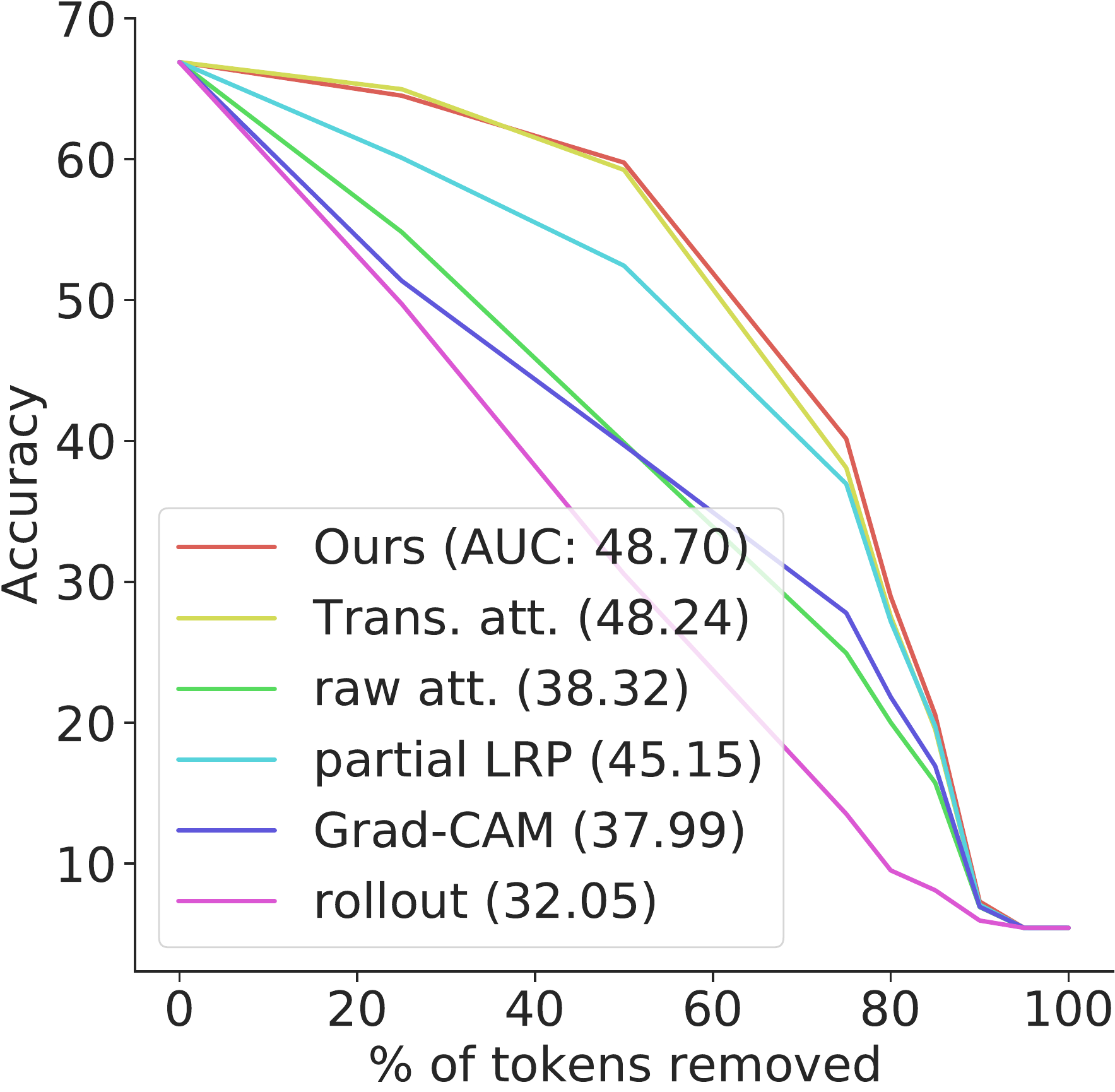} }}&%
    {{\includegraphics[width=6cm]{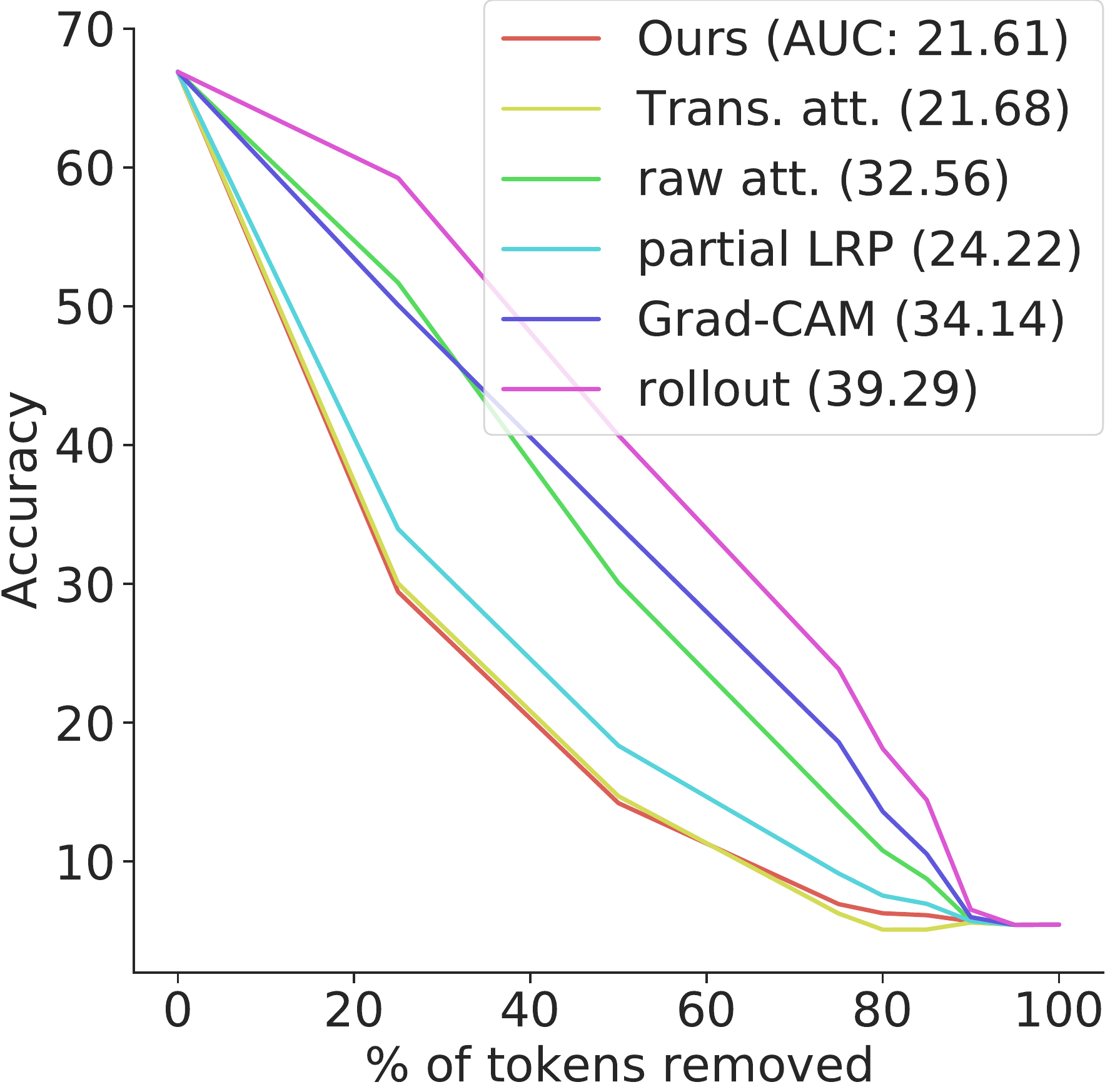} }}\\
    
    (c) & (d) \\
    \end{tabular}
    \caption{LXMERT perturbation test results. For negative perturbation, larger AUC is better; for positive perturbation, smaller AUC is better. (a) negative perturbation on image tokens, (b) positive perturbation on image tokens, (c) negative perturbation on text tokens, and (d) positive perturbation on text tokens.}%
    \label{fig:LXMERT-pert}%
\end{figure*}

\begin{figure*}%
    \centering
    \begin{tabular}{cccc}
    {{\includegraphics[width=6cm]{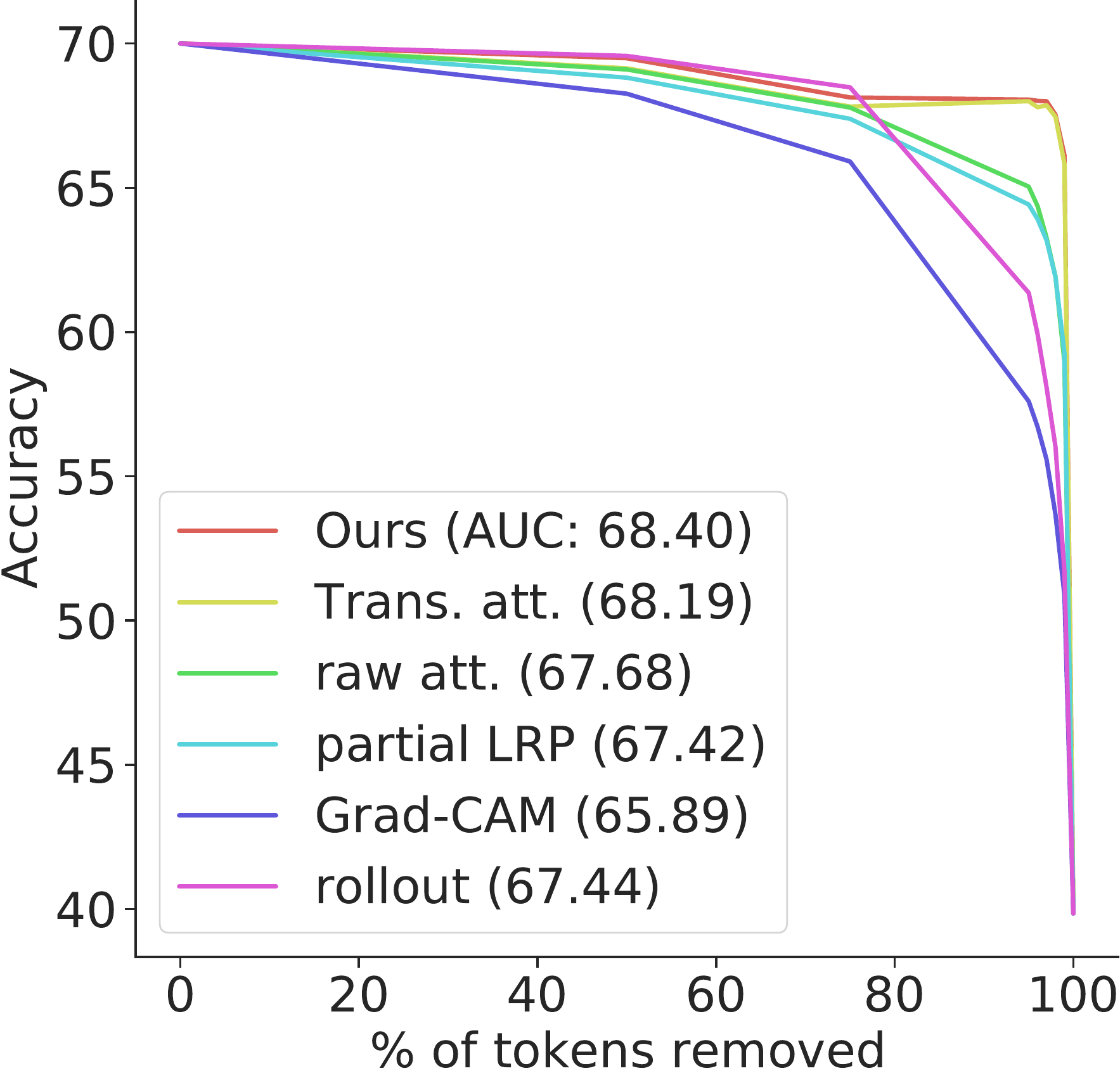} }}&%
    {{\includegraphics[width=6cm]{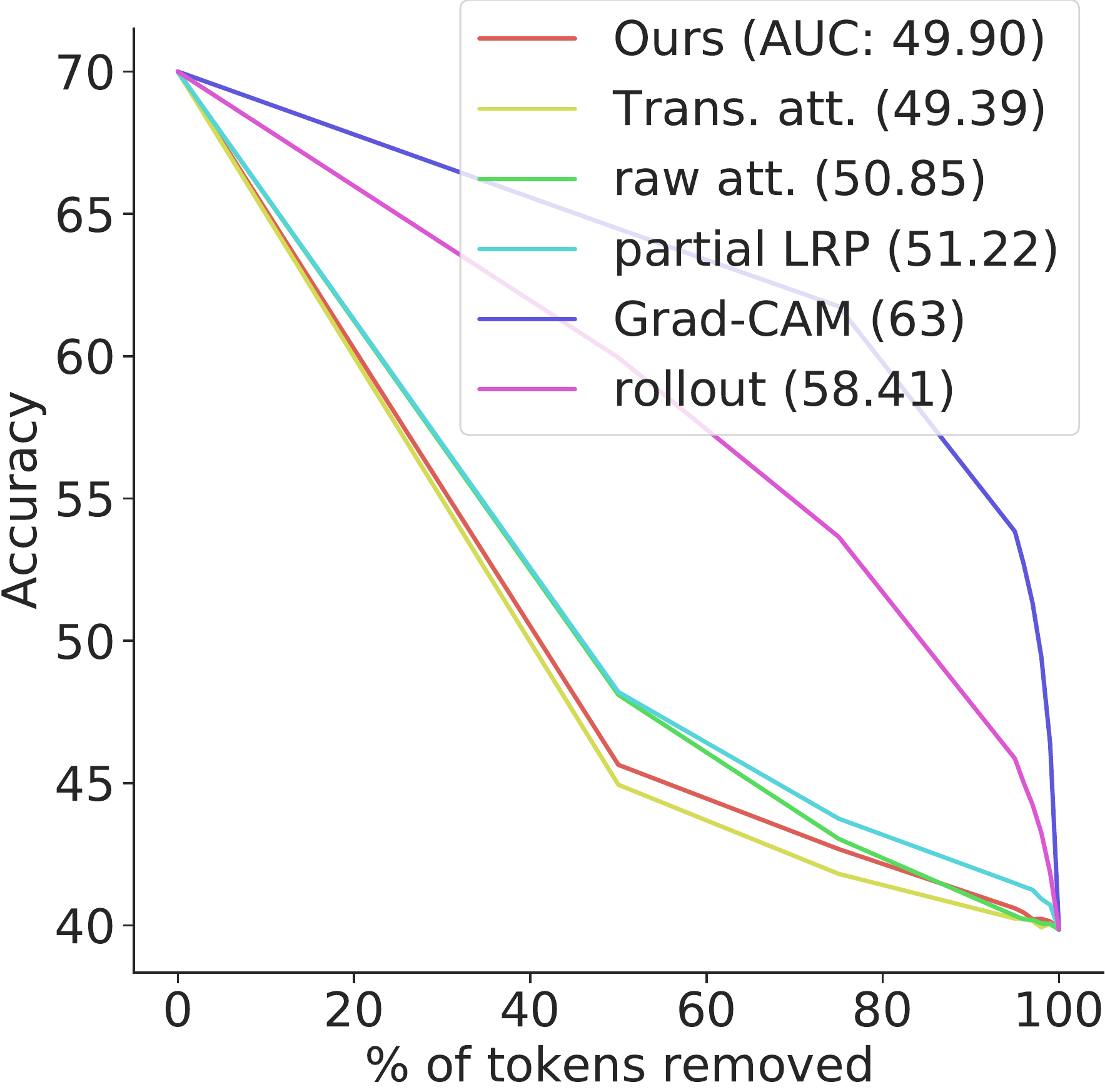} }}\\
    (a) & (b)\\
    {{\includegraphics[width=6cm]{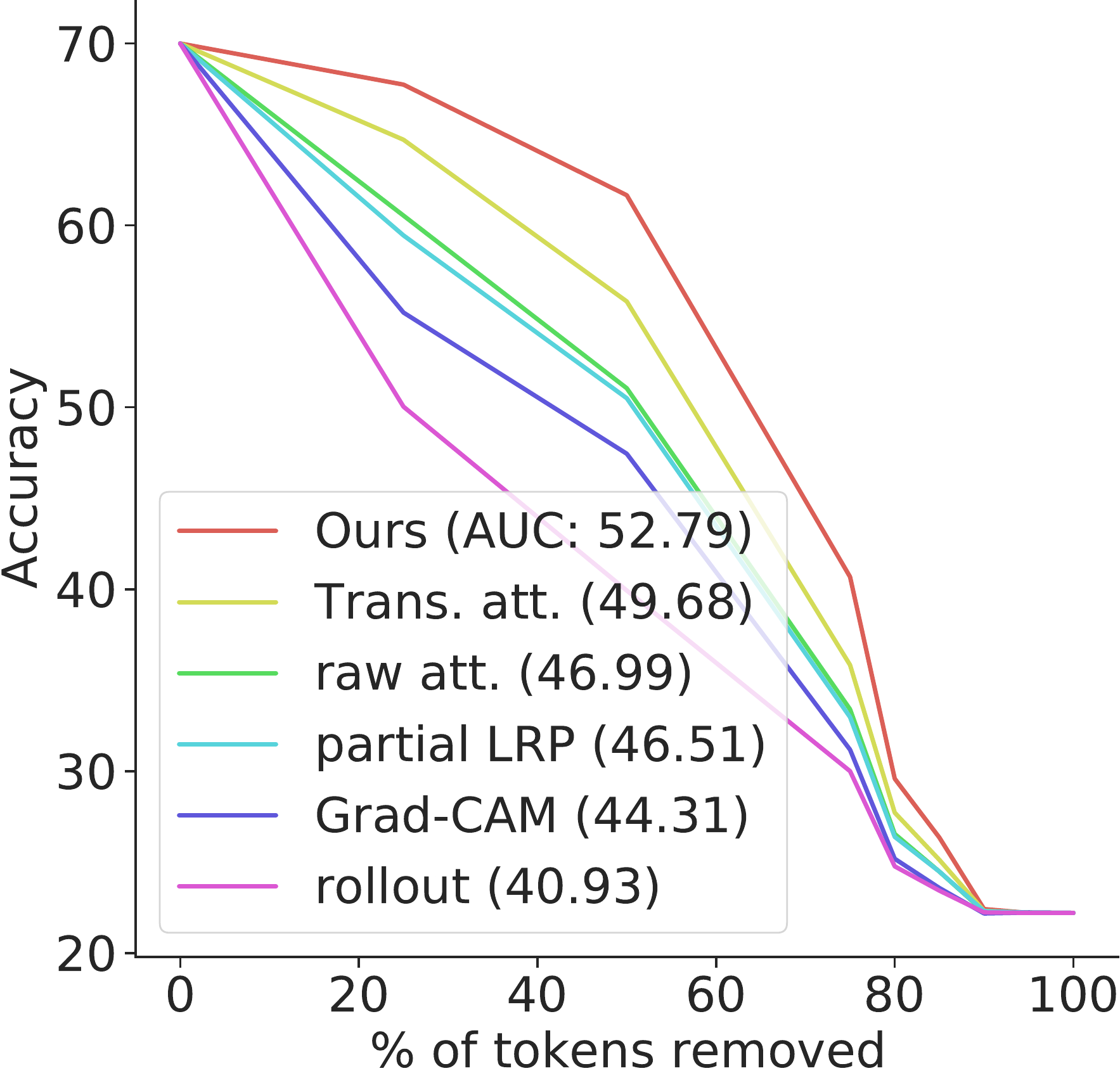} }}&%
    {{\includegraphics[width=6cm]{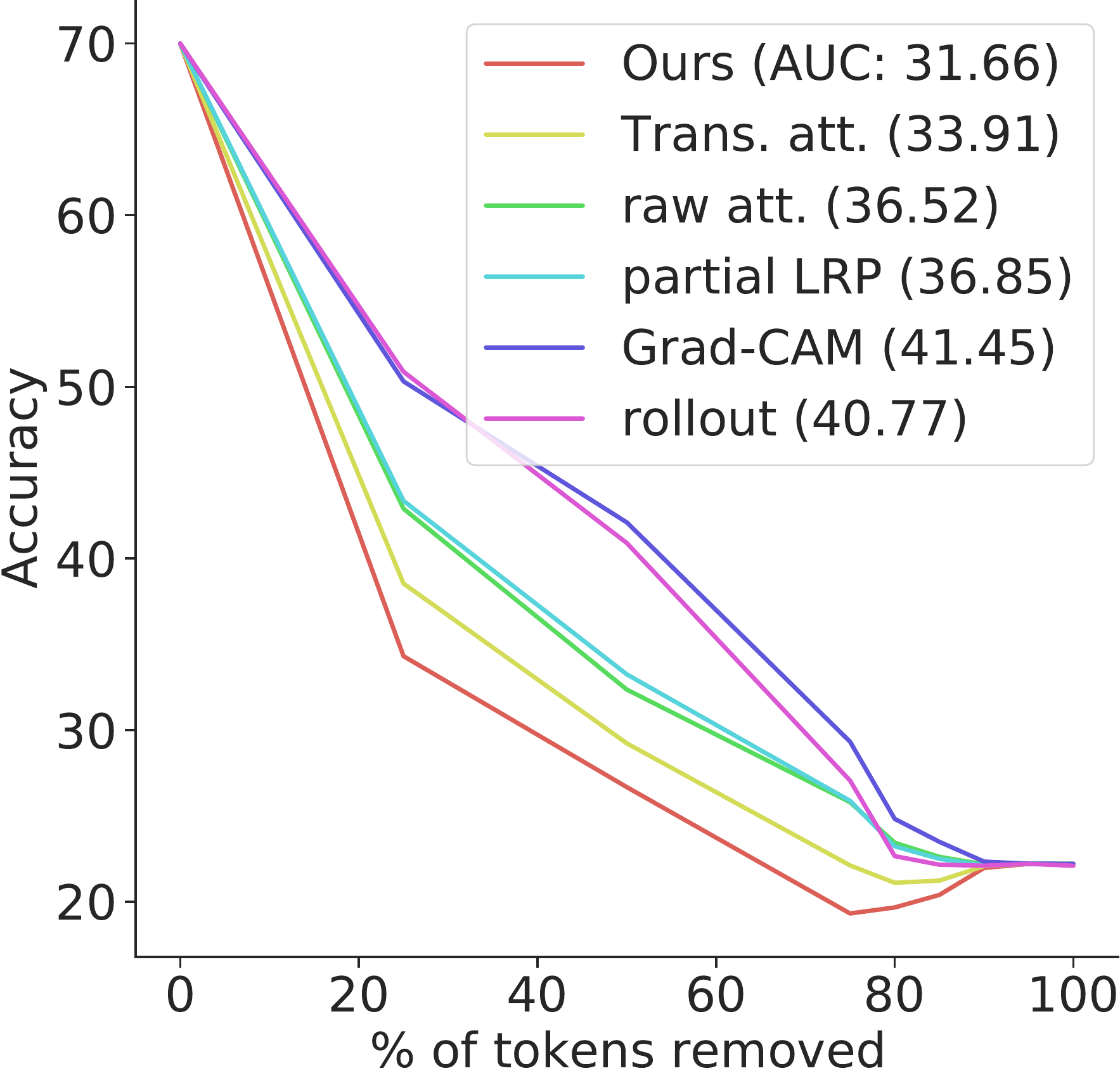} }}\\
    
    (c) & (d) \\
    \end{tabular}
    \caption{VisualBERT perturbation test results. For negative perturbation, larger AUC is better; for positive perturbation, smaller AUC is better. (a) negative perturbation on image tokens, (b) positive perturbation on image tokens, (c) negative perturbation on text tokens, and (d) positive perturbation on text tokens.}%
    \label{fig:VisualBERT-pert}%
\end{figure*}

In Fig.~\ref{fig:LXMERT-pert},~\ref{fig:VisualBERT-pert}, we present enlarged graphs corresponding to our perturbation experiments for better clarity. 

%% file: main.bbl
\begin{thebibliography}{10}\itemsep=-1pt

\bibitem{abnar2020quantifying}
Samira Abnar and Willem Zuidema.
\newblock Quantifying attention flow in transformers.
\newblock {\em arXiv preprint arXiv:2005.00928}, 2020.

\bibitem{Stanislaw2015vqa}
Stanislaw Antol, Aishwarya Agrawal, Jiasen Lu, Margaret Mitchell, Dhruv Batra,
  C.~Lawrence Zitnick, and Devi Parikh.
\newblock Vqa: Visual question answering.
\newblock In {\em Proceedings of the IEEE International Conference on Computer
  Vision (ICCV)}, December 2015.

\bibitem{bach2015pixel}
Sebastian Bach, Alexander Binder, Gr{\'e}goire Montavon, Frederick Klauschen,
  Klaus-Robert M{\"u}ller, and Wojciech Samek.
\newblock On pixel-wise explanations for non-linear classifier decisions by
  layer-wise relevance propagation.
\newblock {\em PloS one}, 10(7):e0130140, 2015.

\bibitem{carion2020end}
Nicolas Carion, Francisco Massa, Gabriel Synnaeve, Nicolas Usunier, Alexander
  Kirillov, and Sergey Zagoruyko.
\newblock End-to-end object detection with transformers.
\newblock {\em arXiv preprint arXiv:2005.12872}, 2020.

\bibitem{chefer2020transformer}
Hila Chefer, Shir Gur, and Lior Wolf.
\newblock Transformer interpretability beyond attention visualization.
\newblock {\em arXiv preprint arXiv:2012.09838}, 2020.

\bibitem{chen2018lshapley}
Jianbo Chen, Le Song, Martin~J. Wainwright, and Michael~I. Jordan.
\newblock L-shapley and c-shapley: Efficient model interpretation for
  structured data.
\newblock In {\em International Conference on Learning Representations}, 2019.

\bibitem{chen2020generative}
Mark Chen, Alec Radford, Rewon Child, Jeff Wu, Heewoo Jun, Prafulla Dhariwal,
  David Luan, and Ilya Sutskever.
\newblock Generative pretraining from pixels.
\newblock In {\em Proceedings of the 37th International Conference on Machine
  Learning}, volume~1, 2020.

\bibitem{dabkowski2017real}
Piotr Dabkowski and Yarin Gal.
\newblock Real time image saliency for black box classifiers.
\newblock In {\em Advances in Neural Information Processing Systems}, pages
  6970--6979, 2017.

\bibitem{dosovitskiy2020image}
Alexey Dosovitskiy, Lucas Beyer, Alexander Kolesnikov, Dirk Weissenborn,
  Xiaohua Zhai, Thomas Unterthiner, Mostafa Dehghani, Matthias Minderer, Georg
  Heigold, Sylvain Gelly, et~al.
\newblock An image is worth 16x16 words: Transformers for image recognition at
  scale.
\newblock {\em arXiv preprint arXiv:2010.11929}, 2020.

\bibitem{erhan2009visualizing}
Dumitru Erhan, Yoshua Bengio, Aaron Courville, and Pascal Vincent.
\newblock Visualizing higher-layer features of a deep network.
\newblock {\em University of Montreal}, 1341(3):1, 2009.

\bibitem{fong2019understanding}
Ruth Fong, Mandela Patrick, and Andrea Vedaldi.
\newblock Understanding deep networks via extremal perturbations and smooth
  masks.
\newblock In {\em Proceedings of the IEEE International Conference on Computer
  Vision}, pages 2950--2958, 2019.

\bibitem{fong2017interpretable}
Ruth~C Fong and Andrea Vedaldi.
\newblock Interpretable explanations of black boxes by meaningful perturbation.
\newblock In {\em Proceedings of the IEEE International Conference on Computer
  Vision}, pages 3429--3437, 2017.

\bibitem{gu2018understanding}
Jindong Gu, Yinchong Yang, and Volker Tresp.
\newblock Understanding individual decisions of cnns via contrastive
  backpropagation.
\newblock In {\em Asian Conference on Computer Vision}, pages 119--134.
  Springer, 2018.

\bibitem{gur2021visualization}
Shir Gur, Ameen Ali, and Lior Wolf.
\newblock Visualization of supervised and self-supervised neural networks via
  attribution guided factorization.
\newblock In {\em AAAI}, 2021.

\bibitem{he2016deep}
Kaiming He, Xiangyu Zhang, Shaoqing Ren, and Jian Sun.
\newblock Deep residual learning for image recognition.
\newblock In {\em Proceedings of the IEEE Conference on Computer Vision and
  Pattern Recognition}, pages 770--778, 2016.

\bibitem{iwana2019explaining}
Brian~Kenji Iwana, Ryohei Kuroki, and Seiichi Uchida.
\newblock Explaining convolutional neural networks using softmax gradient
  layer-wise relevance propagation.
\newblock {\em arXiv preprint arXiv:1908.04351}, 2019.

\bibitem{lewis2019bart}
Mike Lewis, Yinhan Liu, Naman Goyal, Marjan Ghazvininejad, Abdelrahman Mohamed,
  Omer Levy, Ves Stoyanov, and Luke Zettlemoyer.
\newblock Bart: Denoising sequence-to-sequence pre-training for natural
  language generation, translation, and comprehension.
\newblock {\em arXiv preprint arXiv:1910.13461}, 2019.

\bibitem{li2019visualbert}
Liunian~Harold Li, Mark Yatskar, Da Yin, Cho-Jui Hsieh, and Kai-Wei Chang.
\newblock Visualbert: A simple and performant baseline for vision and language.
\newblock {\em arXiv preprint arXiv:1908.03557}, 2019.

\bibitem{li2020oscar}
Xiujun Li, Xi Yin, Chunyuan Li, Pengchuan Zhang, Xiaowei Hu, Lei Zhang, Lijuan
  Wang, Houdong Hu, Li Dong, Furu Wei, et~al.
\newblock Oscar: Object-semantics aligned pre-training for vision-language
  tasks.
\newblock In {\em European Conference on Computer Vision}, pages 121--137.
  Springer, 2020.

\bibitem{ty2014coco}
Tsung-Yi Lin, Michael Maire, Serge Belongie, James Hays, Pietro Perona, Deva
  Ramanan, Piotr Doll{\'a}r, and C~Lawrence Zitnick.
\newblock Microsoft coco: Common objects in context.
\newblock In {\em European conference on computer vision}, pages 740--755.
  Springer, 2014.

\bibitem{lu2019vilbert}
Jiasen Lu, Dhruv Batra, Devi Parikh, and Stefan Lee.
\newblock Vilbert: Pretraining task-agnostic visiolinguistic representations
  for vision-and-language tasks.
\newblock In {\em Advances in Neural Information Processing Systems}, pages
  13--23, 2019.

\bibitem{lundberg2017unified}
Scott~M Lundberg and Su-In Lee.
\newblock A unified approach to interpreting model predictions.
\newblock In {\em Advances in Neural Information Processing Systems}, pages
  4765--4774, 2017.

\bibitem{mahendran2016visualizing}
Aravindh Mahendran and Andrea Vedaldi.
\newblock Visualizing deep convolutional neural networks using natural
  pre-images.
\newblock {\em International Journal of Computer Vision}, 120(3):233--255,
  2016.

\bibitem{montavon2017explaining}
Gr{\'e}goire Montavon, Sebastian Lapuschkin, Alexander Binder, Wojciech Samek,
  and Klaus-Robert M{\"u}ller.
\newblock Explaining nonlinear classification decisions with deep taylor
  decomposition.
\newblock {\em Pattern Recognition}, 65:211--222, 2017.

\bibitem{nam2019relative}
Woo-Jeoung Nam, Shir Gur, Jaesik Choi, Lior Wolf, and Seong-Whan Lee.
\newblock Relative attributing propagation: Interpreting the comparative
  contributions of individual units in deep neural networks.
\newblock {\em arXiv preprint arXiv:1904.00605}, 2019.

\bibitem{otsu1979threshold}
Nobuyuki Otsu.
\newblock A threshold selection method from gray-level histograms.
\newblock {\em IEEE transactions on systems, man, and cybernetics},
  9(1):62--66, 1979.

\bibitem{paul2021local}
Matthieu Paul, Martin Danelljan, Luc Van~Gool, and Radu Timofte.
\newblock Local memory attention for fast video semantic segmentation.
\newblock {\em arXiv preprint arXiv:2101.01715}, 2021.

\bibitem{radford2021learning}
Alec Radford, Jong~Wook Kim, Chris Hallacy, Aditya Ramesh, Gabriel Goh,
  Sandhini Agarwal, Girish Sastry, Amanda Askell, Pamela Mishkin, Jack Clark,
  et~al.
\newblock Learning transferable visual models from natural language
  supervision.
\newblock {\em arXiv preprint arXiv:2103.00020}, 2021.

\bibitem{raffel2019exploring}
Colin Raffel, Noam Shazeer, Adam Roberts, Katherine Lee, Sharan Narang, Michael
  Matena, Yanqi Zhou, Wei Li, and Peter~J Liu.
\newblock Exploring the limits of transfer learning with a unified text-to-text
  transformer.
\newblock {\em arXiv preprint arXiv:1910.10683}, 2019.

\bibitem{ramesh2021zero}
Aditya Ramesh, Mikhail Pavlov, Gabriel Goh, Scott Gray, Chelsea Voss, Alec
  Radford, Mark Chen, and Ilya Sutskever.
\newblock Zero-shot text-to-image generation.
\newblock {\em arXiv preprint arXiv:2102.12092}, 2021.

\bibitem{russakovsky2015ImageNet}
Olga Russakovsky, Jia Deng, Hao Su, Jonathan Krause, Sanjeev Satheesh, Sean Ma,
  Zhiheng Huang, Andrej Karpathy, Aditya Khosla, Michael Bernstein, et~al.
\newblock Imagenet large scale visual recognition challenge.
\newblock {\em International journal of computer vision}, 115(3):211--252,
  2015.

\bibitem{selvaraju2017grad}
Ramprasaath~R Selvaraju, Michael Cogswell, Abhishek Das, Ramakrishna Vedantam,
  Devi Parikh, and Dhruv Batra.
\newblock Grad-cam: Visual explanations from deep networks via gradient-based
  localization.
\newblock In {\em Proceedings of the IEEE international conference on computer
  vision}, pages 618--626, 2017.

\bibitem{shrikumar2017learning}
Avanti Shrikumar, Peyton Greenside, and Anshul Kundaje.
\newblock Learning important features through propagating activation
  differences.
\newblock In {\em Proceedings of the 34th International Conference on Machine
  Learning-Volume 70}, pages 3145--3153. JMLR. org, 2017.

\bibitem{simonyan2013deep}
Karen Simonyan, Andrea Vedaldi, and Andrew Zisserman.
\newblock Deep inside convolutional networks: Visualising image classification
  models and saliency maps.
\newblock {\em arXiv preprint arXiv:1312.6034}, 2013.

\bibitem{smilkov2017smoothgrad}
Daniel Smilkov, Nikhil Thorat, Been Kim, Fernanda Vi{\'e}gas, and Martin
  Wattenberg.
\newblock Smoothgrad: removing noise by adding noise.
\newblock {\em arXiv preprint arXiv:1706.03825}, 2017.

\bibitem{srinivas2019full}
Suraj Srinivas and Fran{\c{c}}ois Fleuret.
\newblock Full-gradient representation for neural network visualization.
\newblock In {\em Advances in Neural Information Processing Systems}, pages
  4126--4135, 2019.

\bibitem{sundararajan2017axiomatic}
Mukund Sundararajan, Ankur Taly, and Qiqi Yan.
\newblock Axiomatic attribution for deep networks.
\newblock In {\em Proceedings of the 34th International Conference on Machine
  Learning-Volume 70}, pages 3319--3328. JMLR. org, 2017.

\bibitem{tan2019lxmert}
Hao Tan and Mohit Bansal.
\newblock Lxmert: Learning cross-modality encoder representations from
  transformers.
\newblock {\em arXiv preprint arXiv:1908.07490}, 2019.

\bibitem{touvron2020training}
Hugo Touvron, Matthieu Cord, Matthijs Douze, Francisco Massa, Alexandre
  Sablayrolles, and Herv{\'e} J{\'e}gou.
\newblock Training data-efficient image transformers \& distillation through
  attention.
\newblock {\em arXiv preprint arXiv:2012.12877}, 2020.

\bibitem{vaswani2017attention}
Ashish Vaswani, Noam Shazeer, Niki Parmar, Jakob Uszkoreit, Llion Jones,
  Aidan~N Gomez, {\L}ukasz Kaiser, and Illia Polosukhin.
\newblock Attention is all you need.
\newblock In {\em Advances in neural information processing systems}, pages
  5998--6008, 2017.

\bibitem{voita2019analyzing}
Elena Voita, David Talbot, Fedor Moiseev, Rico Sennrich, and Ivan Titov.
\newblock Analyzing multi-head self-attention: Specialized heads do the heavy
  lifting, the rest can be pruned.
\newblock In {\em Proceedings of the 57th Annual Meeting of the Association for
  Computational Linguistics}, pages 5797--5808, 2019.

\bibitem{wang2020max}
Huiyu Wang, Yukun Zhu, Hartwig Adam, Alan Yuille, and Liang-Chieh Chen.
\newblock Max-deeplab: End-to-end panoptic segmentation with mask transformers.
\newblock {\em arXiv preprint arXiv:2012.00759}, 2020.

\bibitem{wang2020end}
Yuqing Wang, Zhaoliang Xu, Xinlong Wang, Chunhua Shen, Baoshan Cheng, Hao Shen,
  and Huaxia Xia.
\newblock End-to-end video instance segmentation with transformers.
\newblock {\em arXiv preprint arXiv:2011.14503}, 2020.

\bibitem{zeiler2014visualizing}
Matthew~D Zeiler and Rob Fergus.
\newblock Visualizing and understanding convolutional networks.
\newblock In {\em European conference on computer vision}, pages 818--833.
  Springer, 2014.

\bibitem{zhang2018top}
Jianming Zhang, Sarah~Adel Bargal, Zhe Lin, Jonathan Brandt, Xiaohui Shen, and
  Stan Sclaroff.
\newblock Top-down neural attention by excitation backprop.
\newblock {\em International Journal of Computer Vision}, 126(10):1084--1102,
  2018.

\bibitem{zheng2020rethinking}
Sixiao Zheng, Jiachen Lu, Hengshuang Zhao, Xiatian Zhu, Zekun Luo, Yabiao Wang,
  Yanwei Fu, Jianfeng Feng, Tao Xiang, Philip~HS Torr, et~al.
\newblock Rethinking semantic segmentation from a sequence-to-sequence
  perspective with transformers.
\newblock {\em arXiv preprint arXiv:2012.15840}, 2020.

\bibitem{zhou2018interpreting}
Bolei Zhou, David Bau, Aude Oliva, and Antonio Torralba.
\newblock Interpreting deep visual representations via network dissection.
\newblock {\em IEEE transactions on pattern analysis and machine intelligence},
  2018.

\bibitem{zhou2016learning}
Bolei Zhou, Aditya Khosla, Agata Lapedriza, Aude Oliva, and Antonio Torralba.
\newblock Learning deep features for discriminative localization.
\newblock In {\em Proceedings of the IEEE conference on computer vision and
  pattern recognition}, pages 2921--2929, 2016.

\bibitem{zhu2021deformable}
Xizhou Zhu, Weijie Su, Lewei Lu, Bin Li, Xiaogang Wang, and Jifeng Dai.
\newblock Deformable {\{}detr{\}}: Deformable transformers for end-to-end
  object detection.
\newblock In {\em International Conference on Learning Representations}, 2021.

\end{thebibliography}
